\documentclass[a4paper, 11pt]{article}

\usepackage[T1]{fontenc}

\usepackage[left=2.5cm, top=2.5cm, right=2.5cm, bottom=2.5cm]{geometry}

\usepackage{amsmath}
\usepackage{mathtools}

\usepackage{etoolbox} 
\appto\appendix{\counterwithin{equation}{section}}

\usepackage{setspace}

\usepackage{booktabs}

\usepackage{tabularx}

\usepackage{threeparttable}

\usepackage{adjustbox}

\usepackage{multirow}

\usepackage{lipsum}

\usepackage{verbatim}

\usepackage[numbers]{natbib}
\usepackage{caption}
\usepackage[hidelinks]{hyperref}

\usepackage{siunitx}

\usepackage{subcaption}

\usepackage{chngcntr}

\title{\bf{Consistent multi-animal pose estimation in cattle using dynamic Kalman filter based tracking}}
\author{Maarten Perneel\textsuperscript{a,b}*  (ORCID 0000-0002-4831-5357)\and
        Ines Adriaens\textsuperscript{b} (ORCID 0000-0001-9768-2308)\and 
        Ben Aernouts\textsuperscript{a}$\dagger$ (ORCID 0000-0001-6266-3019)\and
        Jan Verwaeren\textsuperscript{b}$\dagger$ (ORCID 0000-0001-9978-501X)}
\date{%
    \begin{flushleft}
    \small
    \textsuperscript{a}Department of Biosystems, Division of Animal and Human Health Engineering, KU Leuven, Campus Geel, Kleinhoefstraat 4, 2440 Geel, Belgium\\
    \textsuperscript{b}Department of Data Analysis and Mathematical Modelling, Faculty of Bioscience Engineering, Ghent University, Coupure Links 653, 9000 Ghent, Belgium\\
    $\dagger$ Authors contributed equally to this work\\[2ex]
    \normalsize
    * Corresponding author:\\
    Maarten Perneel\\
    maarten.perneel@kuleuven.be\\
    \end{flushleft}    
}

\usepackage{graphicx}

\usepackage{cleveref}

\usepackage{dsfont} 

\usepackage{tikz} 
\usetikzlibrary{3d}
\usetikzlibrary{calc}
\usetikzlibrary{positioning}

\usepackage{xcolor}

\renewcommand{\vec}[1]{\boldsymbol{#1}}

\definecolor{customyellow}{RGB}{255, 191, 0}
\definecolor{customblue}{RGB}{0, 64, 255}

\begin{document}	
    \maketitle
    
    
    \section*{Abstract}
    Over the past decade, studying animal behaviour with the help of computer vision has become more popular. Replacing human observers by computer vision lowers the cost of data collection and therefore allows to collect more extensive datasets. However, the majority of available computer vision algorithms to study animal behaviour is highly tailored towards a single research objective, limiting possibilities for data reuse. In this perspective, pose-estimation in combination with animal tracking offers opportunities to yield a higher level representation capturing both the spatial and temporal component of animal behaviour. Such a higher level representation allows to answer a wide variety of research questions simultaneously, without the need to develop repeatedly tailored computer vision algorithms. In this paper, we therefore first cope with several weaknesses of current pose-estimation algorithms and thereafter introduce KeySORT (Keypoint Simple and Online Realtime Tracking). KeySORT deploys an adaptive Kalman filter to construct tracklets in a bounding-box free manner, significantly improving the temporal consistency of detected keypoints. In this paper, we focus on pose estimation in cattle, but our methodology can easily be generalised to any other animal species. Our test results indicate our algorithm is able to detect up to 80\% of the ground truth keypoints with high accuracy, with only a limited drop in performance when daylight recordings are compared to nightvision recordings. Moreover, by using KeySORT to construct skeletons, the temporal consistency of generated keypoint coordinates was largely improved, offering opportunities with regard to automated behaviour monitoring of animals.  
    
    \textbf{Keywords}: pose estimation, pose tracking, adaptive Kalman filter, behaviour monitoring
	
	\newpage
	
    \section{Introduction}    
 
    Dairy cattle show an extensive behavioural repertoire. The behaviour of an individual or group of animals is often used as an indicator for animal health and welfare, and can be linked to productivity \citep{Magrin_2023}. However, the manual observation of animal behaviour is a laborious task and the observer can be concentrated for only a limited timespan \citep{De_Freslon_2019}. Recent work has shown that, for some behaviours such as mounting and drinking, the human observer can successfully be replaced by a computer vision system \citep{Lodkaew_2023,Tsai_2020}. However, for other behaviours, especially social behaviours involving several animals simultaneously, it proves difficult to replace the human observer completely by a computer vision system. Moreover, the majority of computer vision algorithms developed in the past is highly tailored towards a single research objective. For example, \citep{Tsai_2020} studied the influence of heat stress on the drinking behaviour of cattle, while \citep{Achour_2020} monitored the feeding behaviour of individual animals. These tailored approaches, however, limit the possibilities for data reuse. Alternatively, video data could first be converted into a generic higher level representation. If designed thoughtful, such a representation can retain most of the relevant information, while having a lower memory requirement \citep{Li_2014, Kumar_2009, Song_2014}. Moreover, if the representation space is constructed with meaningful features, subsequent extraction of behavioural information of interest can be performed easily and data reuse will be facilitated. In the case of dairy cattle, such a higher level representation space could be defined by the location of several behaviour-determining keypoints, such as the nose, withers and tail implant. To capture both the spatial and temporal behaviour patterns, these keypoints should not only be detected on separate video frames, but also be tracked. However, current research considers pose estimation and tracking independently from each other \citep{Lauer_2022, Pereira_2022}, whereas the integration of both of them offers remarkable opportunities to improve the accuracy and consistency of the obtained keypoint locations. In this work, we propose an algorithm which integrates pose-estimation and keypoint-tracking. Our algorithm generates for each animal in a pen a pre-defined set of keypoints in a reliable, consistent and computationally efficient manner. As such, the problem we solve is an instance of the more general multiple-object pose estimation and tracking problem. We optimised the algorithm for group-housed dairy cattle, but it can easily be generalised to different animal species or object categories. 
    
    In barn environments, several animals are present in most frames, requiring multi-animal pose estimation (MAPE) algorithms that allow estimation of the pose of each animal per frame. MAPE involves both the detection of keypoints and grouping of the detected keypoints per animal. Classical top-down methods originating from multi-person pose estimation \citep{Fang_2017, Iqbal_2016} handle this problem by first detecting individuals in a frame, often using bounding box detectors, after which keypoint detection is performed for each detected bounding box. On the other hand, most recently published MAPE algorithms such as Deeplabcut \citep{Lauer_2022} and SLEAP \citep{Pereira_2022} use bottom-up methods to perform MAPE. Bottom-up methods first detect keypoints in a frame and associate keypoints per animal in a second step to construct a single skeleton per animal. For the keypoint association step, generally part affinity fields are used, which were introduced by \citep{Cao_2021}, but were already used before under a slightly different form by \citep{Psota_2019}. According to the implementation of \citep{Psota_2019}, for two connected keypoints A and B, such as the tail implant and the shoulders of a pig, part affinity fields indicate, conditionally on the presence of a keypoint of category A, where a corresponding keypoint of category B could be located and vice versa. When compared to top-down implementations of MAPE, bottom-up approaches have two advantages that are important for our application: they allow faster inference and provide more accurate results in dense scenes \citep{Topham_2023}. Especially the last aspect is highly relevant when studying socially interacting animals.


    To study social behaviours and make a distinction between and agonistic and affiliative behaviours, not only the spatial component of animal behaviour, but also the temporal component should be analysed \citep{Ren_2021, Foris_2019}. Therefore, tracking of detected skeletons is necessary. Moreover, since some social behaviours take place on the sub-second to seconds scale, temporal frequency of the tracklets should be relatively high while keeping the computational cost of constructing them feasible. A popular approach to track objects is tracking by re-identification \citep{Wojke_2017, Mahmoudi_2019, Fang_2018}. However, most re-identification algorithms are based on neural networks and require the availability of segments or object detections of the individuals of interest, which would nullify any computational advantage of using bottom-up MAPE. Alternatively, computationally less intensive approaches could be used, such as the Kalman-filter based SORT (Simple and Online Realtime Tracking) \citep{Bewley_2016}. However, all published Kalman-filter based tracking algorithms still require bounding boxes generated by object detection, which would again neutralise the computational advantages of using bottom-up MAPE. We therefore introduce KeySORT (Keypoint Simple and Online Realtime Tracking), a dynamic Kalman filter based tracking algorithm inspired by SORT, but able to track directly (in)complete skeletons constituted of several related keypoints. The advantages of tracking individual keypoints are twofold: (1) no additional bounding box detectors or re-identification steps are needed, and (2) Kalman filtering has a stabilising effect on the keypoint detections, allowing to improve the positional accuracy and consistency of the detected keypoints. Kalman-filter based tracking is able to re-identify individuals on subsequent frames, but is unable to assign an exact identity to an individual by comparison to a reference database. Therefore, to study animal behaviour in an automated way, computer vision based re-identification cannot be eliminated completely. However, by using an optimised Kalman filter based tracking algorithm, the number of computer vision based identifications of individuals can be kept to a minimum. This will result in a major reduction of the computational load required for automated animal behaviour monitoring.
    
    To be able to detect and track the keypoints and skeletons of individual animals from video recordings of a pen or group of animals, we propose a computationally efficient methodology that combines a bottom-up multi-animal pose estimation with an adaptive Kalman filter-based tracking algorithm for keypoints (KeySORT). Moreover, we aim to exploit the properties of Kalman filtering in order to leverage the accuracy and consistency of the pose-estimates, without the necessity to perform extensive hyperparameter tuning. Our methodology can easily be applied for various animal species, but in this paper, we will focus on application of our algorithms to extract spatial and temporal behavioural information for Holstein dairy heifers.
 	
    \section{Methods}
    
    \subsection{Ethics statement}  
     The authors declare that the animal results of the study are reported in accordance with the ARRIVE guidelines. Data were collected at a commercial farm with full consent of the farmers involved. Data collection was carried out according to the Belgian ethical guidelines stating that when there is no interference with the animals, approval of the experiment by an ethical committee is redundant.
    	
    \subsection{Data collection}
    Data were collected on a commercial Belgian dairy farm located in the North-West of Belgium. All animals on the farm are purebred Holstein-Friesian. From December 5, 2021 to February 5, 2022, data were collected in the youngstock barn from a single pen with 8 animals being between 8 and 12 months of age (\cref{fig:visualisation_skeletons}). During the experiment, two groups of eight animals were present in this pen. Of these 16 animals, 11 animals were red-and-white and 5 animals were black-and-white. Video footage was collected continuously during day and night with a roof-mounted 4MP camera (Hikvision, model DS-2CD2643G2-IZS) at 2fps. If the light intensity dropped, the camera automatically switched to night-vision recordings, using built-in infrared illumination. For training purposes, frames were randomly sampled from these recordings to assembly a dataset of 500 frames, equally divided over daylight and night vision frames (\url{https://doi.org/10.5281/zenodo.15016853}). This dataset was split into a training and validation dataset according to a 90/10 ratio. To evaluate the performance of our algorithms, an independent test dataset was constructed by sampling 100 video fragments with a length of 60s (120 frames), equally divided over daylight and night vision recordings (\url{https://doi.org/10.5281/zenodo.15018041}). For each video fragment in the test dataset, keypoint coordinates were annotated in the last frame. For annotation, we used a custom annotation tool. The code for this tool is publicly available via \url{https://github.com/mperneel/Kantool}

    \subsection{Pose-estimation}    
    The algorithm used for multi-animal pose-estimation (MAPE) is extensively elaborated in \cref{sec:appendix_pose_estimation} and contains several important modifications to improve its generalisability and robustness compared to state of the art algorithms used in human \citep{Cao_2021, Cheng_2020} and animal research \citep{Lauer_2022, Pereira_2022, Psota_2019}. The most important modifications are briefly explained below.
    
\begin{figure}[htbp]
    	\centering
    	\includegraphics[width=0.5\textwidth]{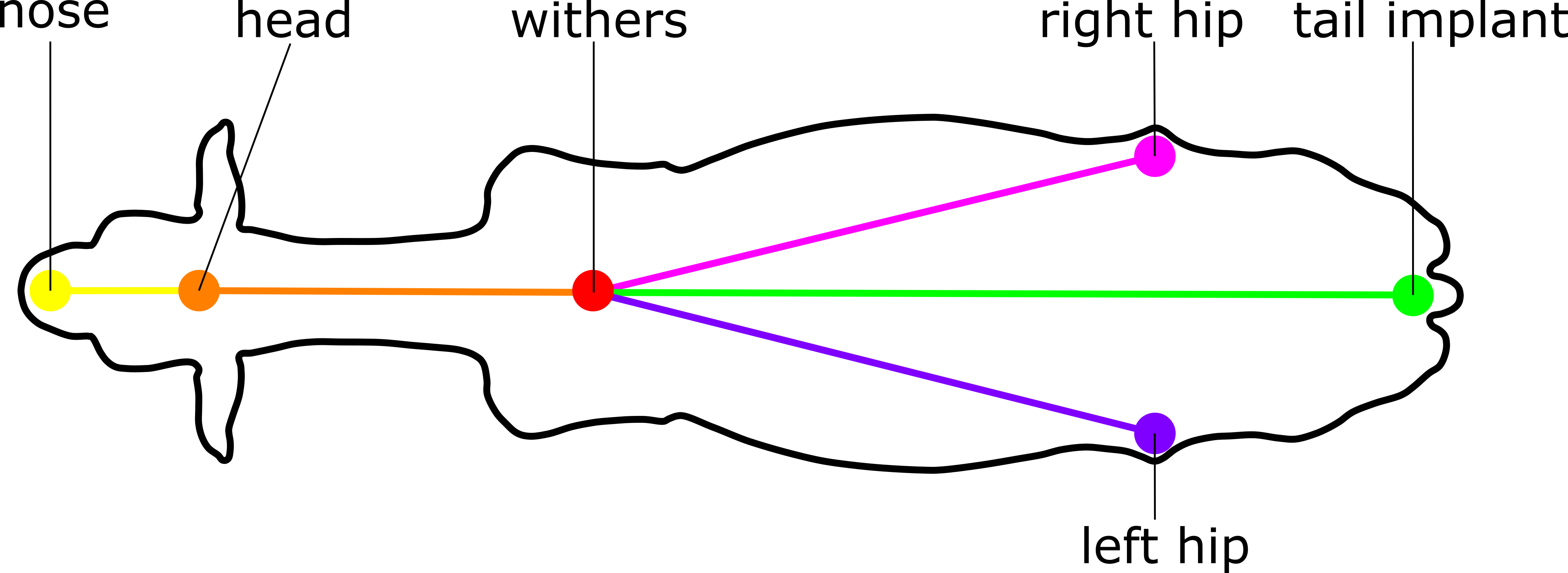}
    	\captionsetup{width=0.5\textwidth}
    	\caption{Overview of the skeleton used for pose-estimation. The withers are the central keypoint, to which all other keypoints are connected in a hierarchic manner. }
    	\label{fig:skeleton}
    \end{figure}

\begin{figure}[htb]	
		\centering
		
		\begin{subfigure}[b]{0.49\textwidth}
			\centering
			\includegraphics[width=\textwidth]{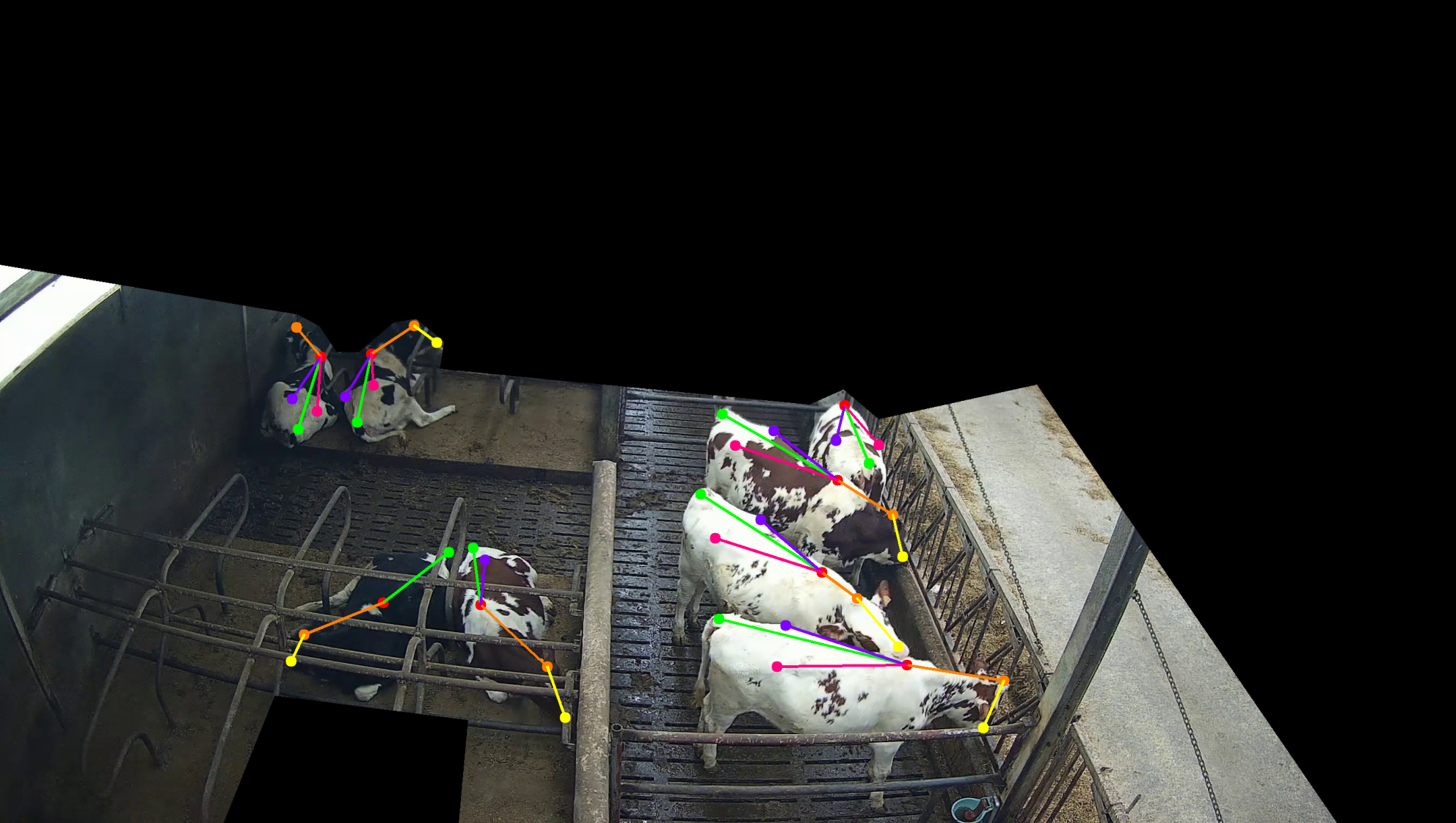}
			\caption{}
			\label{fig:visualisation_skeletons_a}
		\end{subfigure}
		\hfill
		\begin{subfigure}[b]{0.49\textwidth}
			\centering
			\includegraphics[width=\textwidth]{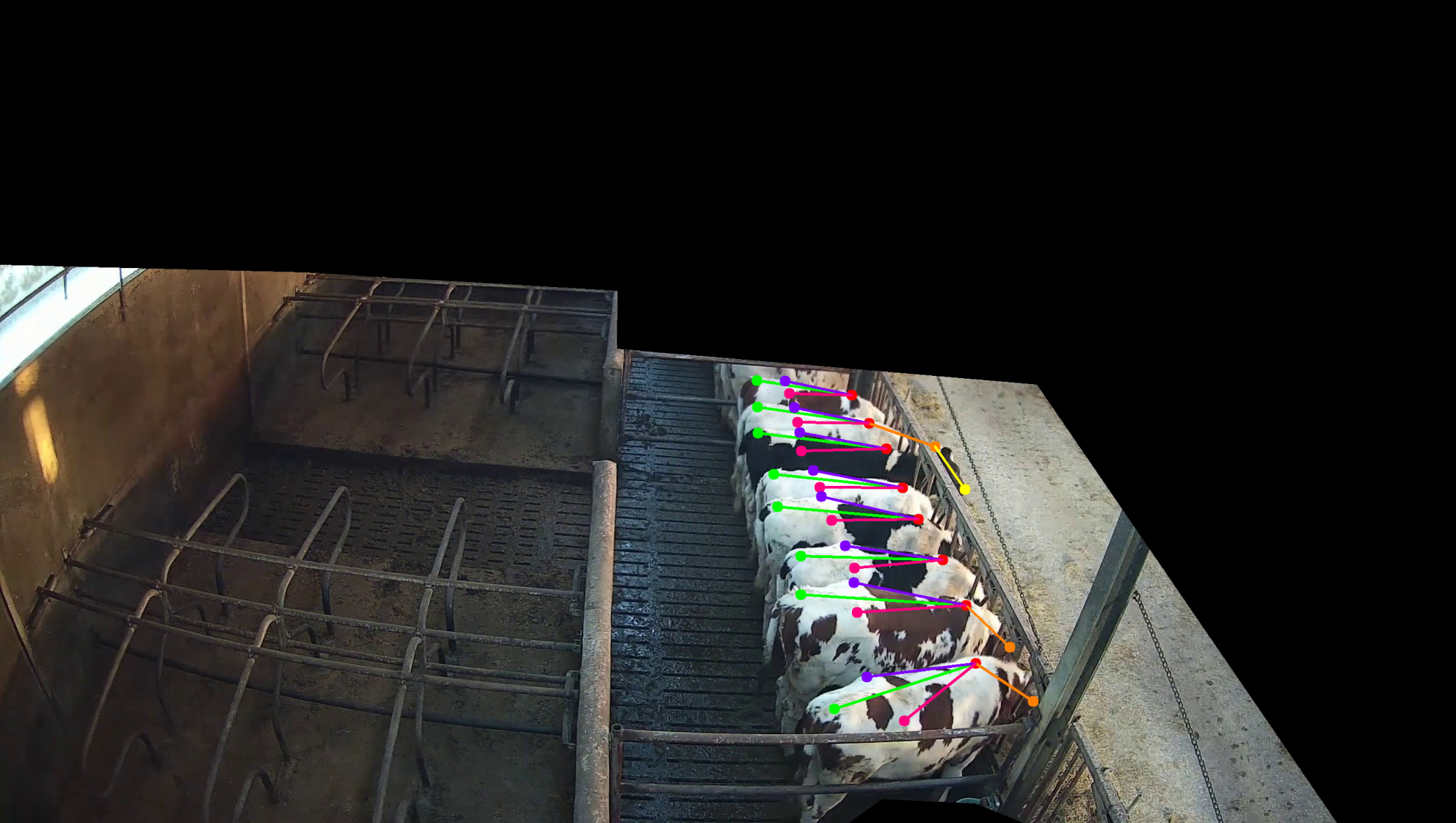}
			\caption{}
			\label{fig:visualisation_skeletons_b}
		\end{subfigure}
		
		\vspace{1em}
		
		\begin{subfigure}[b]{0.49\textwidth}
			\centering
			\includegraphics[width=\textwidth]{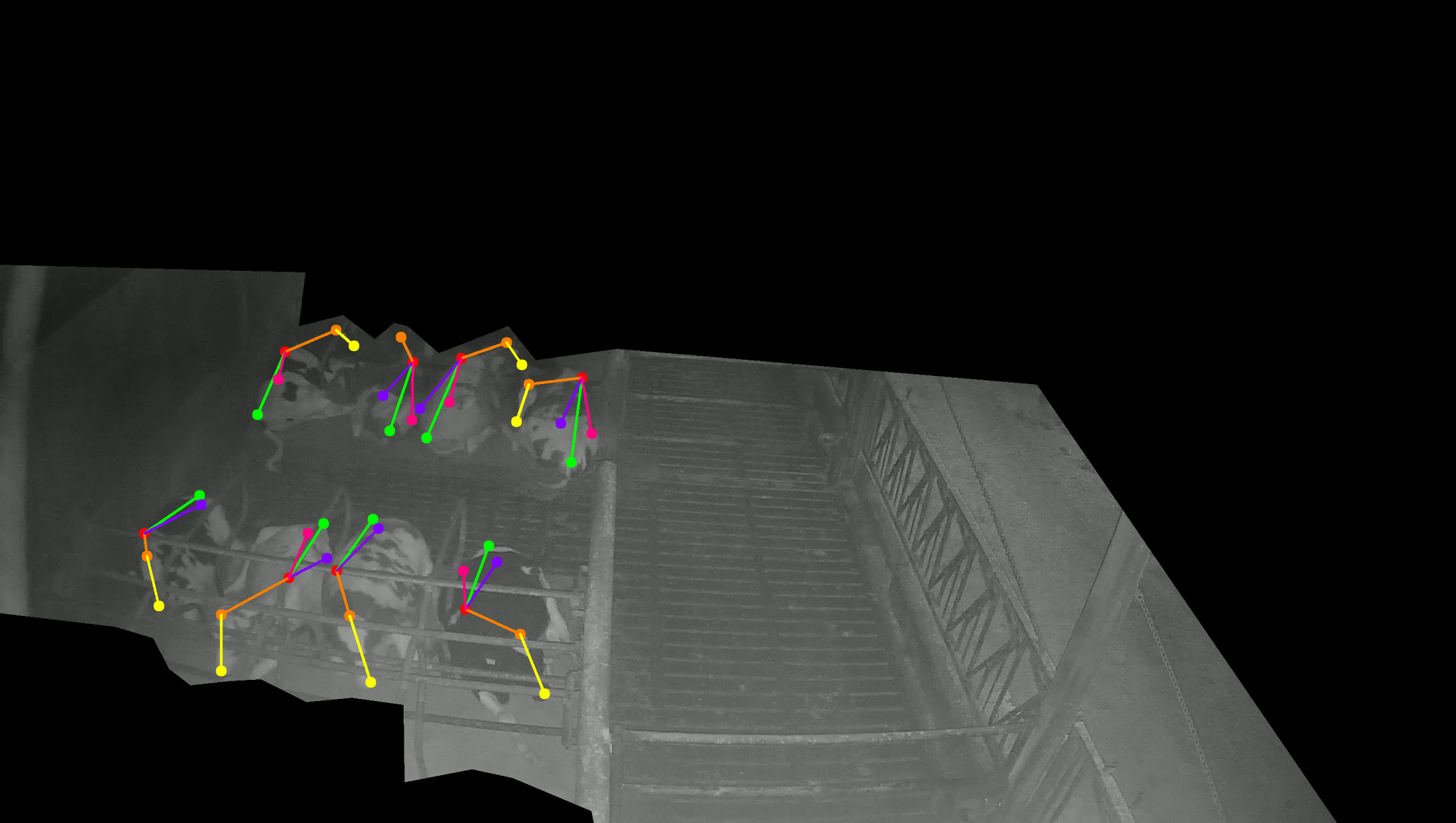}
			\caption{}
			\label{fig:visualisation_skeletons_c}
		\end{subfigure}
		\hfill
		\begin{subfigure}[b]{0.49\textwidth}
			\centering
			\includegraphics[width=\textwidth]{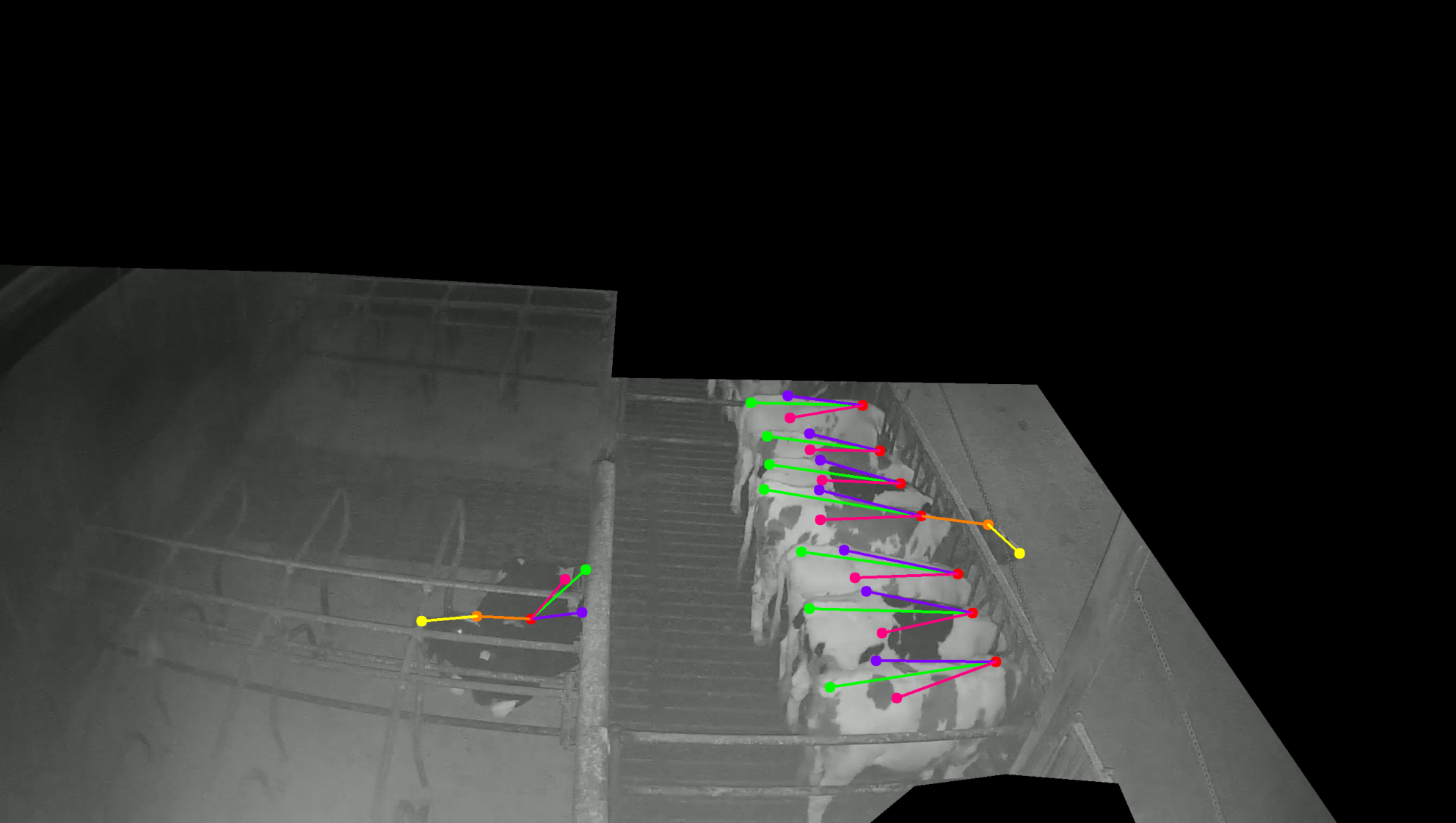}
			\caption{}
			\label{fig:visualisation_skeletons_d}
		\end{subfigure}
		
		\caption{Illustration of ground truth annotations. Recordings were collected both during day and night.}
		\label{fig:visualisation_skeletons}		
	\end{figure}

    First of all, we introduce a framework to formally define an animal skeleton as a hierarchical tree-graph describing the keypoints constituting the skeleton and their mutual connections. Our framework results in skeletons which are easily extendible and can readily be generalised to different animal species. In our study, we use the skeleton illustrated in \cref{fig:skeleton} which focusses on the dorsal axis of the animals and consists of six keypoints: the withers (w), the tail implant (t), the left (lh) and right hip/hook (rh) (tuber coxae of the pelvis), the top of the head (h) and the nose (n). An illustration of the ground truth annotations made according to this skeleton is shown in \cref{fig:visualisation_skeletons}.
    
	Secondly, several important improvements were made to the architecture and training procedure of the hourglass shaped neural network for bottom-up MAPE published by \citep{Psota_2019}. To begin with, all convolutional layers in the neural network were replaced by separable convolutional layers, resulting in a significant reduction of the number of parameters and faster training. Furthermore, the last layer of the neural network was replaced by two single layered branches. The first branch is responsible for the detection of keypoints and has a sigmoid activation function. The second branch has no activation function and is responsible for the prediction of connections between keypoints, which are essential to group keypoints of different categories into skeletons representing individual animals. Thirdly, during training, we specifically target maximisation of the keypoint recovery rate $\eta_k$. The keypoint recovery rate $\eta_k$ quantifies the fraction of ground truth keypoints that is recovered by the pose-estimation pipeline, after post-processing of the neural network output and skeleton assembly. However, $\eta_k$ in itself is non-differentiable and hence not suited to be used as a loss function during training. Therefore, we adopted a multi-loss approach with hyperparameter tuning to maximise the correlation between $\eta_k$ and the actual loss function. Additionally, we applied curriculum learning to maximise knowledge extraction from the training dataset and to obtain a robust algorithm that produces qualitative results, even under challenging circumstances.
	
	Lastly, we deployed a greedy algorithm to associate keypoints with each other, instead of the commonly used Hungarian algorithms. Expecially under circumstances where the animals' position is not random, e.g. when they align next to each other at the feeding fence, the usage of a greedy algorithm improves the robustness of the MAPE to a large extent. 

    \subsection{Pose-tracking}
    \label{sec:pose_tracking}

    To track the skeletons, we introduce KeySORT: keypoint Simple and Online Realtime Tracking. KeySORT is inspired by the SORT algorithm developed by \citep{Bewley_2016} and applies a Kalman filter onto the pose-estimation results in order to track the skeletons. However, in contrast to SORT, which is only able to track bounding boxes, KeySORT can track skeletons directly, without the need of defining a bounding box containing the skeleton. When using a Kalman filter, several matrices have to be designed, of which some also require an initial estimate: the system state $\vec{x}$, the observation vector $\vec{z}$, the transition matrix $\Phi$, the observation matrix $H$, the state covariance matrix $P$, the system noise covariance matrix $Q$ and the observation noise covariance matrix $R$.
    
    The structure of the system state $\vec{x}$ is given by \cref{eq:kalman_state}. It can be observed from \cref{eq:kalman_state} that only the central keypoint (withers) is tracked directly, meaning the corresponding absolute coordinates ($x_w$, $y_w$) and velocities ($\dot{x_w}$, $\dot{y_w}$) are present in the state vector. All the other keypoints are tracked relative to the position of their parent keypoint. Relative tracking in this perspective means that the offset from the parent keypoint ($\Delta^x$,  $\Delta^y$) and the velocity of this offset ($\dot{\Delta}^x$, $\dot{\Delta}^y$) are present in the system state vector $\vec{x}$. By doing so, the Kalman filter implicitly respects the hierarchic nature of the skeleton. Moreover, when keypoints are not detected during several frames, relative tracking of keypoints of rank 1 and higher instead of absolute tracking improves the robustness of the Kalman Filter. For example, when a stationary animal starts moving forward, coinciding with several missing detections of the tail implant, direct tracking of all keypoints would result in fictive lengthening of the spine. This is because the Kalman filter will consider the tail implant still being stationary, while the withers are moving forward. Relative tracking on the contrary, will not suffer from such artefacts since the assumed position of the tail implant will move forward together with the withers' position, even if the former is not detected. The transition matrix $\Phi$ corresponding with $\vec{x}$ is given by \cref{eq:kalman_transition_matrix}, in which $I_{12}$ is a 12-dimensional unit matrix and $0$ is a 12-dimensional square zero matrix.
    
    \begin{equation}
        \label{eq:kalman_state}
        \begin{aligned}
        \vec{x} =& \left[
            x_w, y_w,
            \Delta^x_{w \to t}, \Delta^y_{w \to t},
            \Delta^x_{w \to h}, \Delta^y_{w \to h},
            \Delta^x_{h \to n}, \Delta^y_{h \to n}, \dots
            \right.\\ 
            & \left. 
            \Delta^x_{w \to lh}, \Delta^y_{w \to lh},
            \Delta^x_{w \to rh}, \Delta^y_{w \to rh},
            \dot{x_w}, \dot{y_w},
            \dot{\Delta}^x_{w \to t}, \dot{\Delta}^y_{w \to t},\dots
            \right.\\
            & \left.
            \dot{\Delta}^x_{w \to h}, \dot{\Delta}^y_{w \to h},
            \dot{\Delta}^x_{h \to n}, \dot{\Delta}^y_{h \to n},
            \dot{\Delta}^x_{w \to lh}, \dot{\Delta}^y_{w \to lh},
            \dot{\Delta}^x_{w \to rh}, \dot{\Delta}^y_{w \to rh},
            \right]^T\\
        \end{aligned}
    \end{equation} 

    \begin{equation}
        \label{eq:kalman_transition_matrix}
        \Phi = 
        \begin{bmatrix}
            I_{12} & I_{12} \\
            0_{12} & I_{12}
        \end{bmatrix}       
    \end{equation}    
    
    To facilitate a smooth integration of pose estimation and pose tracking, the observation vector $\vec{z}$, defined by \cref{eq:kalman_observation}, is designed to contain only absolute coordinates. Since $\vec{x}$ contains only relative keypoints, except for the central keypoint, the observation matrix $H$ should consider the hierarchic relationships between keypoints. These hierarchic relationships can be derived directly form the skeleton architecture and are illustrated by \cref{eq:observation_head} for the x-coordinate of the head (rank 1 keypoint) and by \cref{eq:observation_nose} for the x-coordinate of the nose (rank 2 keypoint).

    \begin{equation}
        \label{eq:kalman_observation}
        \begin{aligned}
            \vec{z} = \left[
            x_w, y_w,
            x_t, y_t,
            x_h, y_h,
            x_n, y_n,
            x_{lh}, y_{lh},
            x_{rh}, y_{rh}
            \right]^T\\
        \end{aligned}
    \end{equation}
    \begin{align}
        \label{eq:observation_head}
        x_h =& x_w + \Delta^x_{w \to h}\\		
        \label{eq:observation_nose}
        x_n =& x_w + \Delta^x_{w \to h} + \Delta^x_{h \to n}
    \end{align}

    In order to apply a Kalman filter successfully, correct estimates have to be made for the initial state $\vec{x}_0$, the initial state covariance matrix $P_0$, the measurement noise covariance matrix $R$ and the system noise covariance matrix $Q$. Especially correct estimates for $R$ and $Q$ are crucial to optimise the performance of a Kalman filter. However, when studying animal behaviour, one is often faced with unpredictability and heteroscedasticity of the animals' movements. For example, when walking heifers are compared to lying heifers, the keypoint coordinates of the latter require a significant lower system noise covariance matrix $Q$ in comparison with the former. The presence of heteroscedasticity suggests that using a single universal estimate for $Q$ will result in sub-optimal tracking results. Therefore, we used an improved adaptive Kalman filter (Appendix \ref{sec:adaptive_kalman_filter}) to track animals, which inflates the covariances in the system state covariance matrix $P$ when $Q$ appears to be estimated too low.
    
    An intuitive observation noise covariance matrix $R^*$ was constructed based on the estimated residual variances obtained during evaluation of the pose-estimation accuracy (\S \ref{sec:performance_evaluation}), covariances in $R^*$ are assumed to be zero. $R$ was optimized manually by experimenting with $R^*$, $R^* \times 10^{-1}$, $R^* \times 10^{-2}$, $R^* \times 10^{-3}$ and $R^* \times 10^{-4}$. Since an adaptive Kalman filter is used, we estimated the elements  of the system noise matrix $Q$ deliberately too low. The variances in $Q$ are estimated based on the average observation variance $\bar{\sigma}^2_{R}$, computed as the average of the diagonal elements in $R$. Variances in $Q$ corresponding with position ($x, y, \Delta^x, \Delta^y$) are estimated by $\bar{\sigma}^2_{R} \times 10^{-5}$, whilst variances corresponding to velocities ($\dot{x}, \dot{y}, \dot{\Delta}^x, \dot{\Delta}^y$) are estimated by $\bar{\sigma}^2_{R} \times 10^{-7}$. All covariances in $Q$ are assumed to be zero. For the initial state covariance matrix $P_0$, we used $Q \times 10^{10}$.	

    When new tracklets are initiated, elements in $\vec{x}_0$ corresponding to positions ($x$, $y$, $\Delta^x$ and $\Delta^y$) are estimated based on the first observation using the hierarchic relationships between keypoints, which were illustrated above by \cref{eq:observation_head} and \cref{eq:observation_nose}. Values for $\Delta^x$ and $\Delta^y$ for keypoints missing at the first observation are initialised with zero. Hence, missing keypoints are initialised with the coordinates of their parental keypoint. Since the presence of the central keypoint (withers) is a prerequisite for skeleton validity, the central keypoint $(x_s, y_s)$ is always observed when the Kalman filter is initialised and no measures to handle missing observations of the central keypoint are required. All elements in $\vec{x}_0$ corresponding to velocities ($\dot{x}$, $\dot{y}$, $\dot{\Delta}^x$ and $\dot{\Delta}^y$) are initialised with zero values.
    
    Once tracklets are initiated, the corresponding Kalman filters are updated each time a new pose observation is added to the tracklet. Missing keypoints during this phase are handled by temporarily removing rows from the observation matrix $H$ corresponding to the missing keypoints. In a similar way, rows and columns from the observation noise covariance matrix $R$ which became irrelevant were temporarily removed.

    In order to associate observed skeletons to tracklets, KeySORT uses the average Euclidean distance between observed keypoint coordinates and predicted keypoint coordinates for all observed keypoints, denoted by $\psi(\vec{z}_{obs}, \vec{z}_{pred})$. To associate observed skeletons with the tracklets, $\psi(\vec{z}_{obs}, \vec{z}_{pred})$ is computed for each possible combination of an observation and a tracklet, and pairs are made with a Hungarian algorithm \citep{Kuhn_1955}. To update the tracklets, only pairs with $\psi(\vec{z}_{obs}, \vec{z}_{pred}) \leq$ 25 pixels (on the original image) are retained.

    When applying a Kalman filter to track skeletons, three different sets of coordinates can be distinguished. The first set is given by the observed keypoint coordinates, generated directly by the pose estimation algorithm. The second set is given by the prior coordinates, specifying the predicted keypoint coordinates by the Kalman filter. And thirdly, there are the posterior coordinates, which results from updating the Kalman filter with the observed keypoint coordinates. These posterior coordinates integrate the prior knowledge of the Kalman filter gained from the past with the latest information captured into the observed coordinates. Therefore, using these posterior coordinates allows to reduce the observation noise, resulting in more precise and more consistent pose estimates.   
    
    New tracklets are initiated for each skeleton that could not be associated with a currently existing tracklet. To prevent tracklet initiation from removing true positive skeletons, posterior skeleton coordinates are generated starting immediately from the first observation. Missing observations, however, are only tolerated once tracklets have age 3, corresponding with three consecutive observations after initialisation. Tracklets are terminated when no observation could be associated to the tracklet for more then three consecutive frames.
    
    Kalman filters generate prior coordinate estimates each time step, offering opportunities for imputation of missing (keypoint) detections. However, to prevent the imputation of false positive skeletons, imputation of missing data was performed only at keypoint level, not at skeleton level. In order to be imputed, missing keypoints have to meet two conditions. Firstly, the last detection of that keypoint was less than three frames ago. Missing keypoints are thus imputed for maximum two consecutive frames. Secondly, for each keypoint of a tracklet, an exponential running average observation frequency $f$ was computed, having a memory factor of 0.8 (\cref{eq:running_avg_obs_freq}). Missing keypoints are only imputed if their running average observation frequency $f$ exceeds 0.5.
    
    \begin{equation}
        \label{eq:running_avg_obs_freq}
        f_t = 0.2 * \mathds{1}(\text{keypoint observed}) + 0.8 f_{t-1}
    \end{equation}
    
    \subsection{Performance evaluation}
    \label{sec:performance_evaluation}
    To evaluate the performance of our algorithm for MAPE (Multi Animal Pose Estimation), MAPE was performed for all of the annotated frames from the test dataset and the results were analysed with regard to i) the precision and recall of keypoint detection, ii) the keypoint recovery rate and iii) the relative prediction error. Models were trained on image widths varying between 160 and 1280px to explore the impact of image size on the performance of pose estimation and pose tracking.
     
    Considering the precision and recall, we compared the ground truth keypoint coordinates and the candidate keypoint coordinates before skeleton assembly to the predicted keypoint probability maps and the ground truth keypoint probability maps (\cref{sec:appendix_pose_estimation}), respectively. Ground truth keypoints were considered as being false negative if the corresponding predicted keypoint probability was below 0.5. Analogously, keypoint candidates were considered false positive if the corresponding ground truth keypoint probability was below 0.5. The number of true positive (candidate) keypoints can be determined in two ways. First, from the perspective of the ground truth keypoints, a ground truth keypoint can be considered as true positive if the corresponding predicted keypoint probability is $\geq 0.5$. Secondly, from the perspective of predicted candidate keypoints, a keypoint candidate can be considered as true positive if the corresponding ground truth keypoint probability is $\geq 0.5$. We determined the number of true positive detections in both ways and averaged them to compute the precision and recall.

    After skeleton assembly, predicted skeletons were paired to ground truth skeletons. The distance measure used to pair ground truth skeletons with predicted skeletons is the average Euclidean distance between predicted and ground truth keypoint coordinates for the keypoint categories present in both. Pairs were determined using a Hungarian algorithm with a maximum loss per pair of 50 pixels after rescaling to the original image size. Based on each of these pairs, the keypoint recovery was analysed, quantifying which percentage of ground truth keypoints was also present in the generated skeleton predictions. If a ground truth skeleton was not paired to a predicted skeleton, all of its keypoints were considered as being missed by the algorithm during MAPE. Moreover, the pairs of ground truth skeletons and predicted skeletons were also used to evaluate the prediction error. To account for the scale difference between animals close to the camera and animals further away, all prediction errors were divided by the scale of the corresponding skeleton, which is formally defined in \S\ref{sec:appendix_skeleton} and is computed using \cref{eq:skeleton_scale}.
        
    \begin{equation}
    	\label{eq:skeleton_scale}
    	s_n = \dfrac{1}{\sum_{d \in D}\mathds{1}(\vec{u}_{d,n})}
    	\sum_{d \in D} \beta_d \lVert \vec{u}_{d,n} \rVert
    \end{equation}
	\begin{equation}
		\label{eq:dominant_connections}
		D = \{w \to t, w \to lh, w \to rh\}
	\end{equation}
	\begin{equation}
		\label{eq:connection_weights}
		\begin{aligned}
			\beta_{w \to t} &= 1\\
			\beta_{w \to lh} = \beta_{w \to rh} &= 1.45
		\end{aligned}
	\end{equation}
    \begin{equation}
    	\label{eq:connection}
    	\vec{u}_{a \to b,n} = 
    	\begin{dcases*}
    		\vec{b}_n - \vec{a}_n & if $\vec{a}_n$ and $\vec{b}_n$ exists\\
    		\left[0, 0\right]^\top & otherwise\\
    	\end{dcases*}
    \end{equation}
    \begin{equation}
    	\label{eq:connection_present}
    	\mathds{1}(\vec{u}_{a \to b,n}) = 
    	\begin{dcases*}
    		1 & if $\vec{a}_n$ and $\vec{b}_n$ exists\\
    		0 & otherwise\\
    	\end{dcases*}
    \end{equation}

    To evaluate the performance of KeySORT, all videos from the test dataset were processed and the observed and posterior skeletons coordinates were stored for the last two frames. The observed and posterior skeleton coordinates for the last frame were analysed by computing three performance features: i) keypoint recovery rate, ii) relative prediction error and iii) frame difference compared to previous frame. With regard to the keypoint recovery rate and relative prediction error, the used method is identical to those used to evaluate direct MAPE performance and was described above. To obtain the frame difference between the last and one but last frame, we computed the Euclidean distance between the keypoint coordinates for both frames and this for both the observed and posterior skeleton coordinates, respectively.
    
    \section{Results and discussion}
    \subsection{Direct keypoint detection}
\begin{figure}[htb]
        \centering
        \begin{subfigure}[b]{0.49\textwidth}
            \centering
            \includegraphics[width=\textwidth]{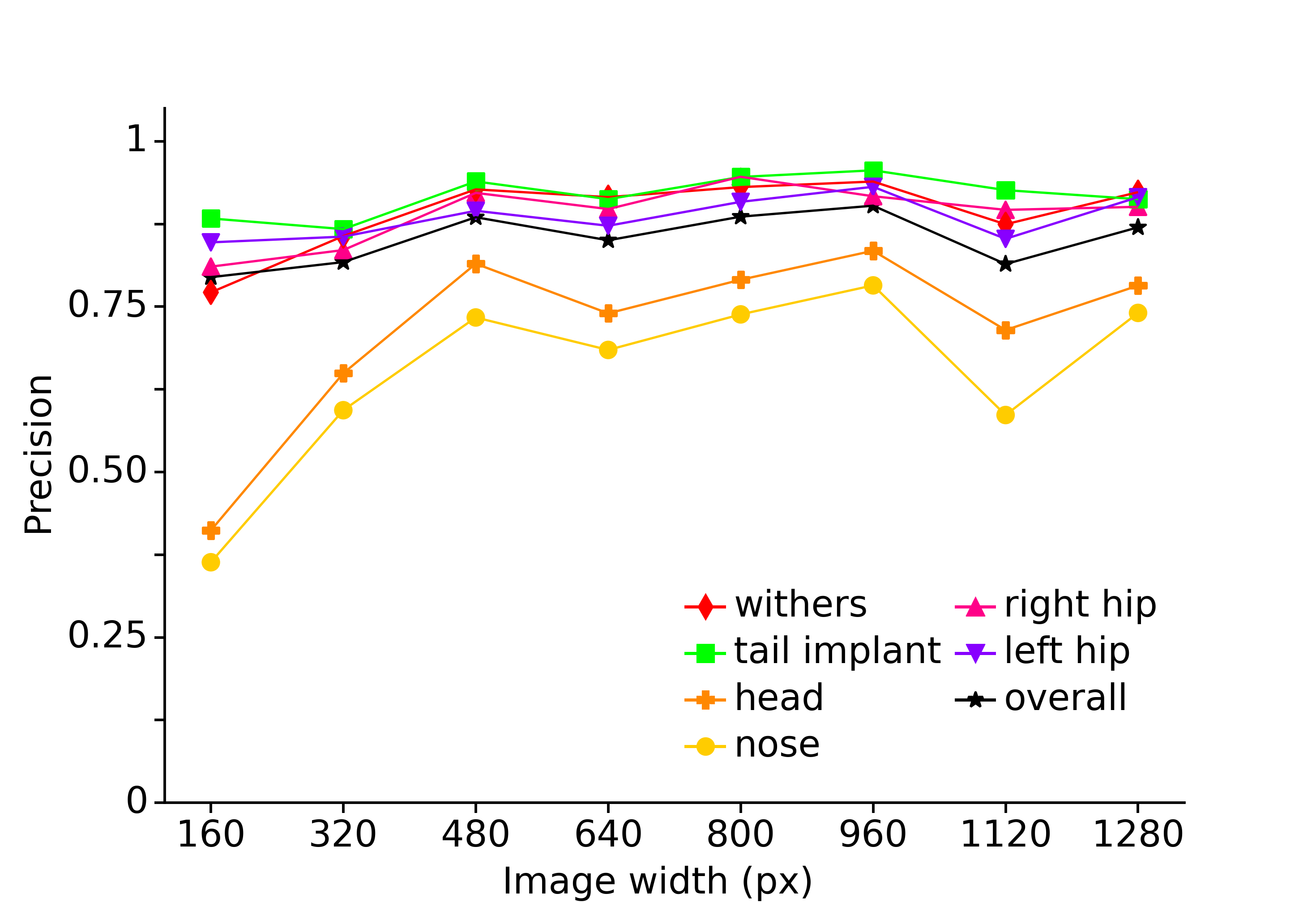}
            \caption{}
            \label{fig:precision_keypoints}
        \end{subfigure}
        \hfill
        \begin{subfigure}[b]{0.49\textwidth}
            \centering
            \includegraphics[width=\textwidth]{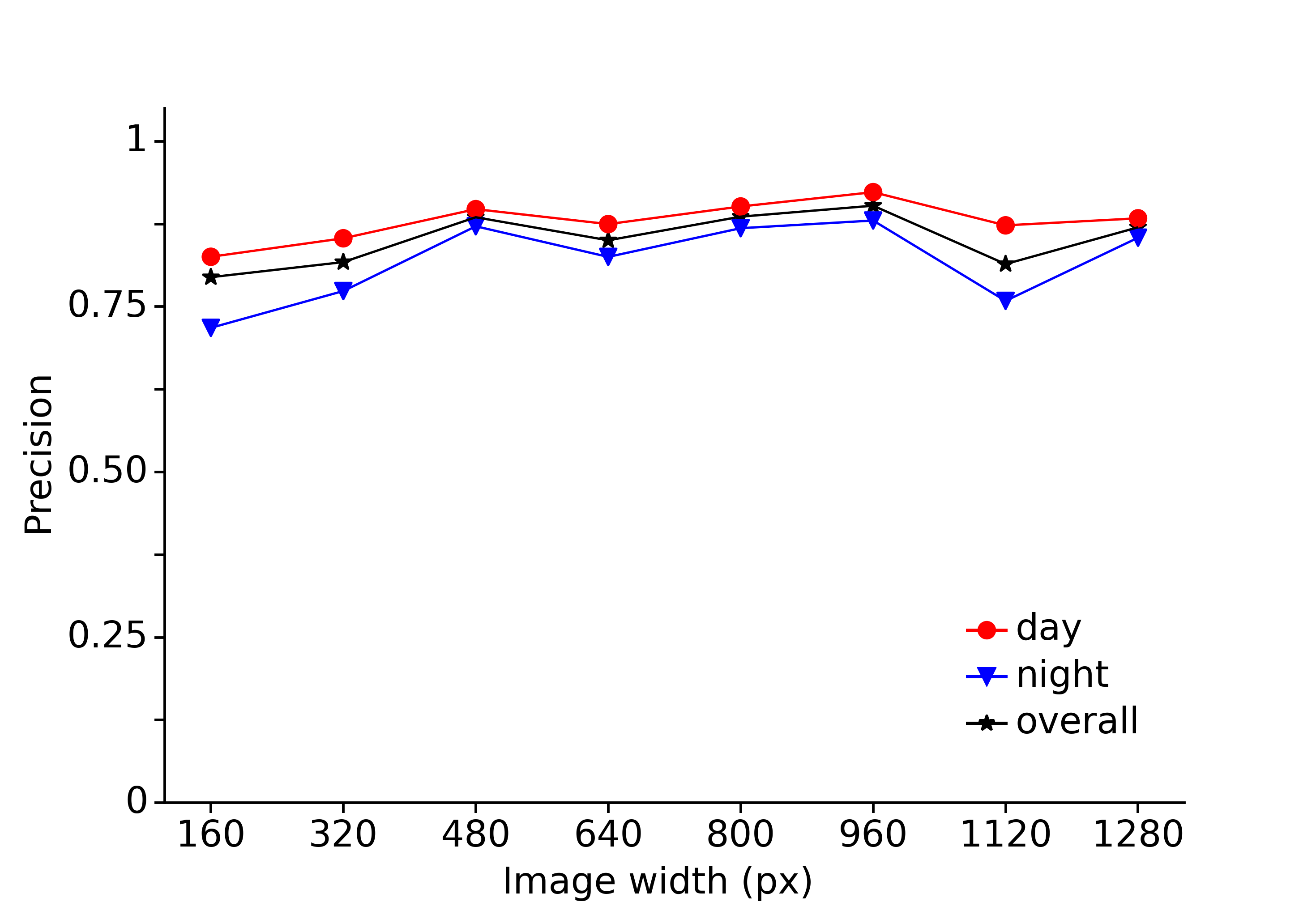}
            \caption{}
            \label{fig:precision_time}
        \end{subfigure}		
        \begin{subfigure}[b]{0.49\textwidth}
            \centering
            \includegraphics[width=\textwidth]{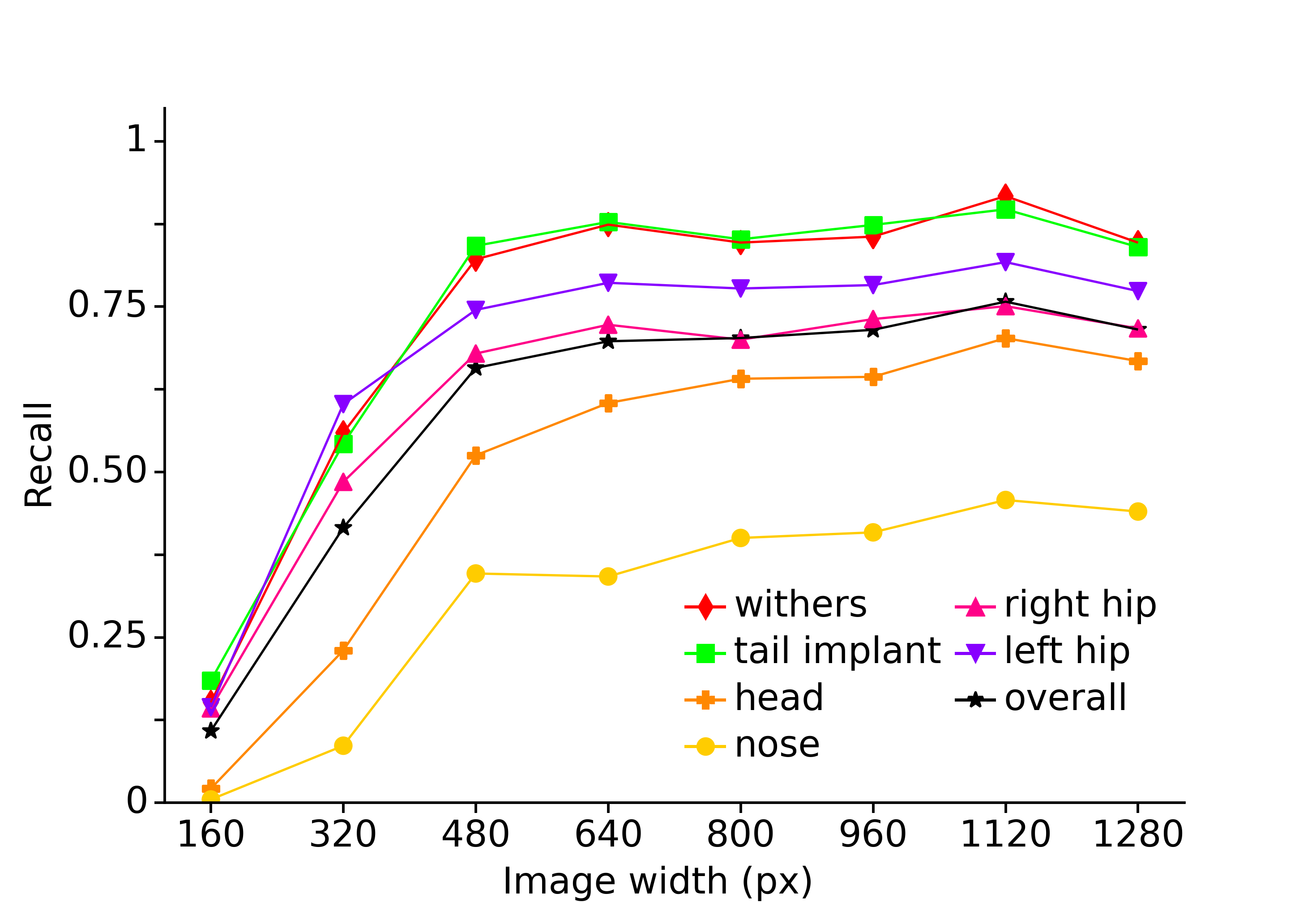}
            \caption{}
            \label{fig:recall_keypoints}
        \end{subfigure}
        \hfill
        \begin{subfigure}[b]{0.49\textwidth}
            \centering
            \includegraphics[width=\textwidth]{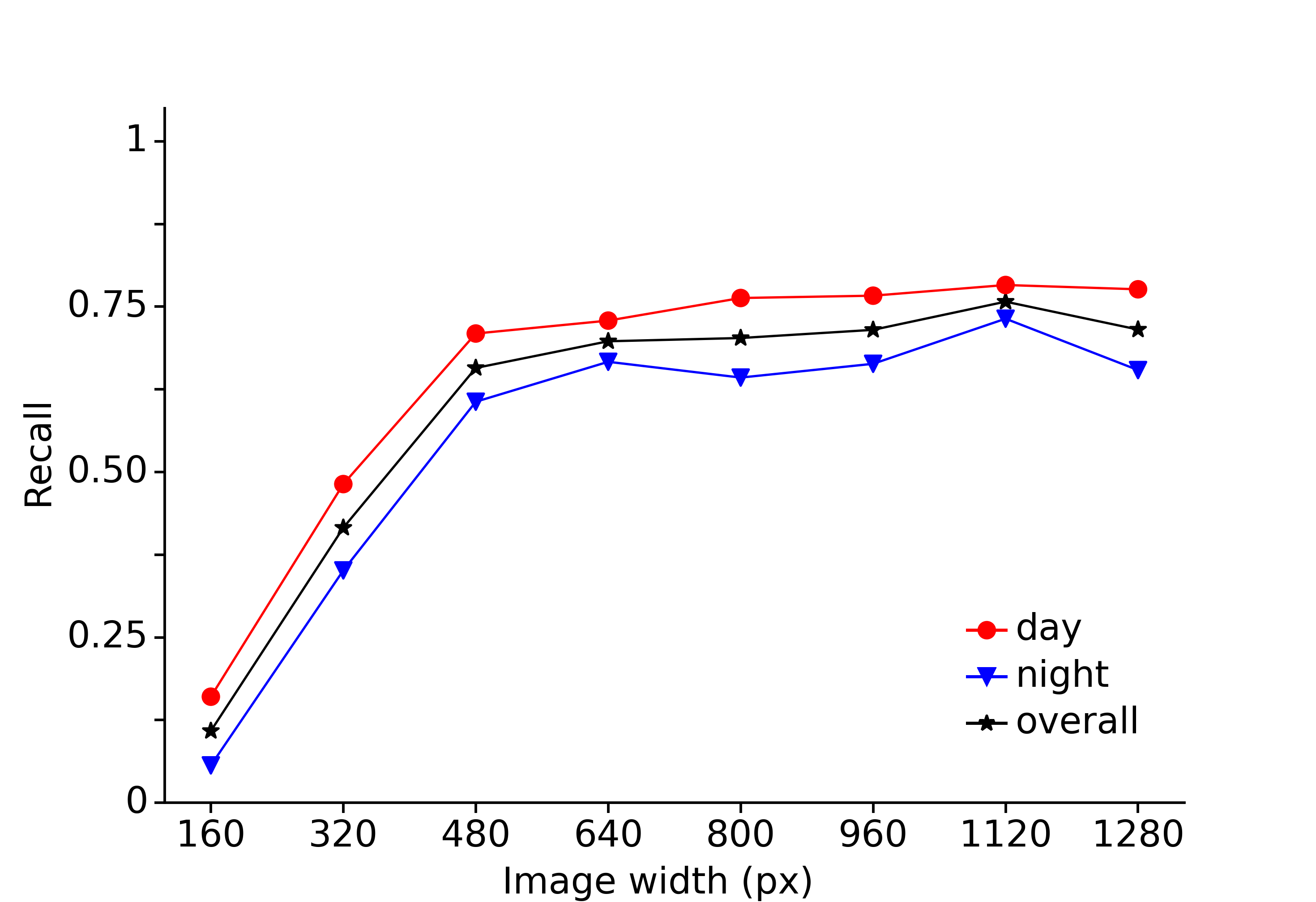}
            \caption{}
            \label{fig:recall_time}
        \end{subfigure}	
    
        \caption{Precision and recall. (a) and (c) show the precision and recall for the different keypoint categories. (b) and (d) compare precision and recall during nightvision and daylight recordings.}	
        \label{fig:precision_recall}	
    \end{figure}

    In \cref{fig:precision_recall}, the precision and recall for the keypoint detection are shown. The absolute values of the precision and recall in these figures should be interpreted with caution, since these are influenced by the kernel size used to train the neural network. If a detected keypoint deviates significantly from its corresponding ground truth keypoint, the probability of it being classified as false positive increases with decreasing kernel size. When a small kernel size was handled, it will probably be classified as false positive, while a large kernel size will allow for a larger margin before it will be considered false positive. However, since using small kernel sizes slows down or even inhibits network training, while too large kernel sizes reduce the resolving power of the neural network, an appropriate choice has to be made with regard to the kernel size, influencing the magnitude of the reported precision and recall. Although the magnitude of the precision and recall measures are thus influenced by the chosen kernel size, trends will be similar across varying kernel sizes.	
    
    When the overall recall is studied, one can observe that at the lowest resolution, the recall is relatively low (0.11). When the resolution increases, the recall quickly rises to 0.66 for 480px wide images. If the image resolution increases more, further improvement of the recall is negligible. This is in contrast with the overall precision, which has already a value of 0.80 at 160 px width, and only slightly increases to 0.89 at 480px width. When the resolution increases even more, no increase of the precision can be observed. At 1120px width, one even can distinguish a slight drop of the precision to 0.81. However, since the models used at different resolutions were trained independently from each other, this is probably due to coincidence rather than an actual decrease of performance at the specified resolution.
    
    The overall precision and recall aggregates the six keypoints of the used skeleton. However, not all keypoints show the same patterns for precision and recall, as can be observed in \cref{fig:precision_keypoints} and \cref{fig:recall_keypoints}. From these figures, it is clear the detection of head and nose is much more difficult in comparison with the detection of withers, tail implant, left- and right hip. Both for the precision and recall, the statistics for the head and nose are significantly worse in comparison with those of the other keypoints. Especially at the resolutions of 160 and 320px, the keypoint detection delivers really poor results for these keypoints. The low precision values at these resolutions are however not caused by a higher number of false positive keypoint detections in comparison with the other keypoints, but are mainly due to an extremely low number of true positive detections as can be deduced from figure \cref{fig:recall_keypoints}. While resolutions over 480px width do not result in a considerable improvement of the overall recall rate, the recall rate of the nose and head show a considerable improvement up to 0.46 and 0.70 respectively at a resolution of 1120px. This indicates that the optimal resolution for detection of these keypoints is higher in comparison with the detection of the other keypoints. However, the recall rates for the head and nose stay low in comparison with those of the withers and tail implant. While the recall rates for the withers and tail implant at a resolution of 1120px are good with values of 0.92 and 0.88 respectively, the recall rate of the top of the head is only fair (0.72) and that of the nose is still quite low (0.59).
    
    When the recall rates for withers and tail implant on the one side and for the right- and left hip on the other side are compared for resolution over 480px, one can see the recall rate for both hips is lower in comparison with the tail implant and the withers. The frequency of not detected hips is thus higher than the frequency of false negatives for the tail implant and withers. A possible explanation for this is the higher variation in appearance for the hips in comparison to the tail implant and withers, especially when a hip is turned away from the camera. In that case, only a slight curvature of the animal's body indicates the location of the hip. Therefore, experience and consistency were essential to produce qualitative annotations for these keypoints. Likely, the model struggles with this variability in appearance in a similar way.
    
    In figures \cref{fig:precision_time} and \cref{fig:recall_time}, the precision and recall are compared between day and night recordings. Since night recordings contain less contrasts and have the tendency to be slightly more blurry, the precision and recall for night recordings were expected to be lower in comparison with those of daylight recordings. Nevertheless, the observed differences in recall and precision are relatively small compared to the influence of the keypoint category: at 480px resolution, the difference in recall rate is only 0.10, while the difference with regard to the precision (0.03) is even smaller.
    
    \begin{figure}[htbp]
        \centering
        \begin{subfigure}[b]{0.49\textwidth}
            \centering
            \includegraphics[width=\textwidth]{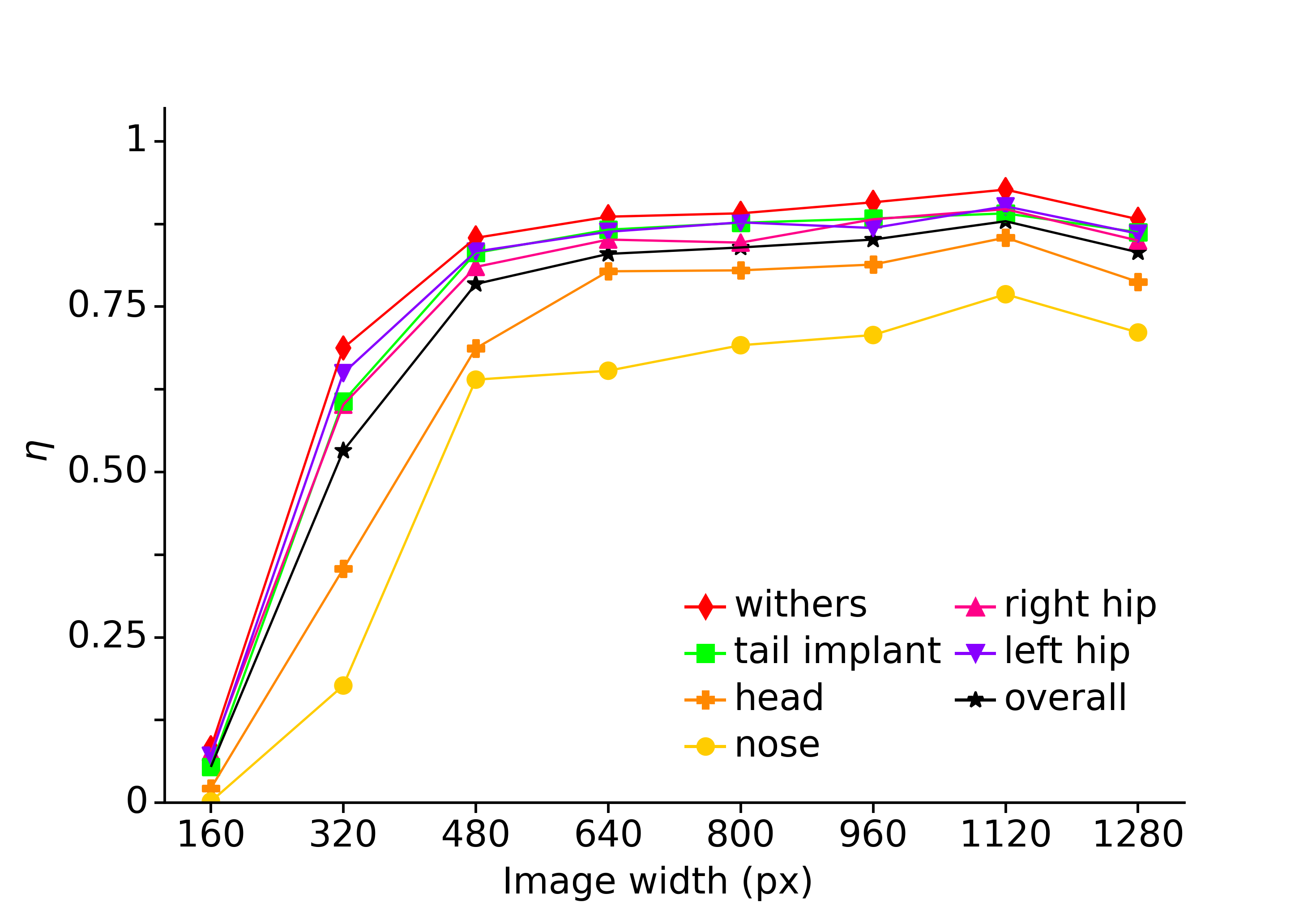}
            \caption{}
            \label{fig:keypoint_eta_keypoints}
        \end{subfigure}
        \hfill
        \begin{subfigure}[b]{0.49\textwidth}
            \centering
            \includegraphics[width=\textwidth]{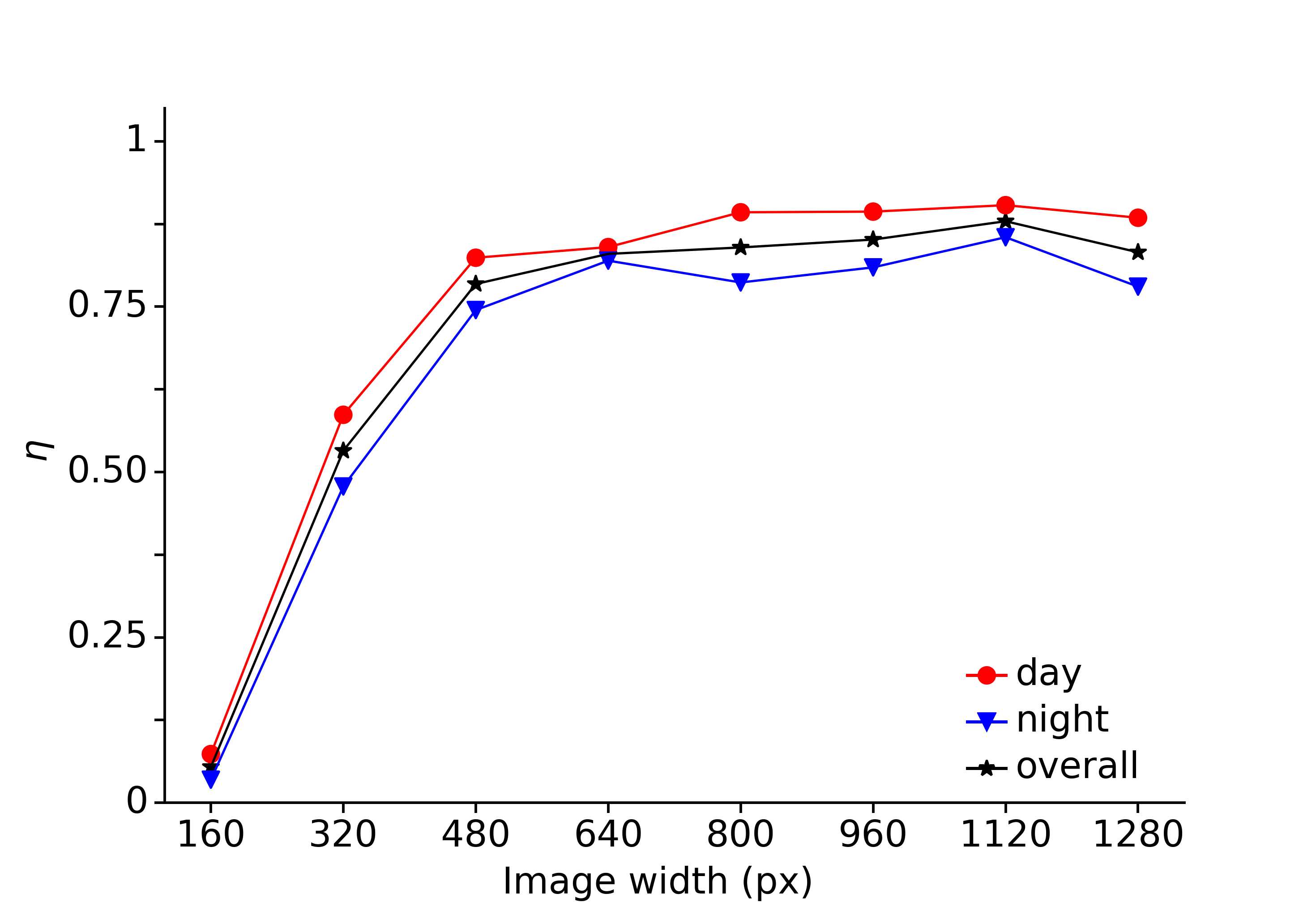}
            \caption{}
            \label{fig:keypoint_eta_time}
        \end{subfigure}
    \caption{keypoint recovery rate.}
    \label{fig:keypoint_eta}	
    \end{figure}
    
    In \cref{fig:keypoint_eta}, the keypoint recovery rate is shown and compared between keypoint categories and between day and night. The keypoint recovery rate quantifies how many ground truth keypoints are found and retained after skeleton assembly. Similar as in figure \cref{fig:precision_recall}, it can be observed that a higher resolution initially results in a higher overall keypoint recovery rate. However, for image widths of 480 px and more, $\eta$ shows saturation, indicating that resolutions over 480px width do not result in better keypoint detections. When the keypoint recovery rate is compared between day and night (\cref{fig:keypoint_eta_time}), the recovery rates during night are lower than for daylight recordings. Nevertheless, similar as in \cref{fig:precision_recall}, the difference in performance is small.
    
    At first glance, the keypoint recovery parameter seems quite similar to the recall (\cref{fig:recall_keypoints} and \cref{fig:recall_time}), however there are two important aspects explaining the differences between \cref{fig:recall_keypoints} and \cref{fig:keypoint_eta_keypoints}. First, skeleton construction respects the hierarchic nature of the used skeleton. Therefore, if only complete ground truth skeletons are considered, the keypoint recovery rate of child keypoints cannot be higher than the one of the parent keypoint. As a consequence, the keypoint recovery rate of the withers (dominant keypoint) is by definition the highest of all keypoints. Secondly, during skeleton construction, keypoint candidates are generated by looking for local maxima on the keypoint probability heatmaps having a probability over 0.4. This is in contrast to the determination of the precision and recall, where a cut-off probability of 0.5 is used. This lower threshold improves the robustness of pose estimation and as a result, the keypoint recovery rate can be higher than the corresponding recall rate. This effect explains the fair recovery rate for the nose (0.64 at 480px), while the corresponding recall rate was only 0.35. Taking this into account, the low recall rate for the nose is probably caused by imprecise detections rather than completely missing of ground truth noses. These imprecise detections result in large patches of low probability on the keypoint heatmaps which yield a keypoint candidate during skeleton construction, but are treated as false negatives during precision-recall analysis.   
    
    \begin{figure}[tb]		
        \centering
        \includegraphics[width=0.6\textwidth]{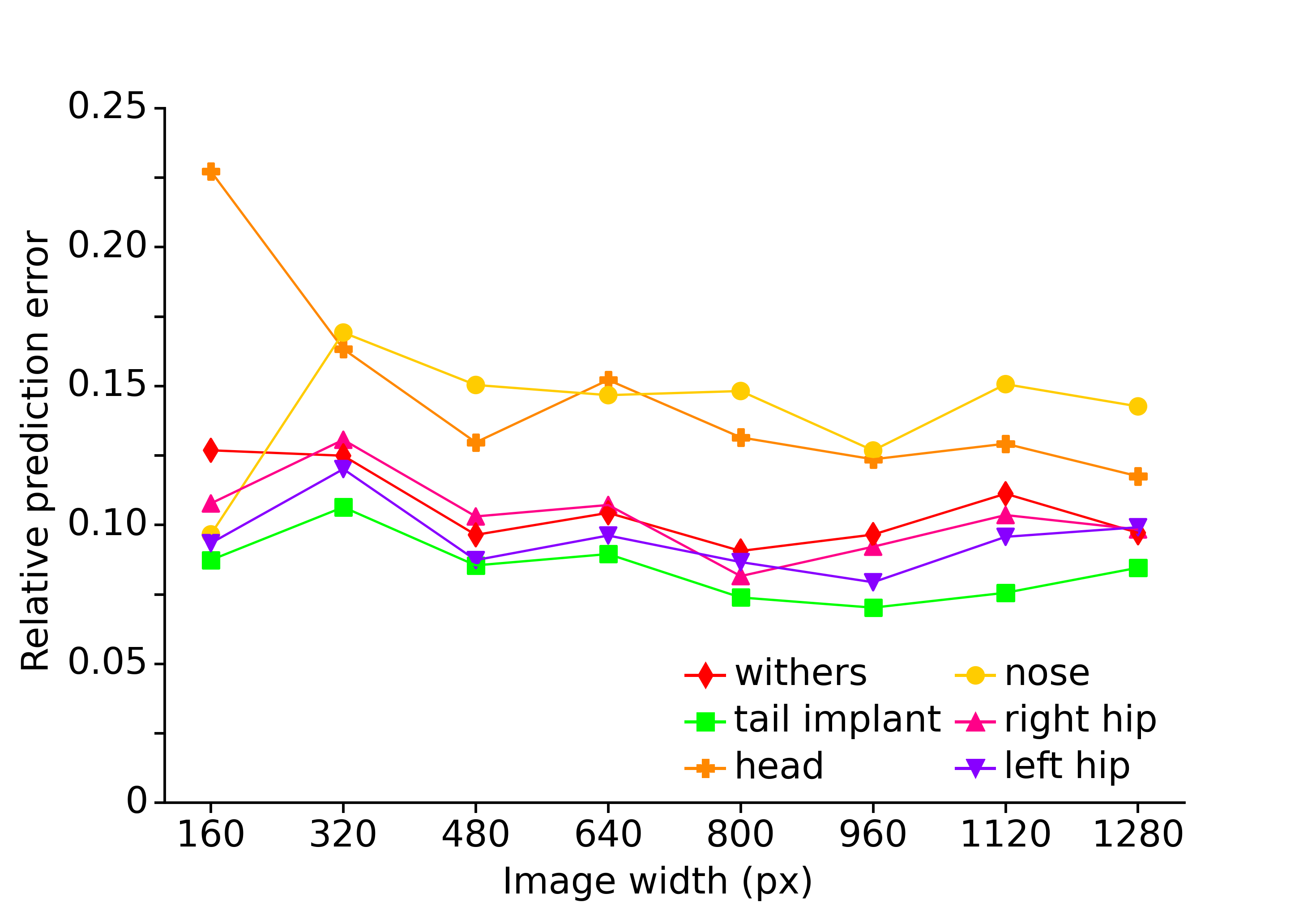}
        \caption{Relative prediction error.}
        \label{fig:relative_error}		
    \end{figure}
    
    This hypothesis is also supported by \cref{fig:relative_error}, which shows the prediction error, expressed relatively to the scale of the skeletons, as defined in \cref{eq:skeleton_scale}. From this figure, it is clear the prediction errors for the nose and head are significantly higher in comparison with those for the other keypoints. At 480px width, when the keypoint recovery rate starts to level off, the relative prediction error for the nose and head are 0.150 and 0.130 respectively, while the relative prediction errors for the other keypoints are all in the interval 0.085 - 0.103. The prediction errors for the head and nose reported at 160px deviate strongly from the errors reported at higher resolutions since these correspond with extremely low keypoint recovery rates: 0.002 for the nose and 0.022 for the head (\cref{fig:keypoint_eta_keypoints}). Due to these low keypoint recovery rates, the estimates for the relative prediction error are inaccurate. Apart from these two inaccurate estimations of the relative prediction error, all prediction errors show a slight decrease up to a resolution of 480px, after which the prediction errors stay relatively constant. Resolutions higher than 480px thus do not result in better performance, which is in concordance with the results reported in \cref{fig:precision_recall} and \cref{fig:keypoint_eta}. Although the prediction errors for the withers, tail implant and hips are quite similar, it is remarkable that the prediction error for the tail implant is always the lowest, while the order of the prediction errors for the withers and hips changes for different resolutions. Possibly, the well-defined nature of the tail implant gives rise to this consistently lower prediction error, whilst not significantly different from the lowest prediction errors reported for the other keypoints (p=0.627 and p=0.099 at 480 and 640 px respectively, t-test).
	
    \subsection{KeySORT}
    As described in \S \ref{sec:pose_tracking}, we finetuned the initial estimate of $R$ by applying various scaling factors to the intuitive observation noise covariance matrix $R^*$ and evaluating the results visually. The best balance between smoothing and responsiveness of the filter was obtained by using $R = R^* \times 10^{-2}$. When $R = R^*$ was used, a strong smoothing of the keypoint coordinates was observed, but the responsiveness of KeySORT to unexpected or abrupt movements was insufficient. This is in contrast to the use of $R = R^* \times 10^{-4}$, when a high responsiveness to unexpected movements was observed, but smoothing of keypoint coordinates was insufficient. The latter was particularly pronounced for the keypoints of lying animals, which continuously vibrated instead of having a stable position over time.
    
    The fact that $R = R^*\times 10^{-2}$ results in the best balance between smoothing and responsiveness, indicates that the variances in $R^*$ are for a large part time-independent. A first explanation for this observation is that subsequent frames are not independent from each other. Therefore, it is expected that the prediction errors from the keypoint detection network for two consecutive frames are correlated. Secondly, it is reasonable that part of the prediction errors reported in \cref{fig:relative_error} does represent the annotation error instead of the prediction error. This could also explain why resolutions higher than 480px-wideness do not result in a lower prediction error.
		
    In \cref{tab:eta_KeySORT}, the results of applying KeySORT are shown with regard to its potential for imputation of missing keypoints. Since the overall keypoint recovery increases with only 1.5\% from 0.784 to 0.799, this potential seems to be limited. However,  when focussing on the two keypoints having the lowest recovery rate, the nose and the head, the improvement of the recovery rate doubles to approximately 3\%. For detecting the nose during daylight recordings, the recovery rate even increases with 6.2\%. As could be expected, the added value of KeySORT with regard to imputation of missing keypoints is thus the largest for keypoints having a lower keypoint recovery rate. The main reason for the modest improvement of the keypoint recovery rate is probably the relatively strict procedure we implemented for imputation of missing keypoints to assure the imputation of missing keypoints would not result in a reduced accuracy of keypoint detection.
	
    \begin{table}[tb]
        \centering
        \begin{tabular}{lcccccc}
            \toprule
            keypoint     & \multicolumn{2}{c}{day} & \multicolumn{2}{c}{night} & \multicolumn{2}{c}{overall} \\
                         & direct &    KeySORT     & direct &     KeySORT      & direct &      KeySORT       \\ \midrule
            withers    &  0883  &     0.888      & 0.825  &      0.825       & 0.854  &       0.857        \\
            tail implant & 0.859  &     0.869      & 0.804  &      0.814       & 0.831  &       0.842        \\
            left hip     & 0.852  &     0.858      & 0.813  &      0.825       & 0.833  &       0.842        \\
            right hip    & 0.859  &     0.862      & 0.761  &      0.780       & 0.810  &       0.821        \\
            head         & 0.743  &     0.773      & 0.634  &      0.659       & 0.687  &       0.715        \\
            nose         & 0.669  &     0.731      & 0.585  &      0.619       & 0.640  &       0.672        \\
            overall      & 0.824  &     0.837      & 0.745  &      0.760       & 0.784  &       0.799        \\ \bottomrule
        \end{tabular}
        \caption{Comparison of the keypoint recovery rate ($\eta$) for direct keypoint detection and keypoint detection with KeySORT for frames analysed at 480px-wide resolution.}		
        \label{tab:eta_KeySORT}
    \end{table}

    A comparison of the relative prediction errors for keypoint detection via direct keypoint detection and via KeySORT is shown in \cref{tab:KeySORT_relative_prediction_error} for frames analysed at a 480px-wide resolution. In this table, it can be noted that for the withers, tail implant and hips, the prediction error shows a minimal increment of 0.1 to 0.2\%. Thus, applying KeySORT does not improve, nor deteriorate the detection accuracy of keypoints for which only a limited number of missing keypoint imputations were performed (\cref{tab:eta_KeySORT}). The prediction error of the head and nose do rise more, by around 1\%, however this increment is not significant ($p=0.183$ and $p=0.273$ for the head and nose respectively, t-test). This prediction error increment is probably caused by the higher number of imputed keypoints, having a higher prediction error. Nonetheless, by applying a relatively strict imputation policy, the resulting increment of the prediction error stays within acceptable limits.	
	
    \begin{table}[htb]
        \centering
        \begin{tabular}{lcc}
            \toprule
            keypoint     &    direct     &    KeySORT    \\ \midrule
            withers    & 0.096 (0.081) & 0.097 (0.084) \\
            tail implant & 0.085 (0.071) & 0.087 (0.073) \\
            left hip     & 0.087 (0.066) & 0.088 (0.071) \\
            right hip    & 0.103 (0.080) & 0.104 (0.075) \\
            head         & 0.130 (0.113) & 0.140 (0.133) \\
            nose         & 0.150 (0.128) & 0.162 (0.142) \\ \bottomrule
        \end{tabular}
        \caption{Relative prediction error for direct keypoint detection and keypoint detection with KeySORT for frames analysed at 480px-wide resolution, the standard deviation of the relative prediction errors is shown between brackets.}
        \label{tab:KeySORT_relative_prediction_error}
    \end{table}

\begin{figure}[tb]		
        \centering
        \includegraphics[width=\textwidth]{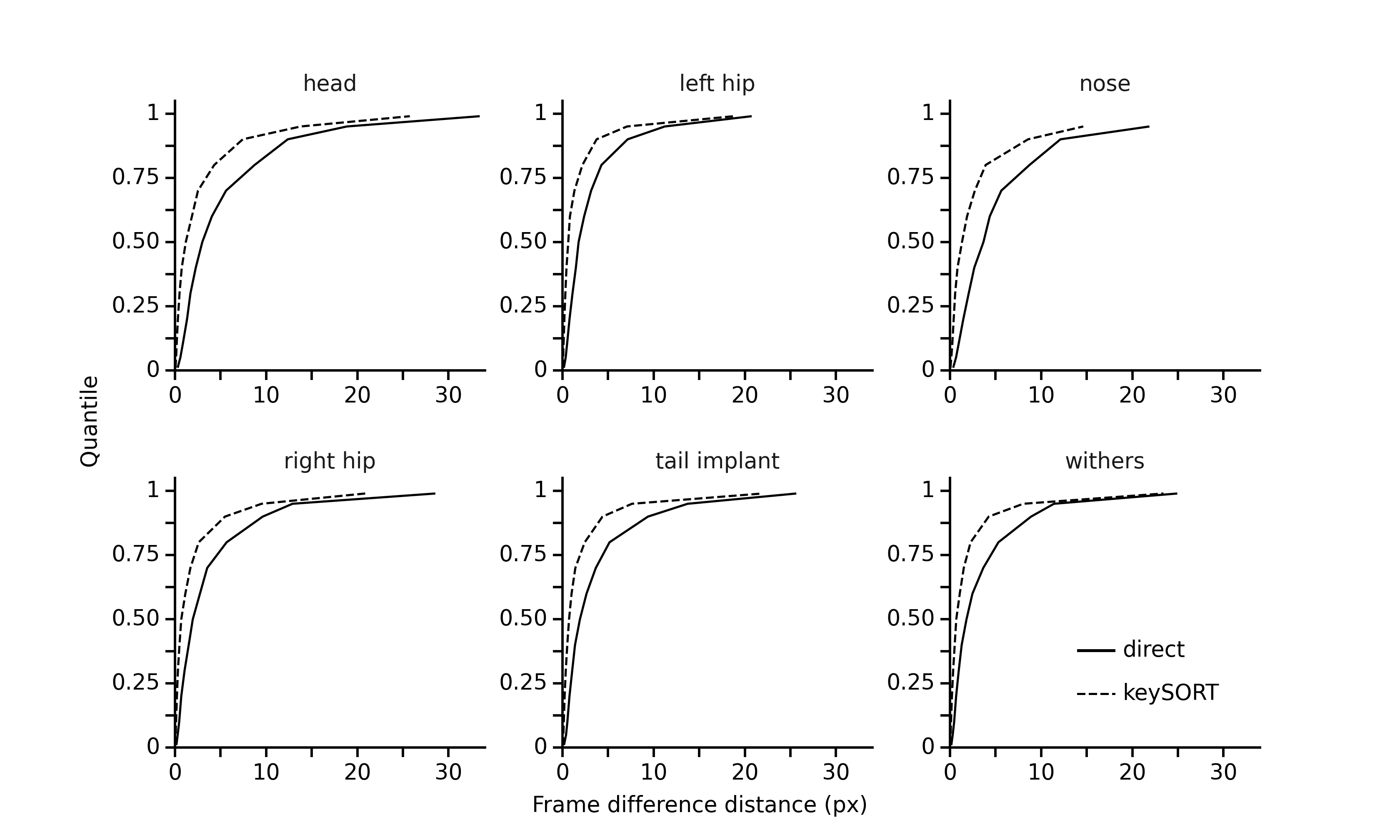}
        \caption{Cumulative distributions for absolute frame differences between keypoint coordinates at the last two frames of video instances, analysed at 480px-wide resolution.}
        \label{fig:frame_differences_quantiles}
    \end{figure}

    The use of KeySORT allows to create tracklets by linking detections on subsequent frames to each other, but shows only a limited added value to improve the keypoint recovery rate and reduce the prediction error. A third aspect is the consistency of keypoint coordinates, which can be studied through quantification of the frame differences of keypoint coordinates of a tracklet. These results are shown in \cref{fig:frame_differences_quantiles} and \cref{tab:quantiles_frame_differences} and illustrate a clear improvement in detection consistency in comparison with direct keypoint detection. In \cref{fig:frame_differences_quantiles}, the cumulative distribution functions for the frame differences clearly have shifted towards lower values and this for all keypoints and all quantiles. The magnitude of the decrease of the quantile values varies hereby when the lower and middle quantiles are compared with the upper quantiles (\cref{tab:quantiles_frame_differences}). For the lower and middle quantiles, most of the KeySORT quantile values are approximately three times lower in comparison with their direct keypoint detection counterpart. For the upper quantiles, the reduction of their values is more modest. The largest absolute decrease for the 95-percent quantile can be observed for the nose, for which the quantile value decreases from 21.87 to 14.62, a decrease with 7.25px. When expressed relatively, however, this decrease is about one third, which is comparable to the decrements of the 95-percent quantiles for the other keypoints.

    \begin{table}[tb]
        \centering
        \begin{tabular}{lcccccc}
            \toprule
            keypoint     & \multicolumn{2}{c}{$Q(0.05)$} & \multicolumn{2}{c}{$Q(0.50)$} & \multicolumn{2}{c}{$Q(0.95)$} \\
            & direct & keySORT              & direct &       keySORT        & direct &       KeySORT        \\
            withers    & 0.30   & 0.07                 &  1.81  &         0.69         & 11.40  &         8.04         \\
            tail implant & 0.39   & 0.09                 &  1.91  &         0.72         & 13.73  &         7.62         \\
            left hip     & 0.35   & 0.07                 &  1.77  &         0.63         & 11.22  &         7.08         \\
            right hip    & 0.30   & 0.08                 &  1.95  &         0.69         & 12.89  &         9.50         \\
            head         & 0.29   & 0.12                 &  3.00  &         1.19         & 18.85  &        13.78         \\
            nose         & 0.35   & 0.14                 &  3.67  &         1.32         & 21.87  &        14.62         \\ \bottomrule
        \end{tabular}
        \caption{5-percent, median and 95-percent quantile for the absolute frame differences between keypoint coordinates at the last two frames of video instances, analysed at 480px-wide resolution.}
        \label{tab:quantiles_frame_differences}
    \end{table}	

    These observations indicate that when animals move relatively little, for example when they are lying, the consistency is largely improved. This improvement is the result of applying the Kalman filter and using the posterior estimates rather than the direct predictions of the keypoint coordinates. When using an ordinary Kalman filter, a highly ameliorated consistency in stable situations, however, results in failure of the Kalman filter in dynamic situations not captured by the transition matrix, for example when an animal abruptly starts moving fast or unpredictable. By using an adaptive Kalman filter, we coped with this drawback. The adaptive Kalman filter is able to react appropriately in these circumstances, resulting in a minor reduction of the largest frame differences. Therefore, the decrement of the upper quantile values is only minor and the standard deviation of the relative prediction error stays approximately the same (\cref{tab:KeySORT_relative_prediction_error}). If the standard deviation of the relative prediction error would have been increased, this would indicate a lag time introduced by the Kalman filter in cases where the movement behaviour of animals changes abruptly, resulting in temporary large relative prediction errors. But since an adaptive kalman filter was used, this is not the case and therefore, we are able to obtain a largely improved detection consistency at virtually no cost with regard to the detection accuracy. In the perspective of behavioural research, this improved consistency results in less noisy data, which will allow to use simpler models to extract behavioural information from keypoint coordinates.
	
    \section{Conclusion}
    In this paper, we presented an improved method for keypoint detection which is easy to generalise to different animal species and requires only a small image resolution. Our method is able to detect up to 80\% of the ground truth keypoints with high accuracy, even when images contain complex scenes of multiple animals close to each other. Moreover, we proposed a theoretically solid adaptive Kalman filter algorithm which can simultaneously cope with both stable and dynamic system behaviour without the need to re-estimate covariance matrices. Based on this adaptive Kalman filter, we developed and validated KeySORT, an algorithm that allows to largely improve the consistency of keypoint detection, which will facilitate automated behaviour monitoring of animals.   
    
     \section{Funding}
    Maarten Perneel received funding from the Flemish Government by FWO (Fonds Wetenschappelijk Onderzoek), grant No. 1131821N. Jan Verwaeren received funding from the Flemish Government under the “Onderzoeksprogramma Artifciële Intelligentie (AI) Vlaanderen”
    
    
    \section{Data availability}
    The data that support the findings of this study are openly available in zenodo at \url{https://doi.org/10.5281/zenodo.15016853} and \url{https://doi.org/10.5281/zenodo.15018041}. 
    
    \section{Competing interests}
    The authors declare that they have no known competing financial interests or personal relationships that could have appeared to influence the work reported in this paper. 
    
    
    \bibliography{papers}

\begin{thebibliography}{10}
\expandafter\ifx\csname url\endcsname\relax
  \def\url#1{\texttt{#1}}\fi
\expandafter\ifx\csname urlprefix\endcsname\relax\def\urlprefix{URL }\fi
\expandafter\ifx\csname href\endcsname\relax
  \def\href#1#2{#2} \def\path#1{#1}\fi

\bibitem{Magrin_2023}
L.~Magrin, B.~Contiero, G.~Cozzi, F.~Gottardo, S.~Segato, Deviation of
  behavioural and productive parameters in dairy cows due to a lameness event:
  a synthesis of reviews, Ital. J. Anim. Sci. 22~(1) (2023) 739--748.
\newblock \href {https://doi.org/10.1080/1828051X.2023.2241870}
  {\path{https://doi.org/10.1080/1828051X.2023.2241870}}.

\bibitem{De_Freslon_2019}
I.~de~Freslon, B.~Mart{\'i}nez-L{\'o}pez, J.~Belkhiria, A.~Strappini, G.~Monti,
  Use of social network analysis to improve the understanding of social
  behaviour in dairy cattle and its impact on disease transmission, Appl. Anim.
  Behav. Sci. 213 (2019) 47--54.
\newblock \href {https://doi.org/10.1016/j.applanim.2019.01.006}
  {\path{https://doi.org/10.1016/j.applanim.2019.01.006}}.

\bibitem{Lodkaew_2023}
T.~Lodkaew, K.~Pasupa, C.~K. Loo, Cowxnet: An automated cow estrus detection
  system, Expert Syst. Appl. 211 (2023) 118550.
\newblock \href {https://doi.org/10.1016/j.eswa.2022.118550}
  {\path{https://doi.org/10.1016/j.eswa.2022.118550}}.

\bibitem{Tsai_2020}
Y.-C. Tsai, J.-T. Hsu, S.-T. Ding, D.~J.~A. Rustia, T.-T. Lin, Assessment of
  dairy cow heat stress by monitoring drinking behaviour using an embedded
  imaging system, Biosystems Eng. 199 (2020) 97--108.
\newblock \href {https://doi.org/10.1016/j.biosystemseng.2020.03.013}
  {\path{https://doi.org/10.1016/j.biosystemseng.2020.03.013}}.

\bibitem{Achour_2020}
B.~Achour, M.~Belkadi, I.~Filali, M.~Laghrouche, M.~Lahdir, Image analysis for
  individual identification and feeding behaviour monitoring of dairy cows
  based on convolutional neural networks (cnn), Biosystems Eng. 198 (2020)
  31--49.
\newblock \href {https://doi.org/10.1016/j.biosystemseng.2020.07.019}
  {\path{https://doi.org/10.1016/j.biosystemseng.2020.07.019}}.

\bibitem{Li_2014}
L.-J. Li, H.~Su, Y.~Lim, L.~Fei-Fei, Object bank: An object-level image
  representation for high-level visual recognition, Int. J. Comput. Vis. 107
  (2014) 20--39.
\newblock \href {https://doi.org/10.1007/s11263-013-0660-x}
  {\path{https://doi.org/10.1007/s11263-013-0660-x}}.

\bibitem{Kumar_2009}
N.~Kumar, A.~C. Berg, P.~N. Belhumeur, S.~K. Nayar, Attribute and simile
  classifiers for face verification, in: 2009 IEEE 12th international
  conference on computer vision, IEEE, 2009, pp. 365--372.
\newblock \href {https://doi.org/10.1109/ICCV.2009.5459250}
  {\path{https://doi.org/10.1109/ICCV.2009.5459250}}.

\bibitem{Song_2014}
F.~Song, X.~Tan, S.~Chen, Exploiting relationship between attributes for
  improved face verification, Comput. Vision Image Understanding 122 (2014)
  143--154.
\newblock \href {https://doi.org/10.1016/j.cviu.2014.02.010}
  {\path{https://doi.org/10.1016/j.cviu.2014.02.010}}.

\bibitem{Lauer_2022}
J.~Lauer, M.~Zhou, S.~Ye, W.~Menegas, S.~Schneider, T.~Nath, M.~M. Rahman,
  V.~Di~Santo, D.~Soberanes, G.~Feng, et~al., Multi-animal pose estimation,
  identification and tracking with deeplabcut, Nat. Methods 19~(4) (2022)
  496--504.
\newblock \href {https://doi.org/10.1038/s41592-022-01443-0}
  {\path{https://doi.org/10.1038/s41592-022-01443-0}}.

\bibitem{Pereira_2022}
T.~D. Pereira, N.~Tabris, A.~Matsliah, D.~M. Turner, J.~Li, S.~Ravindranath,
  E.~S. Papadoyannis, E.~Normand, D.~S. Deutsch, Z.~Y. Wang, et~al., Sleap: A
  deep learning system for multi-animal pose tracking, Nat. Methods 19~(4)
  (2022) 486--495.
\newblock \href {https://doi.org/10.1038/s41592-022-01426-1}
  {\path{https://doi.org/10.1038/s41592-022-01426-1}}.

\bibitem{Fang_2017}
H.-S. Fang, S.~Xie, Y.-W. Tai, C.~Lu, Rmpe: Regional multi-person pose
  estimation, in: Proceedings of the IEEE international conference on computer
  vision, 2017, pp. 2334--2343.

\bibitem{Iqbal_2016}
U.~Iqbal, J.~Gall, Multi-person pose estimation with local joint-to-person
  associations, in: European Conference on Computer Vision, Springer, 2016, pp.
  627--642.
\newblock \href {https://doi.org/10.1007/978-3-319-48881-3_44}
  {\path{https://doi.org/10.1007/978-3-319-48881-3_44}}.

\bibitem{Cao_2021}
Z.~Cao, G.~Hidalgo, T.~Simon, S.-E. Wei, Y.~Sheikh, Openpose: realtime
  multi-person 2d pose estimation using part affinity fields, IEEE transactions
  on pattern analysis and machine intelligence 43~(1) (2021) 172--186.
\newblock \href {https://doi.org/10.1109/TPAMI.2019.2929257}
  {\path{https://doi.org/10.1109/TPAMI.2019.2929257}}.

\bibitem{Psota_2019}
E.~T. Psota, M.~Mittek, L.~C. P{\'e}rez, T.~Schmidt, B.~Mote, Multi-pig part
  detection and association with a fully-convolutional network, Sens. 19~(4)
  (2019) 852.
\newblock \href {https://doi.org/10.3390/s19040852}
  {\path{https://doi.org/10.3390/s19040852}}.

\bibitem{Topham_2023}
L.~K. Topham, W.~Khan, D.~Al-Jumeily, A.~Hussain, Human body pose estimation
  for gait identification: A comprehensive survey of datasets and models, ACM
  computing surveys 55~(6) (2023) 120.
\newblock \href {https://doi.org/10.1145/3533384}
  {\path{https://doi.org/10.1145/3533384}}.

\bibitem{Ren_2021}
K.~Ren, G.~Bernes, M.~Hetta, J.~Karlsson, Tracking and analysing social
  interactions in dairy cattle with real-time locating system and machine
  learning, J Syst Architect 116 (2021) 102139.
\newblock \href {https://doi.org/10.1016/j.sysarc.2021.102139}
  {\path{https://doi.org/10.1016/j.sysarc.2021.102139}}.

\bibitem{Foris_2019}
B.~Foris, M.~Zebunke, J.~Langbein, N.~Melzer, Comprehensive analysis of
  affiliative and agonistic social networks in lactating dairy cattle groups,
  Appl. Anim. Behav. Sci. 210 (2019) 60--67.
\newblock \href {https://doi.org/10.1016/j.applanim.2018.10.016}
  {\path{https://doi.org/10.1016/j.applanim.2018.10.016}}.

\bibitem{Wojke_2017}
N.~Wojke, A.~Bewley, D.~Paulus, Simple online and realtime tracking with a deep
  association metric, in: 2017 IEEE international conference on image
  processing (ICIP), IEEE, 2017, pp. 3645--3649.
\newblock \href {https://doi.org/10.1109/ICIP.2017.8296962}
  {\path{https://doi.org/10.1109/ICIP.2017.8296962}}.

\bibitem{Mahmoudi_2019}
N.~Mahmoudi, S.~M. Ahadi, M.~Rahmati, Multi-target tracking using cnn-based
  features: Cnnmtt, Multimed Tools Appl 78 (2019) 7077--7096.
\newblock \href {https://doi.org/10.1007/s11042-018-6467-6}
  {\path{https://doi.org/10.1007/s11042-018-6467-6}}.

\bibitem{Fang_2018}
K.~Fang, Y.~Xiang, X.~Li, S.~Savarese, Recurrent autoregressive networks for
  online multi-object tracking, in: 2018 IEEE Winter Conference on Applications
  of Computer Vision (WACV), IEEE, 2018, pp. 466--475.
\newblock \href {https://doi.org/10.1109/WACV.2018.00057}
  {\path{https://doi.org/10.1109/WACV.2018.00057}}.

\bibitem{Bewley_2016}
A.~Bewley, Z.~Ge, L.~Ott, F.~Ramos, B.~Upcroft, Simple online and realtime
  tracking, in: 2016 IEEE international conference on image processing (ICIP),
  IEEE, 2016, pp. 3464--3468.
\newblock \href {https://doi.org/10.1109/ICIP.2016.7533003}
  {\path{https://doi.org/10.1109/ICIP.2016.7533003}}.

\bibitem{Cheng_2020}
B.~Cheng, B.~Xiao, J.~Wang, H.~Shi, T.~S. Huang, L.~Zhang, Higherhrnet:
  Scale-aware representation learning for bottom-up human pose estimation, in:
  Proceedings of the IEEE/CVF conference on computer vision and pattern
  recognition, 2020, pp. 5386--5395.

\bibitem{Kuhn_1955}
H.~W. Kuhn, The hungarian method for the assignment problem, Naval research
  logistics quarterly 2~(1-2) (1955) 83--97.

\bibitem{Cao_2017}
Z.~Cao, T.~Simon, S.-E. Wei, Y.~Sheikh, Realtime multi-person 2d pose
  estimation using part affinity fields, in: Proceedings of the IEEE conference
  on computer vision and pattern recognition, 2017, pp. 7291--7299.

\bibitem{Chollet_2017}
F.~Chollet, Xception: Deep learning with depthwise separable convolutions, in:
  Proceedings of the IEEE conference on computer vision and pattern
  recognition, 2017, pp. 1251--1258.

\bibitem{Groenendijk_2021}
R.~Groenendijk, S.~Karaoglu, T.~Gevers, T.~Mensink, Multi-loss weighting with
  coefficient of variations, in: Proceedings of the IEEE/CVF winter conference
  on applications of computer vision, 2021, pp. 1469--1478.

\bibitem{Bengio_2009}
Y.~Bengio, J.~Louradour, R.~Collobert, J.~Weston, Curriculum learning, in:
  Proceedings of the 26th annual international conference on machine learning,
  2009, pp. 41--48.
\newblock \href {https://doi.org/10.1145/1553374.1553380}
  {\path{https://doi.org/10.1145/1553374.1553380}}.

\bibitem{Kingma_2014}
D.~P. Kingma, J.~Ba, Adam: A method for stochastic optimization, arXiv\href
  {https://doi.org/10.48550/arXiv.1412.6980}
  {\path{https://doi.org/10.48550/arXiv.1412.6980}}.

\bibitem{Yang_2006}
Y.~Yang, W.~Gao, An optimal adaptive kalman filter, J. Geod. 80~(4) (2006)
  177--183.
\newblock \href {https://doi.org/10.1007/s00190-006-0041-0}
  {\path{https://doi.org/10.1007/s00190-006-0041-0}}.

\end{thebibliography}
    
    \newpage
    \appendix
    
    \counterwithin{figure}{section}
    \counterwithin{table}{section}
    
    \section{Pose estimation}
\label{sec:appendix_pose_estimation}

Most state of the art algorithms for bottom up multi-individual pose estimation algorithms used in human \citep{Cao_2021, Cheng_2020} and animal research \citep{Lauer_2022, Pereira_2022, Psota_2019} use a deep convolutional neural network to predict simultaneously keypoint locations and keypoint connections. The latter are required to group keypoints into animal skeletons. However, the algorithms used during multiple animal pose estimation (MAPE) often have a lack of generalisability and robustness. Therefore, starting from the methodology published by \citep{Psota_2019}, we propose several significant algorithm modifications, which will be elaborated in this appendix. We start with the introduction of a skeleton as a hierarchical graph of keypoints and their mutual connections. Thereafter, we discuss how keypoints and connections are represented in keypoint heatmaps and part affinity fields generated by the used neural network. After introduction of the neural network architecture and training procedure, we conclude this appendix with the post-processing steps needed to assemble skeletons based on the neural network output.

\subsection{Skeleton}
\label{sec:appendix_skeleton}
During MAPE, animals are modelled according to a predefined skeleton (\cref{fig:skeleton}). A skeleton is a graph describing both the keypoints to be detected and the connections between keypoints that need to be modelled. The skeleton needs to be defined carefully. On the one hand, from a practical perspective, the keypoints constituting the skeleton should be able to describe a wide variety of behaviours. On the other hand, from a computational perspective, the skeleton architecture should facilitate the association of keypoints as much as possible. With regard to the latter, the skeleton is required to have a hierarchical tree-graph architecture to describe the keypoints and their connections. At the root of the tree-graph, a central keypoint (rank 0) is defined. Preferably, this is a keypoint which is almost always visible. Next to the central keypoint, the tree-graph consists of primary keypoints (rank 1) which are directly connected to the central keypoint by a first order edge/connection, secondary keypoints (rank 2) which are only connected to a primary keypoint by a second order edge/connection, etc. In general, the rank of a keypoint is given by the path length to the central keypoint. Mindfull of the hierarchic nature of the skeleton, for a connection of the $k$th order, the constituting keypoint of rank $k-1$ is referred to as the parent keypoint, while the constituting keypoint of rank $k$ is referred to as the child keypoint. Lastly, the rank of the skeleton is defined as the largest path length to the central keypoint occurring in the tree-graph. The assignment of a rank and order to keypoints and edges/connections, respectively, will be used later on, when the construction of skeletons by keypoint association is discussed.

\begin{figure}[htbp]
	\centering
	\includegraphics[width=0.5\textwidth]{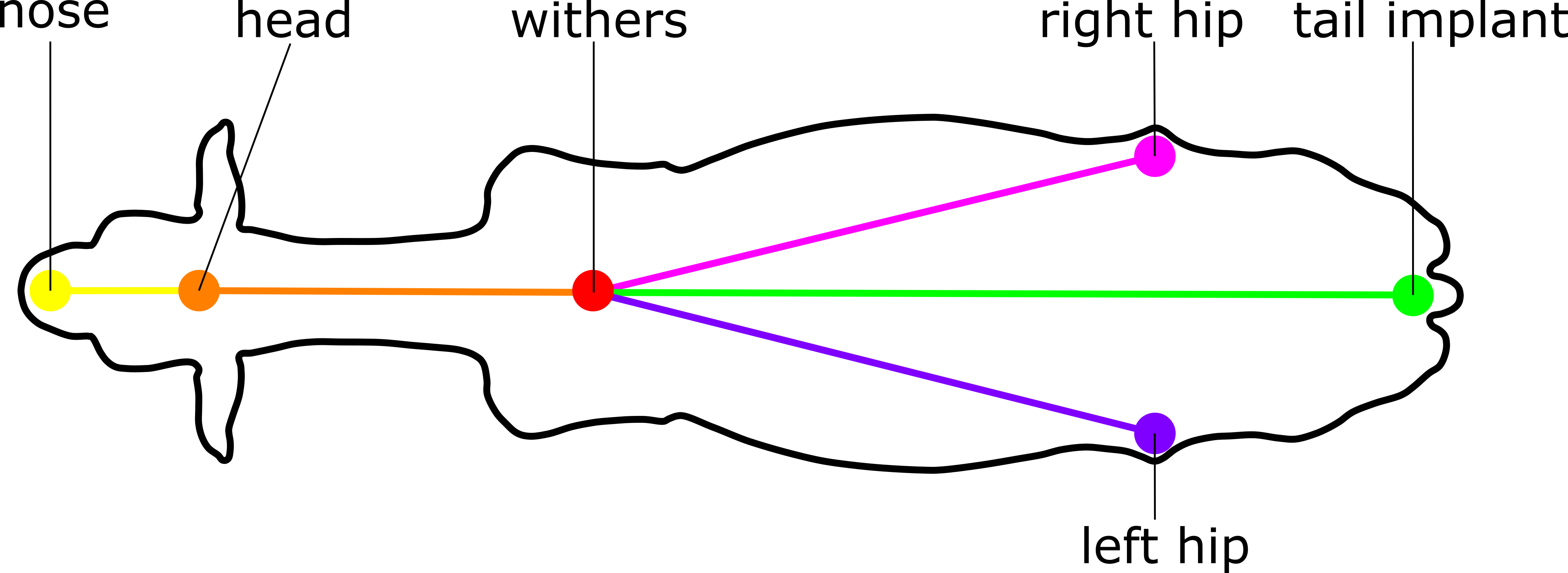}
	\captionsetup{width=0.5\textwidth}
	\caption{Overview of the skeleton used for pose-estimation. The withers are the central keypoint, to which all other keypoints are connected in a hierarchic manner. }
	\label{fig_a:skeleton}
\end{figure}

The hierarchical skeleton that is used in this study has rank 2 and consists of six keypoints (\cref{fig:skeleton}) that are mostly visible from a top/tilted view of a pen: the withers (w), the tail implant (t), the left (lh) and right hook (rh)(tuber coxae of the pelvis), the top of the head (h) and the nose (n). The withers is the central keypoint in the hierarchy since it is the least occluded keypoint and moreover allows to minimise the overall rank of the skeleton. Furthermore, four rank 1 keypoints are present which are directly connected to the central keypoint: head, tail, left and right hip. Lastly, the nose is a rank 2 keypoint, which is connected to the head by a second order connection. The inclusion of both the tail implant and the left and right hook, assures that at least one keypoint of the hindquarters is visible for the camera, independent of the position of the animal. For example, if a heifer is lying on its left side, sometimes the left hook and tail implant are hidden from the camera's perspective, while the right hook will never be occluded by the animal itself and therefore will most likely be visible.

Furthermore, two additional concepts have to be introduced: dominant connections and skeleton scale. The set of dominant connections $D$ is a subset of the first order connections of the skeletons and is used to assess the validity of a skeleton. If none of these dominant connections is present in an assembled skeleton, the skeleton is rejected. In our research, three of the four first order connections are classified as dominant connections: withers to tail ($w \to t$), withers to left hook ($w \to lh$) and withers to right hook ($w \to rh$). The use of three dominant connections is required since none of the keypoints of the hindquarters of the animal is always visible from the camera perspective, as explained above. By using dominant connections, our approach guarantees that minimal skeletons with only a withers keypoint or rudimentary skeletons with only keypoints from the head and forehand of the animal will never be retained during MAPE.

\begin{equation}
	\label{eq_a:dominant_conections}
	D = \{ w \to t, w \to rh, w \to lh \}
\end{equation}

The skeleton scale quantifies the optical size of an animal from the perspective of the camera. Animals closer to the camera appear to be larger in comparison to animals in the back of the recorded area. Therefore, animals closer to the camera will have a larger skeleton scale compared to animals further away. The skeleton scale for animal $n \in \{1, \ldots, N\}$ is obtained via \cref{eq_a:skeleton_scale}, which computes a weighted average with weights $\beta_d$ of the lengths of the dominant connections $\vec{u}_{d,n}$. A connection $\vec{u}_{a \to b,n}$ is defined by \cref{eq_a:connection} as the vector from keypoint $\vec{a}_n=[x^a_n, y^a_n]^\top$ to keypoint $\vec{b}_n=[x^b_n, y^b_n]^\top$, in which both $\vec{a}_n$ and $\vec{b}_n$ are two-dimensional position vectors quantifying the position of a keypoint in image coordinates. If at least one of the constituting keypoints of a connection is missing, $\vec{u}_{d,n}$ is set equal to the two-dimensional zero vector. The indicator function $\mathds{1}(\vec{u}_{d,n})$ in \cref{eq_a:skeleton_scale} is defined by \cref{eq_a:connection_present} and allows to handle missing connections. Since every valid skeleton is required to have at least one dominant connection being present, $s_n$ exists for every valid skeleton.

\begin{equation}
	\label{eq_a:skeleton_scale}
	s_n = \dfrac{1}{\sum_{d \in D}\mathds{1}(\vec{u}_{d,n})}
	\sum_{d \in D} \beta_d \lVert \vec{u}_{d,n} \rVert
\end{equation} 
\begin{equation}
	\label{eq_a:connection}
	\vec{u}_{a \to b,n} = 
	\begin{dcases*}
		\vec{b}_n - \vec{a}_n & if $\vec{a}_n$ and $\vec{b}_n$ exists\\
		\left[0, 0\right]^\top & otherwise\\
	\end{dcases*}
\end{equation}
\begin{equation}
	\label{eq_a:connection_present}
	\mathds{1}(\vec{u}_{a \to b,n}) = 
	\begin{dcases*}
		1 & if $\vec{a}_n$ and $\vec{b}_n$ exists\\
		0 & otherwise\\
	\end{dcases*}
\end{equation}

The coefficients $\beta_d$ quantify the average relationship between the lengths of the different dominant connections. If $M$ dominant connections were defined, $D = \{ a_1 \to b_1, \ldots, a_M \to b_M\}$. One can arbitrarily choose one of the dominant connections as reference dominant connection and set its weight to one. So, if dominant connection $k \in D$ is selected as reference dominant connection, $\beta_k = 1$. The other coefficients $\beta_d$ with  $d \in D \setminus \{k\}$ are thereafter given by the regression coefficients of the $M-1$ linear regression models $\lVert \vec{u}_{k,i} \rVert = \beta_d \lVert \vec{u}_{d,i} \rVert + \epsilon_i$, only taking into account skeletons $i \in \{1, \ldots, N\}$ for which both $\vec{u}_{k,i}$ and $\vec{u}_{d,i}$ exist and with $\epsilon_i$ being an error term. When only a single dominant connection is defined, as implemented by \citep{Psota_2019}, \cref{eq_a:skeleton_scale} can be simplified to a large extend and the skeleton scale becomes equal to the length of the dominant connection.

In this study, we defined three dominant connections, as specified by \cref{eq_a:dominant_conections}. The connection between the withers and the tail implant ($w \to t$) functions as reference dominant connection. To account for the symmetry of an animal's skeleton, the weights for the connections $w \to lh$ and $w \to rh$ were computed independently and thereafter averaged in order to obtain the same weight ($\beta_{w \to lh} = \beta_{w \to rh} = 1.45$) for both connections in \cref{eq_a:skeleton_scale}.

\subsection{Keypoint representation}
\label{sec:keypoint_representation}
To encode the keypoint locations in the training data for an image with width $W$ and height $H$, target probability heatmaps $[0,1]^{H\times W}$ are constructed for each keypoint category in the skeleton. On these probability heatmaps, a keypoint of category $k$ of animal $n \in \{1, \dots, N\}$ is represented by a (discretised) isotropic bivariate Gaussian kernel centred around $(y_{k,n}, x_{k,n})$ with standard deviation $\sigma_n$, rescaled such that is has a value of one at its centre. $\sigma_n$ is dependent on the skeleton scale according to \cref{eq_a:mean scale} and \cref{eq_a:kernel_sigma}, in which $s_n$ is the scale of the skeleton of animal $n$ as defined by \cref{eq_a:skeleton_scale}. Note that in image coordinates, $x$ refers to column coordinates and $y$ to row coordinates. To speed up computations, the domain of the kernels is restricted to $\{x \in [x_{k,n} - 3 \sigma_n, x_{k,n} + 3 \sigma_n], y \in [y_{k,n} + 3 \sigma_n, y_{k,n} - 3 \sigma_n]\}$. This procedure is applied to each keypoint of category $k$ to construct the final output map for keypoint category $k$. If the kernel domains of two neighbouring keypoints show overlap, the maximum values of the respective kernels are retained in the zone of overlap.

\begin{equation}
	\label{eq_a:mean scale} 
	\bar{s} = \dfrac{1}{N} \sum_{n=1}^{N} s_n
\end{equation}
\begin{equation}
	\label{eq_a:kernel_sigma} 
	\sigma_n = \theta \dfrac{s_n + \bar{s}}{2}
\end{equation}

Relating the kernel size to the skeleton scale is important to assure keypoint probability heatmaps do not become cluttered. If animals are located further away from the camera, the optical distance between neighbouring animals becomes smaller and hence the $\sigma_n$ should also become smaller in order to prevent too much overlap between neighbouring kernels. To assure robustness, also the mean scale $\bar{s}$ of all the skeletons in the image is considered in \cref{eq_a:kernel_sigma}, such that $\sigma_n$ does not become too small when an animal shows some deviating pose. For example, when animals are grooming their back, $s_n$ can be significantly smaller in comparison to when the animal is standing in a more common way. With regard to the parameter $\theta$ in \cref{eq_a:kernel_sigma}, preliminary experiments showed that model performance is optimal for scaling factor $\theta \in \left[0.15, 0.25\right]$. Too large values for $\theta$ make it difficult for the model to distinguish nearby keypoints from each other. On the other hand, too small values for $\theta$, slow down training considerably and increase the risk of double detections, where a single keypoint is detected twice by the algorithm. For our study, we chose $\theta = 0.20$.   

\begin{figure}[htb]		
	\centering
	\includegraphics[width=\textwidth]{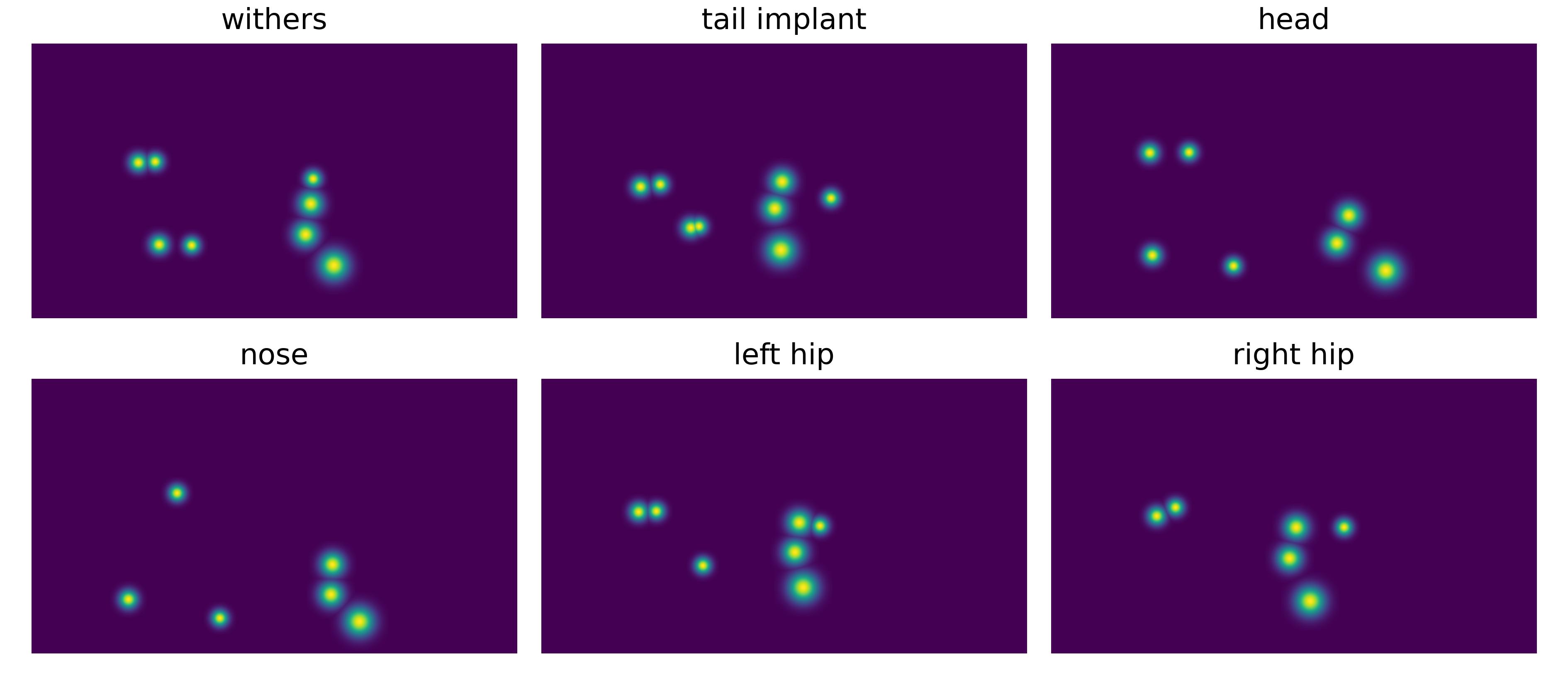}
	\caption{Illustration of ground truth keypoint probability maps. The shown heatmaps correspond with \cref{fig:visualisation_skeletons_a}, remark that keypoint kernels corresponding with animals in the foreground are larger in comparison with those corresponding with animals in the background.}
	\label{fig_a:visualisation_heatmaps}		
\end{figure}

Applied to our pose-model, representing all of the keypoint categories results in 6 probability maps, which will correspond with the first six channels in the neural network output. The probability maps for \cref{fig:visualisation_skeletons_a} are shown in \cref{fig_a:visualisation_heatmaps}.

\begin{figure}[htb]
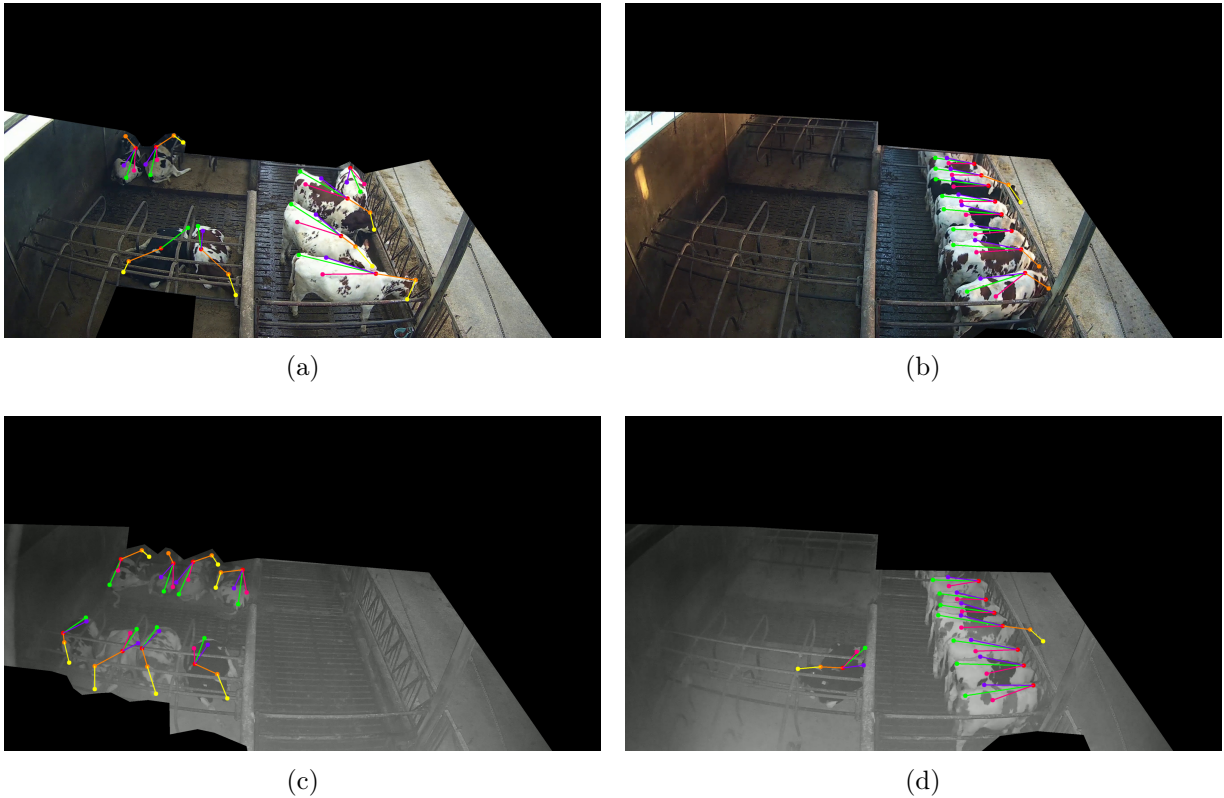
	
	\centering
	
	\begin{subfigure}[b]{0.49\textwidth}
		\centering
		\includegraphics[width=\textwidth]{visualisation_skeletons_00021}
		\caption{}
		\label{fig_a:visualisation_skeletons_a}
	\end{subfigure}
	\hfill
	\begin{subfigure}[b]{0.49\textwidth}
		\centering
		\includegraphics[width=\textwidth]{visualisation_skeletons_00295}
		\caption{}
		\label{fig_a:visualisation_skeletons_b}
	\end{subfigure}
	
	\vspace{1em}
	
	\begin{subfigure}[b]{0.49\textwidth}
		\centering
		\includegraphics[width=\textwidth]{visualisation_skeletons_00002}
		\caption{}
		\label{fig_a:visualisation_skeletons_c}
	\end{subfigure}
	\hfill
	\begin{subfigure}[b]{0.49\textwidth}
		\centering
		\includegraphics[width=\textwidth]{visualisation_skeletons_00069}
		\caption{}
		\label{fig_a:visualisation_skeletons_d}
	\end{subfigure}
	
	\caption{Illustration of ground truth annotations. Recordings were collected both during day and night.}
	\label{fig_a:visualisation_skeletons}		
\end{figure}

\subsection{Connection representation}
Considering a connection between keypoints of type $a$ and $b$, for instance between the withers ($a$) and the tail implant ($b$) (\cref{fig:skeleton}), four different offsets can be considered, each one giving rise to a keypoint association map in the neural network output. The first two keypoint association maps quantify the offset from keypoints $a$ to their paired keypoints $b$ along the x-axis ($\Delta^x_{a \to b}$) and y-axis ($\Delta^y_{a \to b}$), respectively. The last two keypoint association maps quantify in turn the offset from keypoints $b$ to their paired keypoints $a$, again along the x-axis ($\Delta^x_{b \to a}$) and y-axis ($\Delta^y_{b \to a}$), respectively.

Considering the connection between keypoints of type $a$ and $b$, let $\vec{a}_n=[x^a_n, y^a_n]^\top$ and $\vec{b}_n=[x^b_n, y^b_n]^\top$ be the image coordinates of the type $a$ and $b$ keypoint of animal $n \in \{1, \ldots, N\}$, with $N$ being the total number of animals in a frame. The value of the keypoint association map $\Delta^x_{a \to b}$  at $(y,x)$ is obtained via

\begin{equation}
	\label{eq_a:keypoint_association_map}
	\Delta^x_{a \to b}(y,x) = 
	\begin{cases}
		\dfrac{\sum_{n=1}^N \alpha_n^a(y,x) \, (x^b_i - x^a_i)}
		{\sum_{n=1}^N \alpha_n^a(x,y)}
		& \text{if} \; {\sum_{n=1}^N \alpha_n^a(y,x)} > 0\\
		0 & \text{otherwise}\\
	\end{cases}        
\end{equation}

\begin{equation}
	\label{eq_a:kernel_weight_keypoint_association_map}
	\alpha_n^a(y,x) = 
	{\cal N}((x, y); (x^a_n, y^a_n), \sigma^2)
	\mathds{1}\left[{\cal N}((x, y); (x^a_n, y^a_n), \sigma^2) > \gamma \right]
\end{equation}

A similar formula is used to construct $\Delta^y_{a \to b}$, $\Delta^x_{b \to a}$ and $\Delta^y_{b \to a}$. Remark that, in contrast to the part affinity fields used by \citep{Cao_2017}, the clauses in \cref{eq_a:keypoint_association_map} ensure that nonzero values are only assigned to the keypoint association maps conditional to the presence of a keypoint in the neighbourhood. Therefore, keypoint association maps $\Delta^x_{a \to b}$ and $\Delta^y_{a \to b}$ only take on nonzero values in the neighbourhood of a keypoint of type $a$, while for $\Delta^x_{b \to a}$ and $\Delta^y_{b \to a}$ nonzero values can only be observed in the neighbourhood of keypoints of category $b$. The size of this neighbourhood is defined by the parameter $\gamma$ in \cref{eq_a:kernel_weight_keypoint_association_map}, which was set to 0.2.

The skeleton used in our work contains six keypoints. As a result, five connections are present (\cref{fig:skeleton}), resulting in 20 keypoint association maps. However, during preliminary experiments, we observed the neural network had difficulties to decide if a detected hook was a left or a right hook. This because a left hook is on the image often close to the right hook of a neighbouring animal and the other way around, resulting in overlapping kernels on the keypoint probability maps. The previous is especially pronounced when several animals are simultaneously at the feeding fence. Therefore, also the connection between the right and left hook ($rh \to lh$) was considered while constructing the neural network, requiring another four channels in the neural network output. By including the $rh \to lh$ connection, the network is forced to take into account the position of the right hook to determine the position of the corresponding left hook, and vice versa. Inclusion of the $rh \to lh$ connection improved the quality of the respective keypoint probability maps, as well as the quality of the relevant keypoint association maps. The $rh \to lh$ connection is only introduced for training purposes and is not used during post-processing of the neural network output.

\subsection{Neural network architecture and loss function}
The used neural network architecture is illustrated in \cref{fig:network_architecture} and contains several important modifications in comparison with the network architecture used by \citep{Psota_2019}. The first modification entails the replacement of all convolutional layers by separable convolutional layers \citep{Chollet_2017}. Separable convolutional layers have fewer parameters, resulting in faster training. Moreover, separable convolutional layers generally result in a higher model performance \citep{Chollet_2017}, as has appeared to be also the case in our work.    

Secondly, the final layer of the neural network was modified. \citep{Psota_2019}'s network has no activation function. However, this approach has a negative impact on the keypoint probability heatmaps. Since these contain probabilities, they should contain values $\in [0,1]$. In the absence of a sigmoid activation function, it is difficult for the neural network to respect these value boundaries, resulting in slower training and sub-optimal keypoint detection performance. Therefore, instead of using a single final layer, two branches are proposed, each with a single layer. The separable convolutional layer of the first branch has a sigmoid activation function to facilitate the prediction of the six keypoint probability maps. The separable convolutional layer of the second branch, on the other hand, has no activation function and is used to predict the 24 keypoint-association maps.




	
	\definecolor{inputcolor}{RGB}{255, 64, 0}
	\definecolor{convcolor}{RGB}{255, 191, 0}
	\definecolor{maxpoolcolor}{RGB}{0, 64, 255}
	\definecolor{concatcolor}{RGB}{0, 191, 255}
	
	\tikzset{
		pics/box/.style 
		args={
			#1 and #2 and #3 and #4}{
			code={
				\draw[black,ultra thin,fill=#1]  (0,0,0) coordinate(-front-bottom-left) 
				to ++(0,#3,0) coordinate(-front-top-right) 
				-- ++(#2,0,0) coordinate(-front-top-left) 
				-- ++(0,-#3,0) coordinate(-front-bottom-right) 
				-- cycle;
				
				
				\draw[black,ultra thin,fill=#1] (0,#3,0)  
				-- ++(0,0,#4) coordinate(-front-top-left) 
				-- ++(#2,0,0) coordinate(-front-top-right) 
				-- ++(0,0,-#4)  
				-- cycle;
				
				\draw[black,ultra thin,fill=#1!80!black] (#2,0,0) 
				-- ++(0,0,#4) coordinate(-back-bottom-right)
				-- ++(0,#3,0) 
				-- ++(0,0,-#4) 
				-- cycle;
			}
		}
	}
	
	\tikzset{
		pics/concatbox/.style 
		args={
			#1 and #2 and #3 and #4}{
			code={
				\draw[black,ultra thin,fill=#1]  (0,0,0) coordinate(-front-bottom-left) 
				to ++(0,#3,0) coordinate(-front-top-right) 
				-- ++(#2,0,0) coordinate(-front-top-left) 
				-- ++(0,-#3,0) coordinate(-front-bottom-right) 
				-- cycle;
				
				
				\draw[black,ultra thin,fill=#1] (0,#3,0)  
				-- ++(0,0,#4) coordinate(-front-top-left) 
				-- ++(#2,0,0) coordinate(-front-top-right) 
				-- ++(0,0,-#4)  
				-- cycle;
				
				\draw[black,ultra thin,fill=#1!80!black] (#2,0,0) 
				-- ++(0,0,#4) coordinate(-back-bottom-right)
				-- ++(0,#3,0) 
				-- ++(0,0,-#4) 
				-- cycle;
				
				\draw[black,ultra thin, densely dashed] (#2/2,0,0) 
				-- ++(0,#3,0) 
				-- ++(0,0,#4);
			}
		}
	}
	
	\tikzset{
		pics/arrow/.style 
		args={
			#1 and #2 and #3 and #4}{
			code={
				\draw[<-, black, #4] 
				(0,0,0)
				-- ++(0,#3,0)
				-- ++(#1,0,0)
				-- ++(0,-#3 + #2,0);
				
			}
		}
	}

	\begin{figure}[htb]
		\center
		
		\begin{tikzpicture}[x={(1,0)},y={(0,1)},z={({cos(60)},{sin(60)})},
			font=\sffamily\tiny,scale=1.05]
			%
			
			
			\draw pic at (5.42,0.71/2,0.71/4) [transform shape]{arrow= -0.45 and -0.71/2 and 0.1 and densely dashed};
			
			\draw pic at (7.04, 1/2, 1/4) [transform shape]{arrow= -3.81 - 0.05 and -1/2 and 0.1 and densely dashed};
			
			\draw pic at (7.53, 1/2, 1/4) [transform shape]{arrow= -4.63 and -1/2 and 0.2 and solid};
			
			\draw pic at (8.69,1.41/2,1.41/4) [transform shape]{arrow= -6.74 - 0.05 and -1.41/2 and 0.1 and densely dashed};
			
			\draw pic at (9.05,1.41/2,1.41/4) [transform shape]{arrow= -7.35 and -1.41/2 and 0.2 and solid};
			
			\draw pic at (9.93,2/2,2/4) [transform shape]{arrow= -8.94 - 0.05 and -2/2 and 0.1 and densely dashed};
			
			\draw pic at (10.20,2/2,2/4) [transform shape]{arrow= -9.40 and -2/2 and 0.2 and solid};
			
			\draw pic at (10.74,2.83/2,2.83/4) [transform shape]{arrow= -10.30 - 0.07 and -2.83/2 and 0.1 and densely dashed};
			
			\draw pic at (10.94,2.83/2,2.83/4) [transform shape]{arrow= -10.65 and -2.83/2 and 0.2 and solid};

			
			
			

			\foreach \dim in {2.83} {
				\draw pic at (0,-\dim/2,-\dim/4) [transform shape] {box=inputcolor and 0.04 and {\dim} and {\dim/2}};
				\node[anchor=east, rotate=90] at (-0.03,-\dim/2 -0.20, -\dim/4) {input $(3,960,960)$};
				
				\draw pic at (0.09,-\dim/2,-\dim/4) [transform shape] {box=convcolor and 0.10 and {\dim} and {\dim/2}};
				\node[anchor=east, rotate=90] at (0.14,-\dim/2 -0.20, -\dim/4) {conv + relu $(16,960,960)$};
				
				\draw pic at (0.24,-\dim/2,-\dim/4) [transform shape] {box=convcolor and 0.10 and {\dim} and {\dim/2}};
				\node[anchor=east, rotate=90] at (0.33,-\dim/2 -0.20, -\dim/4) {conv + relu + in + drop $(16,960,960)$};
			}
			
			\foreach \dim in {2} {
				\draw pic at (0.39,-\dim/2,-\dim/4) [transform shape] {box=maxpoolcolor and 0.10 and {\dim} and {\dim/2}};
				\node[anchor=east, rotate=90] at (0.44,-\dim/2 - 0.20, -\dim/4) {maxpool $(16,480,480)$};
				
				\draw pic at (0.54,-\dim/2,-\dim/4) [transform shape] {box=convcolor and 0.14 and {\dim} and {\dim/2}};
				\node[anchor=east, rotate=90] at (0.61,-\dim/2-0.20, -\dim/4) {conv + relu $(32,480,480)$};
				
				\draw pic at (0.73,-\dim/2,-\dim/4) [transform shape] {box=convcolor and 0.14 and {\dim} and {\dim/2}};
				\node[anchor=east, rotate=90] at (0.80,-\dim/2-0.20, -\dim/4) {conv + relu + in + drop $(32,480,480)$};
			}
			
			\foreach \dim in {1.41} {
				\draw pic at (0.92,-\dim/2,-\dim/4) [transform shape] {box=maxpoolcolor and 0.14 and {\dim} and {\dim/2}};
				\node[anchor=east, rotate=90] at (0.99,-\dim/2 - 0.20, -\dim/4) {maxpool $(32,240,240)$};
				
				\draw pic at (1.10,-\dim/2,-\dim/4) [transform shape] {box=convcolor and 0.20 and {\dim} and {\dim/2}};
				\node[anchor=east, rotate=90] at (1.20,-\dim/2-0.20, -\dim/4) {conv + relu $(64,240,240)$};
				
				\draw pic at (1.35,-\dim/2,-\dim/4) [transform shape] {box=convcolor and 0.20 and {\dim} and {\dim/2}};
				\node[anchor=east, rotate=90] at (1.45,-\dim/2-0.20, -\dim/4) {conv + relu $(64,240,240)$};
				
				\draw pic at (1.60,-\dim/2,-\dim/4) [transform shape] {box=convcolor and 0.20 and {\dim} and {\dim/2}};
				\node[anchor=east, rotate=90] at (1.70,-\dim/2-0.20, -\dim/4) {conv + relu + in + drop $(64,240,240)$};
			}
			
			\foreach \dim in {1} {
				\draw pic at (1.85,-\dim/2,-\dim/4) [transform shape] {box=maxpoolcolor and 0.20 and {\dim} and {\dim/2}};
				\node[anchor=east, rotate=90] at (1.95,-\dim/2 - 0.20, -\dim/4) {maxpool $(64, 120, 120)$};
				
				\draw pic at (2.10,-\dim/2,-\dim/4) [transform shape] {box=convcolor and 0.28 and {\dim} and {\dim/2}};
				\node[anchor=east, rotate=90] at (2.24,-\dim/2-0.20, -\dim/4) {conv + relu $(128,120,120)$};
				
				\draw pic at (2.43,-\dim/2,-\dim/4) [transform shape] {box=convcolor and 0.28 and {\dim} and {\dim/2}};
				\node[anchor=east, rotate=90] at (2.57,-\dim/2-0.20, -\dim/4) {conv + relu $(128,120,120)$};
				
				\draw pic at (2.76,-\dim/2,-\dim/4) [transform shape] {box=convcolor and 0.28 and {\dim} and {\dim/2}};
				\node[anchor=east, rotate=90] at (2.90,-\dim/2-0.20, -\dim/4) {conv + relu + in + drop $(128,120,120)$};
			}
			
			\foreach \dim in {0.71} {
				\draw pic at (3.09,-\dim/2,-\dim/4) [transform shape] {box=maxpoolcolor and 0.28 and {\dim} and {\dim/2}};
				\node[anchor=east, rotate=90] at (3.23,-\dim/2 - 0.20, -\dim/4) {maxpool $(128, 60, 60)$};
				
				\draw pic at (3.42,-\dim/2,-\dim/4) [transform shape] {box=convcolor and 0.40 and {\dim} and {\dim/2}};
				\node[anchor=east, rotate=90] at (3.62,-\dim/2-0.20, -\dim/4) {conv + relu $(256,60,60)$};
				
				\draw pic at (3.87,-\dim/2,-\dim/4) [transform shape] {box=convcolor and 0.40 and {\dim} and {\dim/2}};
				\node[anchor=east, rotate=90] at (4.07,-\dim/2-0.20, -\dim/4) {conv + relu $(256,60,60)$};
				
				\draw pic at (4.32,-\dim/2,-\dim/4) [transform shape] {box=convcolor and 0.40 and {\dim} and {\dim/2}};
				\node[anchor=east, rotate=90] at (4.52,-\dim/2-0.20, -\dim/4) {conv + relu + in + drop $(256,60,60)$};
			}
			
			\foreach \dim in {0.50} {
				\draw pic at (4.77,-\dim/2,-\dim/4) [transform shape] {box=maxpoolcolor and 0.40 and {\dim} and {\dim/2}};
				\node[anchor=east, rotate=90] at (4.97,-\dim/2 - 0.20, -\dim/4) {maxpool $(256,30,30)$};
			}
			
			\foreach \dim in {0.71} {
				\draw pic at (5.22,-\dim/2,-\dim/4) [transform shape] {box=maxpoolcolor and 0.40 and {\dim} and {\dim/2}};
				\node[anchor=east, rotate=90] at (5.42,-\dim/2 - 0.20, -\dim/4) {maxunpool $(256,60,60)$};
				
				\draw pic at (5.67,-\dim/2,-\dim/4) [transform shape] {box=convcolor and 0.40 and {\dim} and {\dim/2}};
				\node[anchor=east, rotate=90] at (5.87,-\dim/2-0.20, -\dim/4) {conv + relu $(256,60,60)$};
				
				\draw pic at (6.12,-\dim/2,-\dim/4) [transform shape] {box=convcolor and 0.40 and {\dim} and {\dim/2}};
				\node[anchor=east, rotate=90] at (6.32,-\dim/2-0.20, -\dim/4) {conv + relu $(256,60,60)$};
				
				\draw pic at (6.57,-\dim/2,-\dim/4) [transform shape] {box=convcolor and 0.28 and {\dim} and {\dim/2}};
				\node[anchor=east, rotate=90] at (6.71,-\dim/2-0.20, -\dim/4) {conv + relu + in + drop $(128,60,60)$};
			}
			
			\foreach \dim in {1} {
				\draw pic at (6.90,-\dim/2,-\dim/4) [transform shape] {box=maxpoolcolor and 0.28 and {\dim} and {\dim/2}};
				\node[anchor=east, rotate=90] at (7.04,-\dim/2 - 0.20, -\dim/4) {maxunpool $(128,120,120)$};
				
				\draw pic at (7.23,-\dim/2,-\dim/4) [transform shape] {concatbox=concatcolor and 0.40 and {\dim} and {\dim/2}};
				\node[anchor=east, rotate=90] at (7.43,-\dim/2 - 0.20, -\dim/4) {concat $(256,120,120)$};
				
				\draw pic at (7.68,-\dim/2,-\dim/4) [transform shape] {box=convcolor and 0.28 and {\dim} and {\dim/2}};
				\node[anchor=east, rotate=90] at (7.82,-\dim/2-0.20, -\dim/4) {conv + relu $(128,120,120)$};
				
				\draw pic at (8.01,-\dim/2,-\dim/4) [transform shape] {box=convcolor and 0.28 and {\dim} and {\dim/2}};
				\node[anchor=east, rotate=90] at (8.15,-\dim/2-0.20, -\dim/4) {conv + relu $(128,120,120)$};
				
				\draw pic at (8.34,-\dim/2,-\dim/4) [transform shape] {box=convcolor and 0.20 and {\dim} and {\dim/2}};
				\node[anchor=east, rotate=90] at (8.44,-\dim/2-0.20, -\dim/4) {conv + relu + in + drop $(64,120,120)$};
			}
			
			\foreach \dim in {1.41} {
				\draw pic at (8.59,-\dim/2,-\dim/4) [transform shape] {box=maxpoolcolor and 0.20 and {\dim} and {\dim/2}};
				\node[anchor=east, rotate=90] at (8.69,-\dim/2 - 0.20, -\dim/4) {maxunpool $(64,240,240)$};
				
				\draw pic at (8.84,-\dim/2,-\dim/4) [transform shape] {concatbox=concatcolor and 0.28 and {\dim} and {\dim/2}};
				\node[anchor=east, rotate=90] at (8.98,-\dim/2 - 0.20, -\dim/4) {concat $(128,240,240)$};
				
				\draw pic at (9.17,-\dim/2,-\dim/4) [transform shape] {box=convcolor and 0.20 and {\dim} and {\dim/2}};
				\node[anchor=east, rotate=90] at (9.27,-\dim/2-0.20, -\dim/4) {conv + relu $(64,240,240)$};
				
				\draw pic at (9.42,-\dim/2,-\dim/4) [transform shape] {box=convcolor and 0.20 and {\dim} and {\dim/2}};
				\node[anchor=east, rotate=90] at (9.52,-\dim/2-0.20, -\dim/4) {conv + relu $(64,240,240)$};
				
				\draw pic at (9.67,-\dim/2,-\dim/4) [transform shape] {box=convcolor and 0.14 and {\dim} and {\dim/2}};
				\node[anchor=east, rotate=90] at (9.74,-\dim/2-0.20, -\dim/4) {conv + relu + in + drop $(32,240,240)$};
			}
			
			\foreach \dim in {2} {
				\draw pic at (9.86,-\dim/2,-\dim/4) [transform shape] {box=maxpoolcolor and 0.14 and {\dim} and {\dim/2}};
				\node[anchor=east, rotate=90] at (9.93 + 0.05,-\dim/2 - 0.20, -\dim/4) {maxunpool $(32,480,480)$};
				
				\draw pic at (10.05,-\dim/2,-\dim/4) [transform shape] {concatbox=concatcolor and 0.20 and {\dim} and {\dim/2}};
				\node[anchor=east, rotate=90] at (10.15,-\dim/2 - 0.20, -\dim/4) {concat $(64,480,480)$};
				
				\draw pic at (10.30,-\dim/2,-\dim/4) [transform shape] {box=convcolor and 0.14 and {\dim} and {\dim/2}};
				\node[anchor=east, rotate=90] at (10.37,-\dim/2-0.20, -\dim/4) {conv + relu $(32,480,480)$};
				
				\draw pic at (10.49,-\dim/2,-\dim/4) [transform shape] {box=convcolor and 0.10 and {\dim} and {\dim/2}};
				\node[anchor=east, rotate=90] at (10.54,-\dim/2-0.50, -\dim/4) {conv + relu + drop $(16,480,480)$};
			}
			
			\foreach \dim in {2.83} {
				\draw pic at (10.64 + 0.05,-\dim/2,-\dim/4) [transform shape] {box=maxpoolcolor and 0.10 and {\dim} and {\dim/2}};
				\node[anchor=east, rotate=90] at (10.74 + 0.10,-\dim/2 - 0.20, -\dim/4) {maxunpool $(16,960,960)$};
				
				\draw pic at (10.84,-\dim/2,-\dim/4) [transform shape] {concatbox=concatcolor and 0.14 and {\dim} and {\dim/2}};
				\node[anchor=east, rotate=90] at (10.91 + 0.10,-\dim/2 - 0.20, -\dim/4) {concat $(32,960,960)$};
				
				\draw pic at (11.03,-\dim/2,-\dim/4) [transform shape] {box=convcolor and 0.14 and {\dim} and {\dim/2}};
				\node[anchor=east, rotate=90] at (11.10 + 0.10,-\dim/2-0.20, -\dim/4) {conv + relu $(30,960,960)$};
				
				\draw pic at (11.23+0.75,0 + \dim/4,-\dim/4) [transform shape] {box=convcolor and 0.06 and {\dim} and {\dim/2}};
				\node[anchor=north west, rotate=90] at (11.23 + 0.0 +0.75,\dim/4 - 0.10, \dim/4) {conv + sigmoid $(6,960,960)$};
				
				\draw pic at (11.23 + 0.75,-\dim - \dim/4,-\dim/4) [transform shape] {box=convcolor and 0.12 and {\dim} and {\dim/2}};
				\node[anchor=north west, rotate=90] at (11.23 + 0.06 + 0.75,-\dim -\dim/4, \dim/4) {conv $(24,960,960)$};
				
				\draw pic at (12.14 + 1,-\dim/2,-\dim/4) [transform shape] {box=inputcolor and 0.12 and {\dim} and {\dim/2}};
				\node[anchor=north west, rotate=90] at (12.14 + 0.06 + 1,-\dim/2, \dim/4) {concat $(30,960,960)$};
			}
			
		\end{tikzpicture}
		\caption{Visualisation of the used neural network architecture. Specified tensor sizes are indicative for a $960 \times 960$ image. The number of channels in the last four tensors is in function of the used skeleton architecture. Solid arrows represent tensor transfer, dashed arrows represent transfer of maxpool indices. conv = separable convolution, in=instance normalization, drop=dropout, concat=concatenation}
		\label{fig:network_architecture}
	\end{figure}
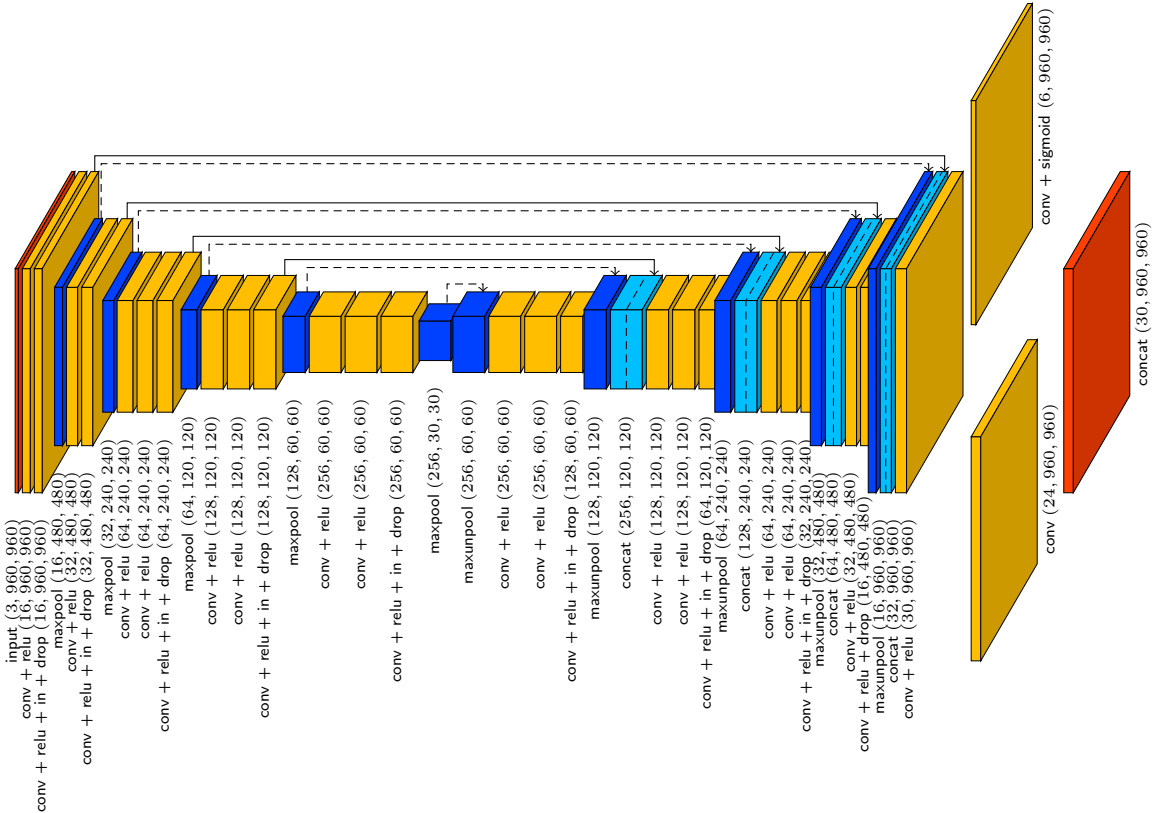	
	

Lastly, the 25 batch normalisation layers were replaced with respectively 8 instance normalisation layers and 9 dropout layers which were positioned before each maxpool or maxunpool layer. We used instance normalisation instead of batch normalisation to allow training and evaluating the network with batches of size 1.

The output of the network is a $w \times h \times 30$ tensor denoted $\hat{Y}$ for a training image with size $w \times h$. We use P to denote the indices of the (6) channels that represent the keypoint probability maps and $A$ to denote the (24) indices of the keypoint association maps. The loss function used during training is defined as

\begin{equation}
	\label{eq_a:loss_function}
	L = \; \theta_1 +
	\theta_2
	\underbrace{
		\dfrac{1}{wh|P|}
		\sum_{i \in P} || \hat{Y}_i - Y_i || ^2}
	_{\text{Location loss}}
	+ \theta_3
	\underbrace{\dfrac{1}{\sum_{i \in A}\lVert \mathds{1}\left[ Y_i \neq 0\right] \rVert}
		\sum_{j \in A}
		\bigg\lVert \mathds{1}\left[ Y_i \neq 0\right] \odot \left(\dfrac{\hat{Y}_i - Y_i}{\gamma}\right) \bigg\rVert^2}
	_{\text{Association loss}}
\end{equation}

where $|P|$ is the cardinality of $P$ (6 in this paper). $\mathds{1}[.]$ is the element-wise indicator function and $\odot$ is element-wise multiplication operator. Neglecting hyperparameters $\theta_1$, $\theta_2$ and $\theta_3$, the loss function consists out of a location loss and an association loss. The association loss only takes into account the squared difference between the predicted and the ground truth value if the ground truth value is nonzero. By doing so, the keypoint detection and keypoint association are decoupled from each other, which facilitates training to a large extent. The $\gamma$ hyperparameter in the association loss is used to scale the location loss and association loss appropriately and was set to 512.

\subsection{Training procedure}
Models are trained for 100 epochs of 1000 batches of a single image.  During training, the performance of the models was monitored based on the value of the loss function for a validation dataset and showed always convergence without signs of overfitting after 100 epochs. The training procedure was split into 50 epochs for initial training and 50 epochs for finetuning of the network. 

During training, extensive image augmentation was applied. Firstly, we applied hue, saturation and brightness jittering. Secondly, left-right flipping was applied, requiring to flip the left and right hook in the annotations. Furthermore, images were rotated with a random angle $\in [-15, 15]$ degrees. During rotation, it was ensured no pixels of the original image were removed by adjusting the image size and aspect ratio appropriately, which required training the network with batches of size one. Finally, random translation (0-5\% width/height) and scaling with a factor varying between 0.8 and 1.2 was applied. When during these modifications keypoints fell outside the image borders, they were removed from the annotations.

During the initial training, training instances are sampled at random and the hyperparameters $\theta_1$, $\theta_2$ and $\theta_3$ of \cref{eq_a:loss_function} are set to 0, 1 and 1, respectively. During the finetuning phase of the training procedure, however, we combine the principles of multi-loss weighing \citep{Groenendijk_2021} and curriculum learning \citep{Bengio_2009} to optimise the keypoint recovery rate $\eta_k$. The keypoint recovery rate $\eta_k$ quantifies the fraction of ground truth keypoints that is recovered by the pose-estimation pipeline after post-processing. During preliminary experiments, we found that optimising the loss function used during the phase of initial training with solvers such as ADAM \citep{Kingma_2014} does not result in maximisation of the network performance in terms of $\eta_k$. However, $\eta_k$ cannot be directly used as loss function. Therefore, the hyperparameters $\theta_1$, $\theta_2$ and $\theta_3$ of \cref{eq_a:loss_function} are adapted each tenth epoch to optimise the correlation between the loss and the keypoint recovery rate $\eta_k$.

To determine optimal values for $\theta_1$, $\theta_2$ and $\theta_3$, five augmented images were generated and evaluated for each instance of the training dataset. For each of these augmented images, the location loss and association loss are stored, after which new values for $\theta_1$, $\theta_2$ and $\theta_3$ are obtained by regression analysis. $\theta_1$ in \cref{eq_a:loss_function} does not have any effect on the training results, but improves the interpretability of the overall loss in terms of $\eta_k$. To prevent the location loss or the association loss to become neglected in the loss function, which can result in a fast and severe deterioration of $\eta_k$, the larger of $\theta_2$ and $\theta_3$ is required to be at most five times the value of the smaller of $\theta_2$ and $\theta_3$. Otherwise, an alternative linear regression is performed assuring this requirement is met. Similarly, if the regression analysis does not result in an $R^2$ over 0.05, or negative values for $\theta_2$ or $\theta_3$ are obtained, a simplified regression is performed so both $\theta_2$ and $\theta_3$ are assigned the same value. 

To perform curriculum learning during the finetuning phase of the training procedure, the $\eta_k$-values obtained are averaged per instance of the training dataset and used to determine a probability mass function (pmf), which is used to sample training instances. The pmf is defined such that the image with the lowest $\eta_k$ is sampled ten times more frequent in comparison to images for which all of the ground-truth keypoints are recovered ($\eta_k = 1$). 

\subsection{Skeleton assembly}
The final step of bottom-up pose estimation is skeleton assembly. To obtain skeletons, the neural network output is post-processed using a two-step procedure: (i) the identification of candidate keypoints; and (ii) keypoint association. Both steps use the output of the neural network, after smoothing with a $5 \times 5$ averaging kernel. Candidate keypoints are located by selecting all local maxima and applying quadratic interpolation to obtain sub-pixel resolution. Only local maxima with a probability above 0.4 are retained. In case the distance between two local maxima is less than 7 pixels, only the candidate keypoint with the highest probability is retained.

Once keypoint candidates are obtained, skeletons are constructed following the hierarchy of the pose-model. First, all dominant connections are assembled by associating their constituting keypoints. Once this is done, all detected central keypoints (withers) and rank-1 keypoints part of dominant connections (left hook, right hook and tail implant) which were not associated, are removed. This ensures that only valid skeletons containing at least one dominant connection will be constructed. Next, connections are made according to their order. To this end, first all non-dominant first order connections are assembled (top of head - withers), after which, all rank 1 keypoints which were not associated are removed. Thereafter, all second order connections can be assembled and again, once this is done, non-associated rank 2 keypoints are removed. If a skeleton would have an order over two, the latter assembly step is repeated until also the keypoints with the highest order are associated and the final skeletons are obtained.

To illustrate the algorithm used to associate keypoints into connections, consider the $k$th order connection $a \to b$ between keypoints of category $a$ and $b$, with $a$ being of rank $k-1$ and $b$ being of rank $k$. Let $\vec{a}_n$ be a keypoint of category $a$, with $n \in [1, \dots, N]$ and $N$ the number of detected keypoints of category $a$ that were retained after assembly of all $k-1$th order connections. Let $\vec{b}_m$ be a keypoint of category $b$, with $m \in [1, \dots, M]$ and $M$ the number of keypoint candidates of category $b$ found on the corresponding keypoint probability map. 

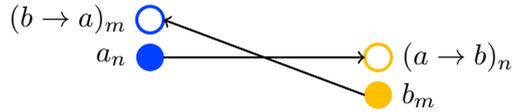
\begin{figure}[htbp]
	\center
	\begin{tikzpicture}[scale=1]
		\filldraw [customblue] (0,0) circle (5pt);
		\node[anchor=east, black] at (-5pt,0){$a_n$};
		\filldraw [customyellow] (3,-0.5) circle (5pt);
		\node[anchor=west, black] at (3cm + 5pt,-0.5){$b_m$};
		
		\draw [customblue, very thick] (0,0.5) circle (5pt);			
		\node[anchor=east, black] at (-5pt,0.5){$(b \to a)_m$};
		\draw [customyellow, very thick] (3, 0) circle (5pt);
		\node[anchor=west, black] at (3cm + 5pt,0){$(a \to b)_n$};
		
		\draw [->, black, thick] (0cm + 5pt,0) -- (3cm - 5pt, 0);
		\draw [->, black, thick] (3cm - 5 pt,-0.5) -- (0cm + 5pt,0.5);
		
	\end{tikzpicture}
	\captionsetup{width=0.5\textwidth}
	\caption{Detected keypoints and predicted locations for associated keypoints.}
	\label{fig:association_methodology}
\end{figure}

The first step to associate connections of type $a \to b$ is the prediction of the location of the complementary keypoints $b$ for all $\vec{a}_n$, denoted as $\vec{(a \to b)}_n$ and the locations of the complementary keypoints $a$ for all $\vec{b}_n$, denoted as $\vec{(b \to a)}_m$. The prediction of complementary keypoints is illustrated in \cref{fig:association_methodology}. To predict the coordinates of $\vec{(a \to b)}_n$ and $\vec{(b \to a)}_m$ we use

\begin{equation}
	\label{eq_a:a_to_b_n}
	\vec{(a \to b)}_n = \vec{a}_n + [\Delta^x_{a \to b}(J_2 \vec{a}_n), \Delta^y_{a \to b}(J_2 \vec{a}_n)]^\top
\end{equation}

\begin{equation}
	\label{eq_a:b_to_a_m}
	\vec{(b \to a)}_m = \vec{b}_m + [\Delta^x_{b \to a}(J_2 \vec{b}_m), \Delta^y_{b \to a}(J_2 \vec{b}_m)]^\top
\end{equation}

In which $\Delta^x_{a \to b}(J_2 \vec{a}_n)$ is the value of the association map $\Delta^x_{a \to b}$ at the position of $\vec{a}_n$, using quadratic interpolation to achieve sub-pixel precision. Since $\vec{a}_n$ contains image coordinates (column, row), while $\Delta^x_{a \to b}$ requires matrix coordinates (row, column), multiplication of $\vec{a}_n$ by the two-dimensional exchange matrix $J_2$, given by \cref{eq_a:exchange_matrix}, is required in order to get the correct values from $\Delta^x_{a \to b}$.

\begin{equation}
	\label{eq_a:exchange_matrix}
	J_2 = 
	\begin{bmatrix}
		0 & 1 \\
		1 & 0 \\
	\end{bmatrix}
\end{equation}

In a second step, keypoints are actually associated into connections. Therefore, we make use of an association penalty \cref{eq_a:association_penalty} which is computed based on the coordinates of the (candidate) keypoints. This penalty is calculated for each possible pair $(\vec{a}_n, \vec{b}_m)$ of keypoints, resulting in a $n \times m$ matrix $C$. To decide which keypoints should be associated with each other, we solve the assignment problem with cost matrix $C$ using a greedy approach. The first connection is made by associating the pair of keypoints resulting in the lowest association penalty. Amongst the remaining keypoints, a second connection is made between the pair of keypoints which in turns results in the lowest association penalty. This process is repeated until no keypoints are remaining for at least one of the keypoint categories. After connections are associated in this way, connections having an association penalty above 5\% of the diagonal length of the image are removed, since these connections probably were made erroneously.

\begin{equation}
	\label{eq_a:association_penalty}
	d(\vec{a}_n,\vec{b}_m) = \dfrac{
		\lVert \vec{(a \to b)}_n  - \vec{b}_m\rVert
		+ \lVert \vec{(b \to a)}_m - \vec{a}_n \rVert}
	{2}
\end{equation}

\begin{figure}[htbp]
    \center
    \begin{tikzpicture}[scale=0.8]
        \node[] at (1.5,6) {Ground truth};
        \draw [customblue, very thick] (0,5) circle (5pt);
        \filldraw [customyellow] (3,5) circle (5pt);
        \draw [black, thick, dashed] (5pt,5) -- (3cm-5pt,5);
        
        \filldraw [customblue] (0,4) circle (5pt);
        \filldraw [customyellow] (3,4) circle (5pt);
        \draw [black, thick] (5pt,4) -- (3cm-5pt,4);
        
        \filldraw [customblue] (0,3) circle (5pt);
        \filldraw [customyellow] (3,3) circle (5pt);
        \draw [black, thick] (5pt,3) -- (3cm-5pt,3);
        
        \filldraw [customblue] (0,2) circle (5pt);
        \filldraw [customyellow] (3,2) circle (5pt);	
        \draw [black, thick] (5pt,2) -- (3cm-5pt,2);
        
        \filldraw [customblue] (0,1) circle (5pt);
        \filldraw [customyellow] (3,1) circle (5pt);
        \draw [black, thick] (5pt,1) -- (3cm-5pt,1);
        
        \filldraw [customblue] (0,0) circle (5pt);
        \draw [customyellow, very thick] (3,0) circle (5pt);
        \draw [black, thick, dashed] (5pt,0) -- (3cm-5pt,0);
        
        \node[] at (6.5,6) {Hungarian association};
        
        \draw [customblue, very thick] (5,5) circle (5pt);
        \filldraw [customyellow] (8,5) circle (5pt);
        
        \filldraw [customblue] (5,4) circle (5pt);
        \filldraw [customyellow] (8,4) circle (5pt);
        \draw [black, thick] (5cm + 5pt,4) -- (8cm-5pt,5);
        
        \filldraw [customblue] (5,3) circle (5pt);
        \filldraw [customyellow] (8,3) circle (5pt);
        \draw [black, thick] (5cm + 5pt,3) -- (8cm-5pt,4);
        
        \filldraw [customblue] (5,2) circle (5pt);
        \filldraw [customyellow] (8,2) circle (5pt);	
        \draw [black, thick] (5cm + 5pt,2) -- (8cm-5pt,3);
        
        \filldraw [customblue] (5,1) circle (5pt);
        \filldraw [customyellow] (8,1) circle (5pt);
        \draw [black, thick] (5cm + 5pt,1) -- (8cm-5pt,2);
        
        \filldraw [customblue] (5,0) circle (5pt);
        \draw [customyellow, very thick] (8,0) circle (5pt);
        \draw [black, thick] (5cm + 5pt,0) -- (8cm-5pt,1);
        
        \node[] at (11.5,6) {Greedy association};
        
        \draw [customblue, very thick] (10,5) circle (5pt);
        \filldraw [customyellow] (13,5) circle (5pt);
        
        \filldraw [customblue] (10,4) circle (5pt);
        \filldraw [customyellow] (13,4) circle (5pt);
        \draw [black, thick] (10cm + 5pt,4) -- (13cm-5pt,4);
        
        \filldraw [customblue] (10,3) circle (5pt);
        \filldraw [customyellow] (13,3) circle (5pt);
        \draw [black, thick] (10cm + 5pt,3) -- (13cm-5pt,3);
        
        \filldraw [customblue] (10,2) circle (5pt);
        \filldraw [customyellow] (13,2) circle (5pt);	
        \draw [black, thick] (10cm + 5pt,2) -- (13cm-5pt,2);
        
        \filldraw [customblue] (10,1) circle (5pt);
        \filldraw [customyellow] (13,1) circle (5pt);
        \draw [black, thick] (10cm + 5pt,1) -- (13cm-5pt,1);
        
        \filldraw [customblue] (10,0) circle (5pt);
        \draw [customyellow, very thick] (13,0) circle (5pt);
        \draw [black, thick, dashed] (10cm + 5pt,0) -- (13cm-5pt,5);
    \end{tikzpicture}
    \captionsetup{width=0.8\textwidth}
    \caption{Comparison between using a Hungarian algorithm and our greedy algorithm to associate keypoints into connections. Open circles represent ground truth keypoints that were not detected. Dashed connections represent connections that cannot be made (ground truth) or are removed by the algorithm based on a threshold for the association penalty (greedy association).}
    \label{fig:keypoint_association_algorithms}
\end{figure}
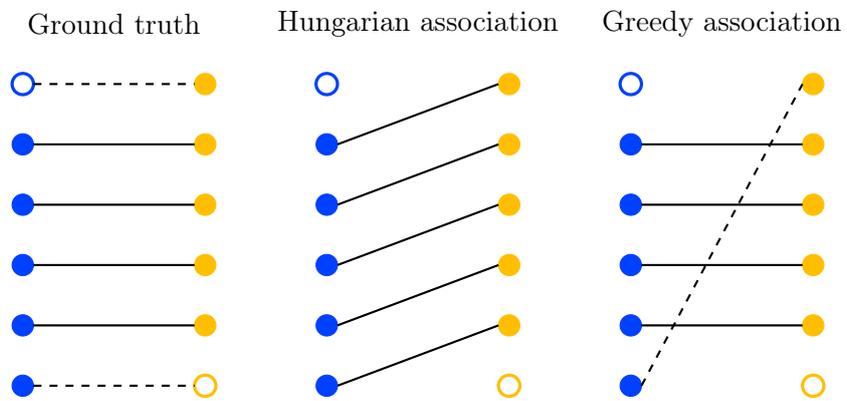

Our greedy method is fundamentally different from the Hungarian algorithm used by \citep{Psota_2019} to solve the assignment problem. If the animals in the pen show random orientation (\cref{fig:visualisation_skeletons_b} and \cref{fig:visualisation_skeletons_d}) both our greedy method and the Hungarian algorithm deliver satisfying results. If, however, multiple animals close to each other show parallel orientation (\cref{fig:visualisation_skeletons_b} and \cref{fig:visualisation_skeletons_d}) and some keypoints are not detected by the keypoint detector because they are occluded or unclear, the Hungarian algorithm results in a very poor skeleton assembly performance, as is illustrated in \cref{fig:keypoint_association_algorithms}. This is in contrast to our greedy method, which will still associate keypoints correctly in such situations.

    \section{Adaptive Kalman filter}
\label{sec:adaptive_kalman_filter}	

To perform pose tracking, we made use of an adaptive Kalman filter. In this appendix we elaborate the mathematical aspects of this Kalman filter. Firstly, the system is defined (\S\ref{sec:system_definition}), after which the equations corresponding with a standard Kalman filter are listed ((\S\ref{sec:standard_kalman_filter})). In \S\ref{sec:adaptive_kalman_filter_Yang_2006} the equations of the adaptive Kalman filter of \cite{Yang_2006} are explained. Finally, in \S\ref{sec:adaptive_kalman_filter_ours}, we present the modifications we did to the adaptive Kalman filter of \cite{Yang_2006} in order to obtain an adaptive Kalman filter that not only has a theoretical basis, but also performs good in practice.

\subsection{System definition}
\label{sec:system_definition}
The modelled system is defined by equations \cref{eq:system_transition} and \cref{eq:observation}, in which $\vec{x_n}$ is the unobserved state vector and $\vec{z_n}$ is the observed output vector. Common elements of the state vector are the position of the system, its velocity and its acceleration. The output vector on the other hand contains the observed variables, these are often elements of the state vector, however this is not a requirement and variables derived from the state vector are also frequently used.
\begin{equation}
	\label{eq:system_transition}
	\vec{x_n} = \Phi \vec{x_{n-1}} + \vec{w_n}
\end{equation}
\begin{equation}
	\label{eq:observation}
	\vec{z_n} = H \vec{x_n} + \vec{v_n}
\end{equation}
In \cref{eq:system_transition} and \cref{eq:observation}, $\Phi$ is the transition matrix which defines the dynamics of the system, quantifying how the current system state depends on the previous state in time. $H$ on the other hand is the observation matrix and defines the relationship between the system state and the observed variables in $\vec{z_n}$. Next to the previously described matrices and vectors, \cref{eq:system_transition} and \cref{eq:observation} also contain two noise vectors. The first one is $\vec{w_n}$, which quantifies the process noise, i.e. the behaviour of the system that cannot be captured by the term $\Phi \vec{x_{n-1}}$ in the state transition equation \cref{eq:system_transition}. The second noise vector is $\vec{v_n}$ and describes the observation noise, often referred to as the measurement errors. The noise vectors are considered to be normally distributed multivariate stochastic variables distributed according to
\begin{equation}
	\vec{w_n} \sim N \left( \vec{0}, Q \right)
\end{equation}
\begin{equation}
	\vec{v_n} \sim N \left( \vec{0}, R \right)
\end{equation}
In which $Q$ is the system noise covariance matrix and $R$ is the observation noise covariance matrix.

\subsection{Standard Kalman filter}
\label{sec:standard_kalman_filter}
The standard Kalman filter consists out of nine steps, each one with their own equations. To introduce notation, they are summarised bellow. Thereafter, we explain the risks of incorrect estimates of the observation noise covariance matrix $R$ and the system noise covariance matrix $Q$.

\begin{enumerate}
	\item Construct matrices $\Phi$ (transition matrix), $H$ (observation matrix), $Q$ (system noise covariance matrix) and $R$ (observation noise covariance matrix) based on system knowledge or prior knowledge
	\label{item:matrices_initialization}
	
	\item Generate initial estimates of the system state $\vec{\hat{x}_0}$ and the covariance matrix $P_0$
	\label{item:state_covariance_initialization}
	
	\item Predict the system state at the next point in time
	\label{item:state_prediction}
	\begin{equation}
		\label{eq:state_extrapolation_equation_2}
		\vec{\hat{x}_{n+1,n}} = \Phi \vec{\hat{x}_n}
	\end{equation}
	
	\item Predict the state covariance matrix $P_{n+1,n}$ at the next point in time
	\label{item:covariance_prediction}
	\begin{equation}
		\label{eq:covariance_extrapolation}
		P_{n+1,n} = \Phi P_n \Phi^\top + Q
	\end{equation}
	
	\item Proceed one step in time $\left( n+1 \mapsto n\right)$, and collect a new observation $\vec{z_n}$
	
	\item Calculate the Kalman gain
	\label{item:kalmen_gain}
	\begin{equation}
		\label{eq:kalman_gain}
		K_n =	
		P_{n,n-1} H^\top \left( H P_{n,n-1} H^\top + R \right)^{-1}		
	\end{equation}
	
	\item Update the system state estimate
	\begin{equation}
		\label{eq:state_update_2}
		\vec{\hat{x}_n} = \vec{\hat{x}_{n,n-1}} + K_n \left( \vec{z_n} - H \vec{\hat{x}_{n, n-1}} \right)
	\end{equation}
	
	\item Update the state covariance matrix
	\label{item:covariance_update}
	\begin{equation}
		\label{eq:covariance_update_2}
		P_n = \left( I - K_n H \right) P_{n,n-1} \left( I - K_n H \right)^\top
		+ K_n R K_n^\top\\
	\end{equation}
	\begin{equation}
		\label{eq:covariance_update_simplified_2}
		P_n = (I - K_n H) P_{n,n-1}
	\end{equation}
	
	\item Return to step \ref{item:state_prediction}	
\end{enumerate}

\begin{figure}[htb]
	\begin{center}
		\includegraphics[width=0.6\textwidth]{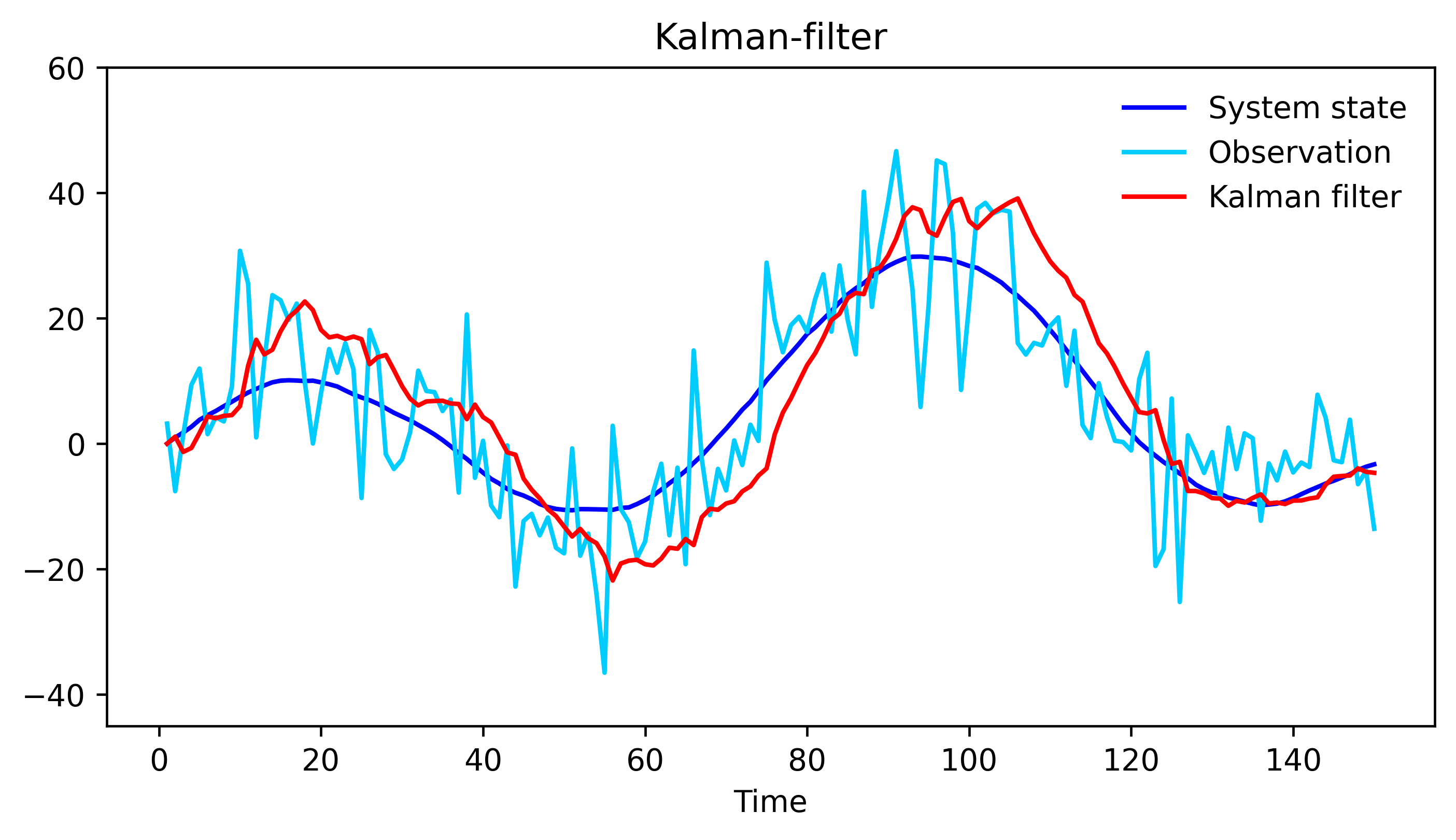}
		\captionsetup{width=0.6\textwidth}
		\caption{First order Kalman-filter with  correct estimates of the system covariance matrix $Q$ and the measurement noise covariance matrix $R$}
		\label{fig:kalman_correct}
	\end{center}
\end{figure}

An illustration of the results of applying the standard Kalman filter algorithm with correct estimates for $Q$ and $R$ is given in \cref{fig:kalman_correct}. Attention should be payed towards the construction of the covariance matrices describing the system noise ($Q$) and observation noise ($R$), since they define how the filter will react on subsequent observations. For example, if the variances in $R$, the covariance matrix of the observation noise, are low, the filter will quickly react on observations that deviate strongly from the predicted observations $H \vec{\hat{x}_{n,n-1}}$. If the variances in $R$ however are high, the filter will react way slower, since it knows the measurement errors can be quite high and therefore, the algorithm will be hesitant to adapt its state predictions abruptly after a strongly deviating measurement.

\begin{figure}[htb]
	\begin{center}
		\includegraphics[width=0.6\textwidth]{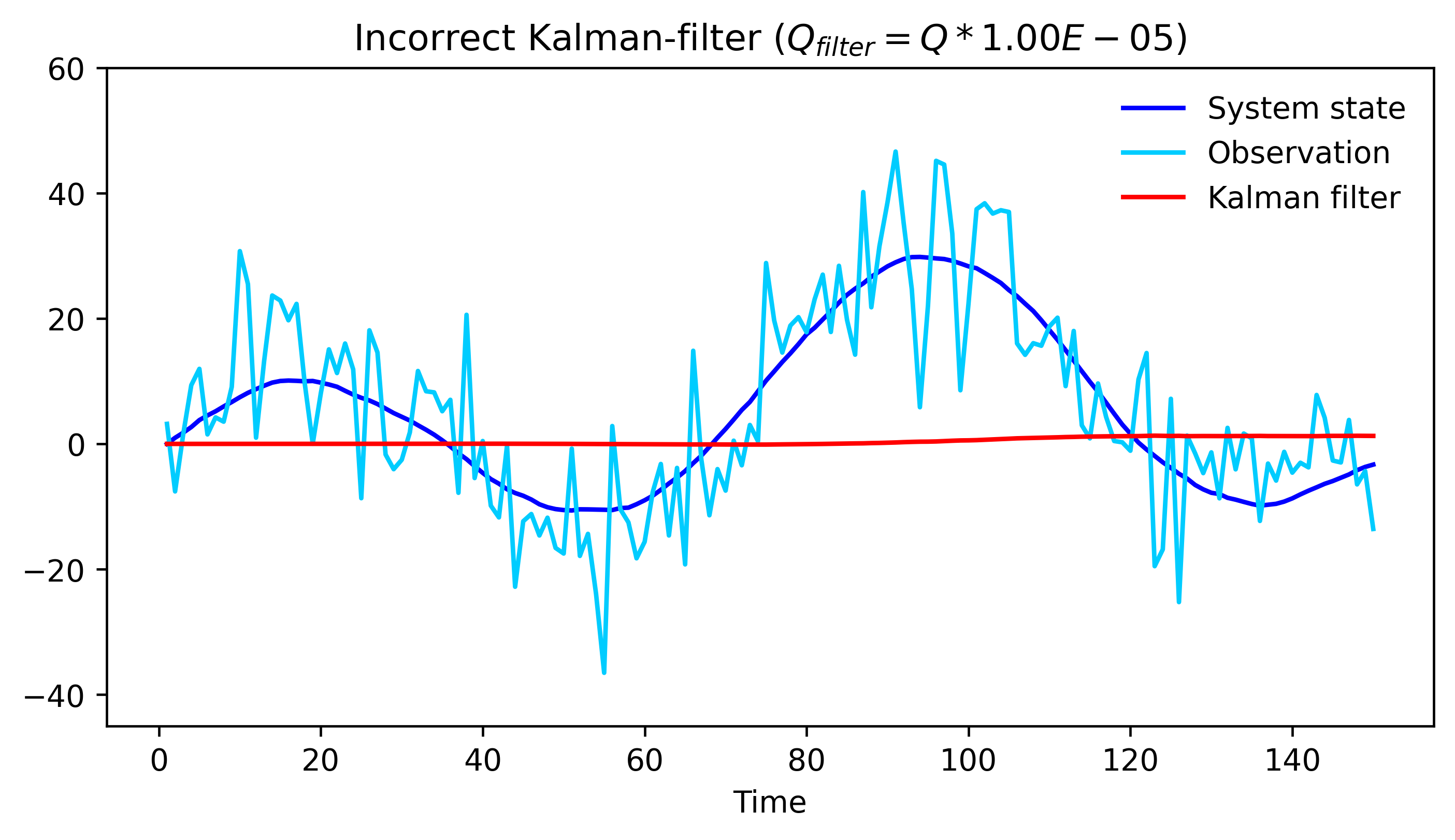}
		\captionsetup{width=0.6\textwidth}
		\caption{First order Kalman-filter with a too low estimate of the system covariance matrix $Q$, resulting in inadequate modelling of system dynamics not captured by the transition matrix $\Phi$}
		\label{fig:kalman_incorrect}		
	\end{center}
\end{figure}

Similarly, the covariances in $Q$ determine to which extent the filter can cope with differences between the expected behaviour of the system, quantified in the transition matrix $\Phi$ and the real world behaviour of the system. If the covariances in $Q$ are too low (fig. \ref{fig:kalman_incorrect}), the Kalman-filter will stiffly hold on to the system dynamics defined in $\Phi$, which often results in low performance, characterised by a systematic bias between predicted and observed measurements. On the other hand, if the covariances in $Q$ are too high, the filter can have the tendency to largely neglect the defined system dynamics, and therefore the filter also ignores the prior system knowledge quantified in the transition matrix $\Phi$. 

\subsection{Adaptive kalman filter}
\label{sec:adaptive_kalman_filter_Yang_2006}

As elaborated in \S\ref{sec:standard_kalman_filter}, correct estimates of the covariance matrices $Q$ and $R$, quantifying the system noise and the observation noise, respectively, are essential for a correct behaviour of a Kalman-filter. Whilst $R$ often can be estimated by using known measurement errors from used instruments or models, estimating the system noise $Q$ is mostly more complicated. For low-dimensional states, one can perform a trial-and-error procedure to estimate $Q$, but for more complex systems with multi-dimensional states $\vec{x_n}$ this is less obvious.

The solution to this issue is adaptive Kalman filtering, in which the covariance matrix of the predicted state $P_{n, n-1}$ is altered each time step in order to assure the modelled system noise corresponds on average with the actual system noise. By applying adaptive Kalman filtering, one even can relax the constraint specifying the system covariance matrix $Q$ should be independent of time, since adaptive Kalman filtering is able to handle possible time-dependent dynamics of $Q$. Various mathematical implementations of adaptive Kalman filtering are available, but here we will elaborate on the methodology developed by \cite{Yang_2006}.

In \cref{eq:state_update_2}, the prior state estimate is updated to the posterior state estimate. For this, there is made use of the difference between the actual observation $\vec{z_n}$ and the predicted observation $H \vec{\hat{x}_{n,n-1}}$, which is called the innovation $\vec{y_n}$ \cref{eq:innovation}
\begin{equation}
	\vec{y_n} = \vec{z_n} - H \vec{\hat{x}_{n,n-1}}	
	\label{eq:innovation}
\end{equation}
The covariance matrix of the innovation $\Sigma_{\vec{y_n}}$ is given by
\begin{equation}
	\label{eq:covariance_innovation}
	\Sigma_{\vec{y_n}} = R + HP_{n, n-1}H^\top
\end{equation}
In which can be observed $\Sigma_{\vec{y_n}}$ depends both on the measurement noise $R$, as well as on the uncertainty on the predicted state $\vec{x_{n, n-1}}$, given by $P_{n, n-1}$ and influenced by $Q$ via \cref{eq:covariance_extrapolation}.

\cite{Yang_2006} propose to multiply the covariance matrix of the predicted state $P_{n, n-1}$ with a factor $1/\alpha_n$, such that the theoretic covariance matrix of $\vec{y_n}$, which is used to compute the Kalman gain \cref{eq:kalman_gain}, is given by
\begin{equation}
	\label{eq:adaptive_kalman_derivation_1}
	\tilde{\Sigma}_{\vec{y_n}} = \dfrac{1}{\alpha_n} HP_{n, n-1}H^\top + R
\end{equation}
It is desired $\alpha_n$ takes a value such that the observed covariance matrix $\hat{\Sigma}_{\vec{y_n}}$ is equal to the theoretical covariance matrix $\tilde{\Sigma}_{\vec{y_n}}$ used to compute the Kalman gain.
\begin{equation}
	\label{eq:adaptive_kalman_derivation_2}
	\hat{\Sigma}_{\vec{y_n}} = \tilde{\Sigma}_{\vec{y_n}}
\end{equation}
Substitution of \cref{eq:adaptive_kalman_derivation_2} in \cref{eq:adaptive_kalman_derivation_1} results in
\begin{equation}
	\label{eq:adaptive_kalman_derivation_3}
	\hat{\Sigma}_{\vec{y_n}} = \dfrac{1}{\alpha_n} HP_{n, n-1}H^\top + R
\end{equation}
Multiplication of both sides of the equation $\alpha_n$, delivers
\begin{equation}
	\label{eq:adaptive_kalman_derivation_4}
	\alpha_n \hat{\Sigma}_{\vec{y_n}} =  HP_{n, n-1}H^\top + \alpha_n R
\end{equation}
\begin{equation}
	\label{eq:adaptive_kalman_derivation_5}
	\Leftrightarrow
	\alpha_n \left( \hat{\Sigma}_{\vec{y_n}} - R \right) =  HP_{n, n-1}H^\top
\end{equation}
If subsequently $R$ is added to both sides of \cref{eq:adaptive_kalman_derivation_5}, this results in
\begin{equation}
	\label{eq:adaptive_kalman_derivation_6}
	\alpha_n \left( \hat{\Sigma}_{\vec{y_n}} - R \right) + R =  HP_{n, n-1}H^\top + R
\end{equation}
Which can be simplified by using \cref{eq:covariance_innovation}
\begin{equation}	
	\label{eq:adaptive_kalman_derivation_7}
	\alpha_n \left( \hat{\Sigma}_{\vec{y_n}} - R \right) + R =  \Sigma_{\vec{y_n}}
\end{equation}
\begin{equation}	
	\label{eq:adaptive_kalman_derivation_8}
	\Leftrightarrow
	\alpha_n \left( \hat{\Sigma}_{\vec{y_n}} - R \right) =  \Sigma_{\vec{y_n}} - R
\end{equation}
If the trace of \cref{eq:adaptive_kalman_derivation_8} is taken, an expression for the optimal adaptive factor is obtained
\begin{equation}
	\label{eq:adaptive_kalman_derivation_9}
	\Leftrightarrow
	\alpha_n  = \dfrac{tr(\Sigma_{\vec{y_n}} - R)}{tr( \hat{\Sigma}_{\vec{y_n}} - R )}
\end{equation}
In practical applications however, it is undesired to shrink $P_{n,n-1}$ and hence we want $\alpha_n \leq 1$. Therefore, the expression for $\alpha_n$ is reformulated to
\begin{equation}
	\label{eq:optimal_adaptive_factor}
	\alpha_n = 
	\begin{cases}
		1 & tr( \hat{\Sigma}_{\vec{y_n}}) < tr(\Sigma_{\vec{y_n}})\\
		\dfrac{tr(\Sigma_{\vec{y_n}} - R)}{tr( \hat{\Sigma}_{\vec{y_n}} - R )} & otherwise
	\end{cases}       
\end{equation}
Since both the numerator and the denominator of \cref{eq:optimal_adaptive_factor} contain $R$, an approximative formula can be formulated as
\begin{equation}
	\label{eq:optimal_adaptive_factor_simplified}
	\alpha_n = 
	\begin{cases}
		1 & tr( \hat{\Sigma}_{\vec{y_n}}) < tr(\Sigma_{\vec{y_n}})\\
		\dfrac{tr(\Sigma_{\vec{y_n}})}{tr( \hat{\Sigma}_{\vec{y_n}})} & otherwise
	\end{cases}       
\end{equation}
In some cases, \cref{eq:optimal_adaptive_factor} can result in a negative value for $\alpha_n$, which makes no sense since (co)variances should always be greater than or equal to zero. A negative value for $\alpha_n$ is obtained when $tr(\hat{\Sigma}_{\vec{y_n}}) < tr(R)$ and hence, the denominator becomes negative. In this case, \cref{eq:optimal_adaptive_factor_simplified} can be used instead of \cref{eq:optimal_adaptive_factor}, since \cref{eq:optimal_adaptive_factor_simplified} will always result in a positive value of $\alpha_n$

In order to compute $\alpha_n$, it is thus required to know $\Sigma_{\vec{y_n}}$, $\hat{\Sigma}_{\vec{y_n}}$ and $R$. $R$ is defined during the construction of the Kalman-filter and $\Sigma_{\vec{y_n}}$ can easily be obtained by using \cref{eq:covariance_innovation}. $\hat{\Sigma}_{\vec{y_n}}$, however, should be estimated based on the observed innovations(s) $\vec{y_n}$. Assuming the expectation value of the innovation $\vec{y_n}$ equals zero
\begin{equation}
	\label{eq:expectation_value_involution}
	E\left[ \vec{y_n} \right] = \vec{0}
\end{equation}
The covariance matrix of $\vec{y_n}$ corresponds with
\begin{equation}
	\Sigma_{\vec{y_n}} = E\left[ \vec{y_n} \vec{y_n}^\top \right] 
\end{equation}
Therefore, to compute an estimate for $\hat{\Sigma}_{\vec{y_n}}$, one can use moving windows of size $N$
\begin{equation}
	\label{eq:observed_state_variance_moving_window}
	\hat{\Sigma}_{\vec{y_n}} = \dfrac{1}{N - 1} \sum_{i=0}^{N - 1} \vec{y_{n-i}} \vec{y_{n-i}}^\top
\end{equation}
However, this strategy entails the risk $\hat{\Sigma}_{\vec{y_n}}$ is not estimated accurately since averaging $\vec{y_{n}} \vec{y_{n}}^\top$ over $N$ time steps can result in unwanted smoothing. Therefore, the preferred approach to estimate $\hat{\Sigma}_{\vec{y_n}}$ is by using only the involution of the current time step
\begin{equation}
	\label{eq:observed_state_variance_single_point}
	\hat{\Sigma}_{\vec{y_n}} = \vec{y_n} \vec{y_n}^\top
\end{equation}

\begin{figure}[htbp]
	\begin{center}
		\includegraphics[width=0.7\textwidth]{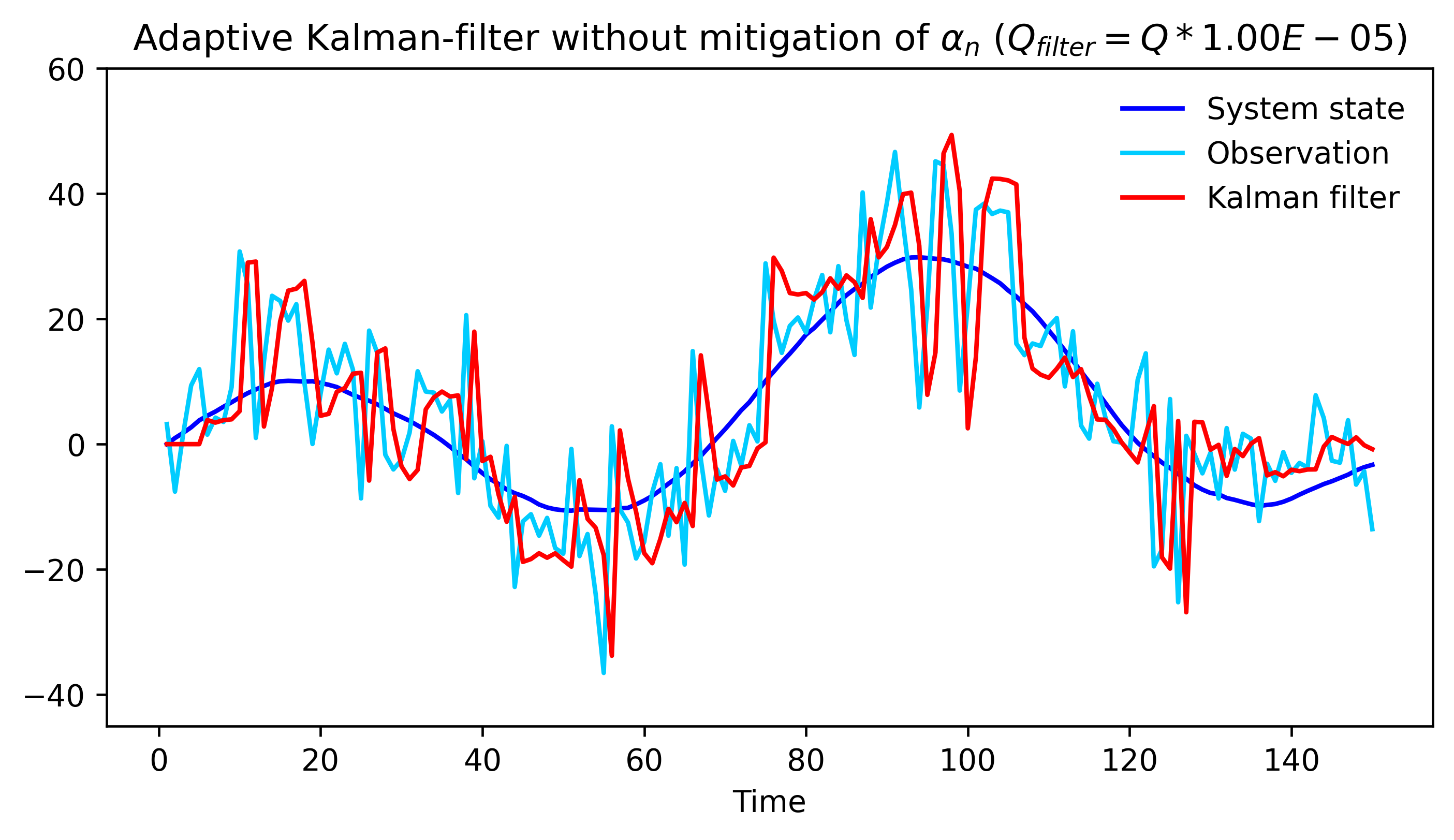}
		\captionsetup{width=0.7\textwidth}
		\caption{First order adaptive Kalman-filter using \cref{eq:observed_state_variance_single_point} to compute $\hat{\Sigma}_{\vec{y_n}}$ without mitigation of the resulting values for $\alpha_n$, resulting in strong overfitting of the observations.}
		\label{fig:kalman_adaptive_no_mitigation}		
	\end{center}
\end{figure}

Despite the above derivation of formulas for an adaptive Kalman-filter is theoretically solid, the performance during simulation studies does not meet the expectations of a good adaptive Kalman-filter. In practice, the adaptive Kalman-filter has the tendency to overestimate the system noise and therefore, putting too much importance into the most recent measurement and largely neglecting the specified system dynamics (fig. \ref{fig:kalman_adaptive_no_mitigation}). Therefore, adaptations to the algorithm of \cite{Yang_2006} are necessary in order to obtain a useable adaptive Kalman-filter.

\subsection{Improved adaptive Kalman filter}
\label{sec:adaptive_kalman_filter_ours}
The adaptation we propose to the algorithm of \cite{Yang_2006} is a mitigation strategy of $\alpha_n$. After calculating $\alpha_n$, the effect of alpha is mitigated by a factor $\gamma_n$ according to
\begin{equation}
	\label{eq:alpha_mitigation_strategy}
	\alpha_n \mapsto 1 - \gamma_n (1 - \alpha_n)
\end{equation}
Hereby, $\gamma_n$ is a mitigation factor based on the sign of the innovations during the last $m$ time steps. If the innovations have constantly the same sign during the time period covered by the moving window, this is an indication the Kalman-filter shows a significant bias when compared to the actual process and therefore, no mitigation of the effect of $\alpha_n$ is required, which corresponds with $\gamma_n$ being equal to one. If however half of the time the involutions have a negative sign whilst the other half of the time they have a positive sign, the Kalman-filter models the system well and by consequence, no modification of the state covariance matrix $P_{n, n-1}$ is necessary, which can be obtained by assigning a value of zero to $\gamma_n$. To obtain $\gamma_n$ in the case only the $i$-th dimension of the Kalman-filter would be taken into consideration ($\gamma_{n,i}$), \cref{eq:alpha_mitigation_factor_single_dim} can be used, in which $y_{n-j,i}$ is the $i$-th element of the involution vector $\vec{y_{n-j}}$ at time $n-j$.
\begin{equation}
	\label{eq:alpha_mitigation_factor_single_dim}
	\gamma_{n,i} = 
	\dfrac{1}{m}
	\left|
	\sum_{j=0}^{m-1}\dfrac{y_{n-j,i}}{\left| y_{n-j,i} \right|}
	\right|
\end{equation}
In \cref{eq:alpha_mitigation_factor_single_dim}, the average involution sign is computed over the past $m$ time steps. Since it is only the inequality between the number of positive and negative involutions that is relevant with respect to the mitigation factor $\gamma_{n,i}$, the absolute value of the obtained average is calculated, resulting in a most suited mitigation factor $\gamma_{n,i} \in [0, 1]$ if only dimension $i$ of the Kalman-filter would have been relevant. However, typical Kalman filters have multiple observation dimensions which are all relevant from the perspective of dynamic Kalman filtering and therefore, a final value of $\gamma_n$ in the case of a Kalman-filter with an observation vector $\vec{z_n}$ with dimension $k$, is obtained by averaging the different $\gamma_{n,i}$, as is shown in \cref{eq:alpha_mitigation_factor}.
\begin{equation}
	\label{eq:alpha_mitigation_factor}
	\begin{split}
		\gamma_n =&
		\dfrac{1}{k} \sum_{i=1}^{k}
		\gamma_{n,i}\\
		=&
		\dfrac{1}{km} \sum_{i=1}^{k}
		\left|
		\sum_{j=0}^{m-1}\dfrac{y_{n-j,i}}{\left| y_{n-j,i} \right|}
		\right|\\
	\end{split}
\end{equation}

\begin{figure}[htb]
	\begin{center}
		\includegraphics[width=0.7\textwidth]{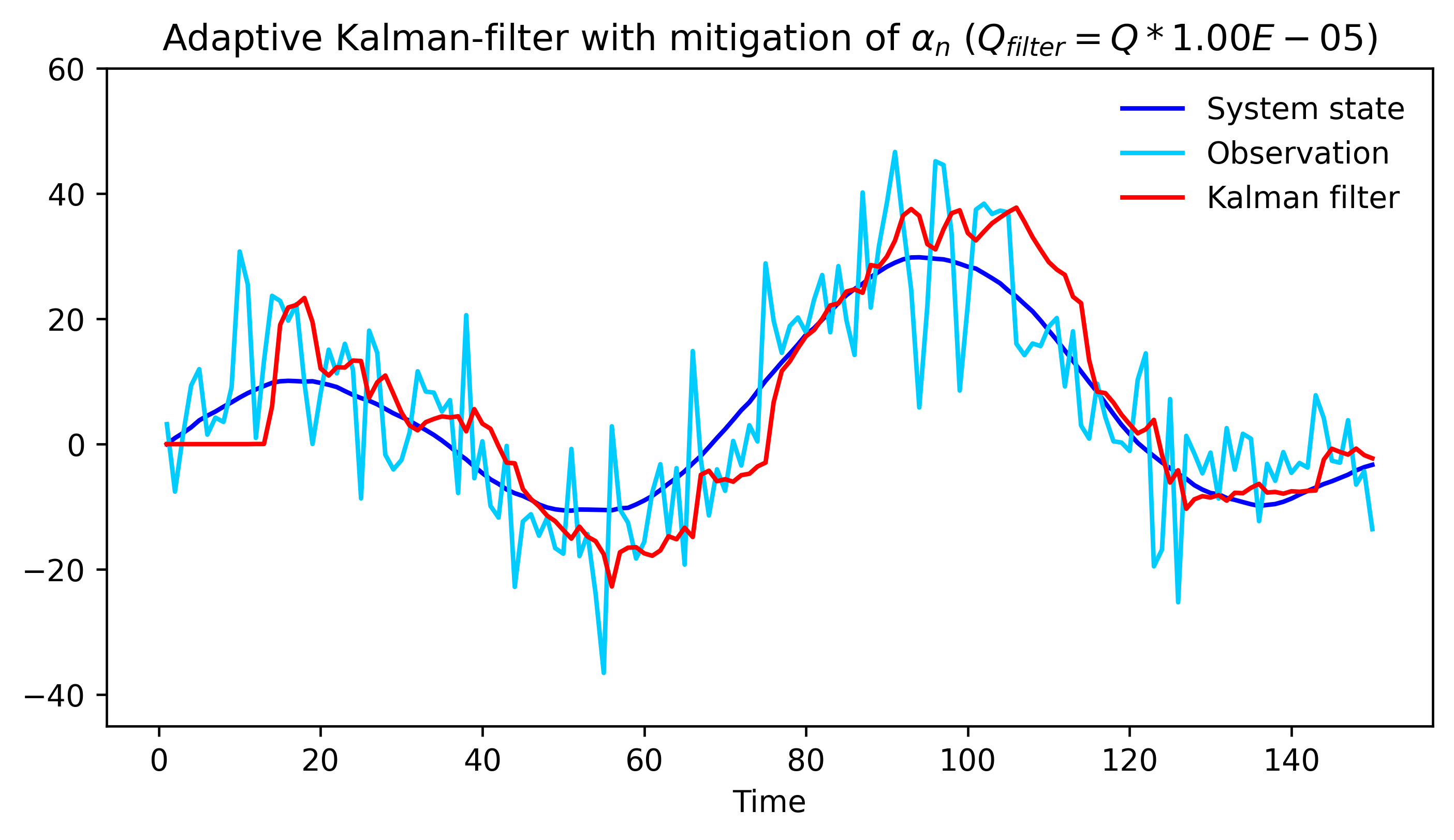}
		\captionsetup{width=0.7\textwidth}
		\caption{First order adaptive Kalman-filter using \cref{eq:observed_state_variance_single_point} to compute $\hat{\Sigma}_{\vec{y_n}}$, mitigating the resulting values for $\alpha_n$ afterwards with \cref{eq:alpha_mitigation_strategy}, resulting in correct adaptation of the filter to the system dynamics not captured by the system transition matrix $\Phi$}
		\label{fig:kalman_adaptive_mitigation}				
	\end{center}
\end{figure}

In contrast to \cref{eq:observed_state_variance_moving_window}, using \cref{eq:observed_state_variance_single_point} in combination with \cref{eq:alpha_mitigation_factor} limits the risk of unwanted smoothing of $\alpha_n$ to a large extent. Consider a process that is at time $t$ modelled correctly by a given Kalman-filter with system covariance matrix $Q$. At time $t + 1$, however, all covariances of $Q$ increase abruptly with a factor $10^5$. When \cref{eq:observed_state_variance_moving_window} would be used, some lag time is present before $\alpha_n$ takes on lower values and the system state covariance matrix $P_{n, n-1}$ will be enlarged. The presence of this lag time is desired, since it is unwanted that the Kalman-filter neglects all historical information by inflating $P_{n, n-1}$ largely as a result of a single deviating measurement. After some time steps, the estimated innovation covariance matrix $\hat{\Sigma}_{\vec{y_n}}$ will gradually start to increase, resulting in inflation of $P_{n, n-1}$, as desired. However, the disadvantage of using \cref{eq:observed_state_variance_moving_window} is that $P_{n, n-1}$ will be inflated too long as a result smoothing of $\hat{\Sigma}_{\vec{y_n}}$ due to the averaging operation performed in \cref{eq:observed_state_variance_moving_window}. This persisted inflation of $P_{n, n-1}$ results in overfitting of the modelled system, leading to to large innovations. By consequence, the covariances in $P_{n, n-1}$ are unlikely to decrease again, resulting to persistent overfitting.

In order to avoid this persistent overfitting, $\alpha_n$ should take on values close to 1 nearly immediate when the innovations are in line with the values that would be expected based on the system state covariance matrix $P_{n, n-1}$. Hence, at first glance, \cref{eq:observed_state_variance_single_point} seems to be a reasonable alternative for \cref{eq:observed_state_variance_moving_window}. However, when using \cref{eq:observed_state_variance_single_point}, the adaptive Kalman-filter has no initial lag time, and by consequence, a single deviating measurement can lead to a strong inflation of $P_{n, n-1}$, which will sign to the Kalman-filter it has to neglect historic information and once again will result in overfitting. Again, once the Kalman-filter starts overfitting the process, it is very unlikely for $P_{n, n-1}$ to decrease again, since overfitting of the process will lead to larger innovations, reinforcing the overfitting of the process continuously.

\begin{figure}[htb]
	\begin{center}
		\includegraphics[width=0.7\textwidth]{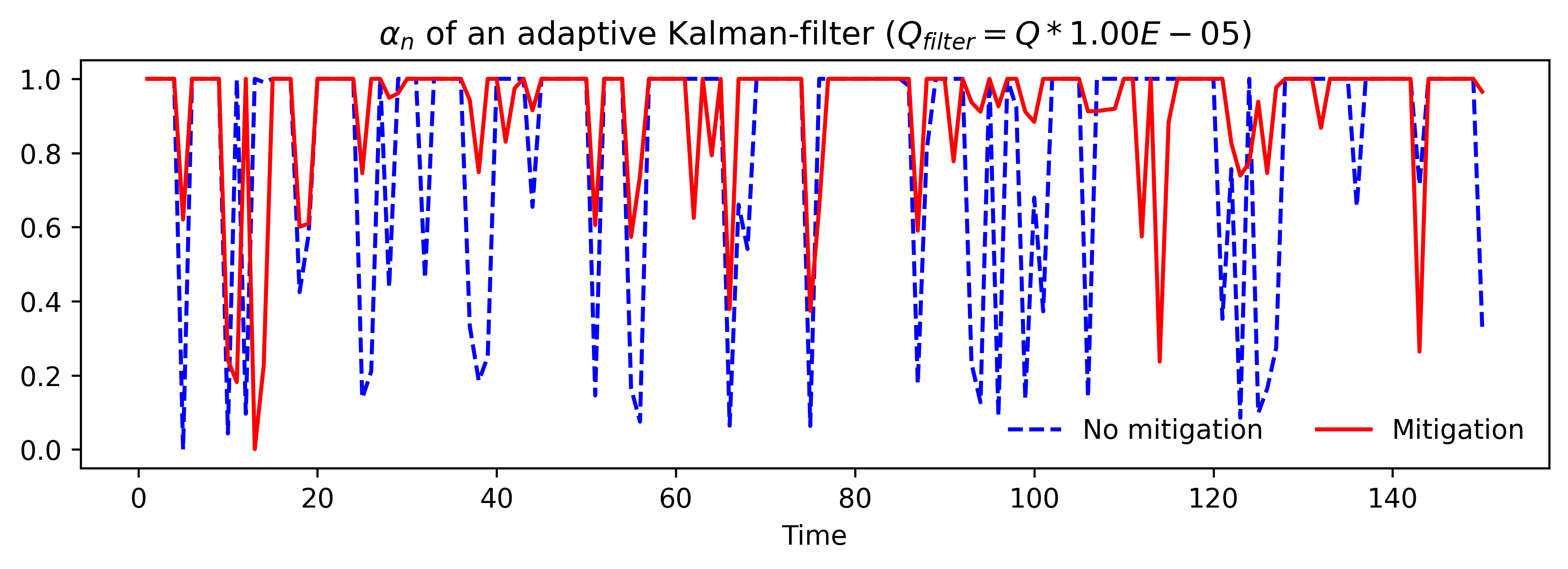}		
		\captionsetup{width=0.7\textwidth}	
		\caption{Comparison between the values for $\alpha_n$ used by an adaptive Kalman-filter without mitigation of $\alpha_n$ (fig. \ref{fig:kalman_adaptive_no_mitigation}) and those used by an adaptive Kalman-filter with mitigation of $\alpha_n$ (fig. \ref{fig:kalman_adaptive_mitigation}). It is clear from the figure that the unmitigated values for $\alpha_n$ more frequently take on values close to zero, causing overfitting of the observations}	
		\label{fig:kalman_alpha_values}
	\end{center}
\end{figure}

By using \cref{eq:observed_state_variance_single_point} in combination with \cref{eq:alpha_mitigation_strategy} and \cref{eq:alpha_mitigation_factor}, the advantages of both \cref{eq:observed_state_variance_moving_window} and \cref{eq:observed_state_variance_single_point} are exploited, while the disadvantages are avoided (fig. \ref{fig:kalman_adaptive_mitigation} and fig. \ref{fig:kalman_alpha_values}). When, starting from time $t + 1$, strongly deviating measurements would be observed, the mitigation of $\alpha_n$ by \cref{eq:alpha_mitigation_strategy} and \cref{eq:alpha_mitigation_factor} will introduce some lag time before the system state covariance matrix $P_{n, n-1}$ will start to inflate. On the other hand, as soon as $P_{n, n-1}$ has been inflated to the right amount, using \cref{eq:observed_state_variance_single_point} to estimate $\hat{\Sigma}_{\vec{y_n}}$ will immediately result in values of $\alpha_n$ close to 1, avoiding persisted inflation of $P_{n, n-1}$. Moreover, using \cref{eq:alpha_mitigation_strategy} and \cref{eq:alpha_mitigation_factor} to mitigate $\alpha_n$, will in the case of strong overfitting of the process, strongly mitigate $\alpha_n$, since $\gamma_n$ will take on values close to zero because one can expect an approximately equal number of positive and negative innovations. Hence, when the Kalman-filter is strongly overfitting the process due to a hyperinflated $P_{n, n-1}$, the mitigation strategy specified by \cref{eq:alpha_mitigation_strategy} and \cref{eq:alpha_mitigation_factor} will stop the continuous inflation of $P_{n, n-1}$ and allow the Kalman-filter to consolidate knowledge from the past observations in order to lower the covariances in $P_{n, n-1}$ and therefore do more stable and reliable predictions for the system state $\vec{x_{n+1}}$ at the next point in time.

\end{document}